\documentclass[10pt,journal,compsoc]{IEEEtran}

\usepackage{amsmath}
\usepackage{times}
\usepackage[utf8]{inputenc}
\usepackage{graphicx,subcaption}
\usepackage{color}
\usepackage[dvipsnames]{xcolor}
\usepackage{multirow}
\usepackage{amsfonts}
\usepackage{amssymb}
\usepackage{mathrsfs}
\usepackage{caption}
\usepackage{booktabs}
\usepackage{url}
\usepackage{mathrsfs}
\usepackage{fourier}

\usepackage{algorithm}
\usepackage{algpseudocode}

\usepackage{tabularx,booktabs}
\newcolumntype{C}{>{\centering\arraybackslash}X} 
\usepackage{lipsum}
\usepackage{etoolbox}
\usepackage{tikz}
\usetikzlibrary{tikzmark}
\usetikzlibrary{calc}
\usepackage{lscape}

\usepackage{rotating}
\usepackage{bbm}
\usepackage{latexsym}
\newtheorem{definition}{Definition}
\newtheorem{theorem}{Theorem}

\newtheorem{proposition}{Proposition}

\usepackage{mathtools} 
\usepackage{booktabs} 
\usepackage{tikz} 

\ifCLASSOPTIONcompsoc
  \usepackage[nocompress]{cite}
\else
  \usepackage{cite}
\fi

\begin{document}
\title{A Physics inspired Functional Operator for Model Uncertainty Quantification in the RKHS}

\author{Rishabh~Singh,~\IEEEmembership{Student Member,~IEEE,}
        and~Jose~C.~Principe,~\IEEEmembership{Life~Fellow,~IEEE}
\IEEEcompsocitemizethanks{\IEEEcompsocthanksitem R. Singh is with the Department
of Electrical and Computer Engineering, University of Florida, Gainesville,
FL, 32607.\protect\\
E-mail: rish283@ufl.edu
\IEEEcompsocthanksitem J. Principe is with the Department
of Electrical and Computer Engineering, University of Florida. E-mail: principe@cnel.ufl.edu}}




\IEEEtitleabstractindextext{%
\begin{abstract}
Accurate uncertainty quantification of model predictions is a crucial problem in machine learning. Existing Bayesian methods, being highly iterative, are expensive to implement and often fail to accurately capture a model’s true posterior because of their tendency to select only central moments. We propose a fast single-shot uncertainty quantification framework where, instead of working with the conventional Bayesian definition of model weight probability density function (PDF), we utilize physics inspired \textit{functional operators} over the projection of model weights in a reproducing kernel Hilbert space (RKHS) to quantify their uncertainty at each model output. The RKHS projection of model weights yields a potential field based interpretation of model weight PDF which consequently allows the definition of a functional operator, inspired by perturbation theory in physics, that performs a moment decomposition of the model weight PDF (the potential field) at a specific model output to quantify its uncertainty. We call this representation of the model weight PDF as the \textit{quantum information potential field} (QIPF) of the weights. The extracted moments from this approach automatically decompose the weight PDF in the local neighborhood of the specified model output and determine, with great sensitivity, the local heterogeneity of the weight PDF around a given prediction. These moments therefore provide sharper estimates of predictive uncertainty than central stochastic moments of Bayesian methods. Experiments evaluating the error detection capability of different uncertainty quantification methods on covariate shifted test data show our approach to be more precise and better calibrated than baseline methods, while being faster to compute.
\end{abstract}
\begin{IEEEkeywords}
Uncertainty quantification, RKHS, neural networks, perturbation, physics, moment decomposition, covariate shift.
\end{IEEEkeywords}}

\maketitle


\noindent Deep neural network (DNN) models have become the predominant choice for pattern representation in a wide variety of machine learning applications due to their remarkable performance advantages in the presence of large amount data \cite{lec}. The increased adoption of DNNs in safety critical and high stake problems such as medical diagnosis, chemical plant control, defense systems and autonomous driving has led to growing concerns within the research community on the \textit{performance trustworthiness} of such models \cite{kend, lund}. This becomes particularly imperative in situations involving data distributional shifts or the presence of out of distribution data (OOD) during testing. The model may lack robustness in such cases due to poor choice of training parameters or lack of sufficiently labeled training data, especially since machine learning algorithms do not have extensive prior information like humans to deal with such situations \cite{amod}. An important way through which trust in the performance of machine learning algorithms (particularly DNNs) can be established is through accurate techniques of \textit{predictive uncertainty quantification}, which allow practitioners to determine how much they should rely on model predictions.\par

Although there have been several categories of methods developed in the recent years, the Bayesian approach \cite{mack, neal, bishop} has for long been regarded as the gold standard for natural representation of uncertainty in neural networks. However, due to their highly iterative nature, they are unable to scale to modern applications and often fail to capture the true data distribution \cite{laks}. Moreover, in practice, Bayesian approaches are mostly only able select measurements of \textit{central tendency} (or central moments) from the posterior thus significantly reducing their sensitivity in the quantification of local uncertainty information which is especially prevalent in modern applications where models can capture very complex patterns with significant local variations.\par

The reproducing kernel Hilbert space (RKHS) is well known in machine learning research to enable an elegant, mathematically rigorous and powerful functional representation of non-linear local patterns in input data, without having to explicitly know those patterns. This is made possible through inner products that can be computed in the input space using the kernel reproducing property. Two fundamental theorems governing the properties of the RKHS are the Moore-Aronszajn theorem \cite{moore, aron11}, which states that a uniquely defined RKHS exists for every symmetric non-negative definite function, and the representer theorem \cite{schol}, which reduces optimization problems formulated in the RKHS (which may be high or infinite dimensional space) into finite dimensional spaces on scalar coefficients, thereby simplifying the analysis. It is therefore well established that the linear structure of the RKHS can be directly used to build non-linear deterministic functions that can bypass the need for highly iterative training procedures such as those used in artificial neural networks. In a similar spirit, our goal in this paper is to propose a framework that utilizes the RKHS to build a \textit{non-linear operator that provides a deterministic description of uncertainty} associated with a given function/model (that has been optimized over the training set) and bypasses the need for iterative Bayesian and ensemble approaches. We consider neural networks in this paper without loss of generality towards other learning algorithms. \par

Fundamentally, the proposed RKHS based operator quantifies model uncertainty by evaluating the \textit{variability} in the local density of training points around a test-set prediction point, with high variability indicating high uncertainty in the prediction. For a trained model, this is accomplished by evaluating the variability of the weight PDF (which quantifies the training data distribution) in the local neighborhood of the model prediction point. From a Bayesian perspective, this measures the degree of regularization of weights around an output, albeit locally. To this end, our approach begins by defining an RKHS based functional constructed using model weights, denoted by $\psi_\mathbf{w}(.)$, that evaluates the weight PDF at any point. The evaluation of $\psi_\mathbf{w}(.)$ at any prediction point $y^*$ (denoted by $\psi_\mathbf{w}(y^*)$) is taken to be an approximation of the predictive PDF of the model $p(y^*|\mathbf{w}, \mathbf{x}^*)$ \footnote{We use the standard shorthand notation $p(y^*)$ to represent $P(Y = y^*)$.}, at $y^*$. The next step is to add a small perturbation $\Delta{y^*}$ to $y^*$ to evaluate $\psi_\mathbf{w}(y^* + \Delta{y^*})$ which quantifies the local variability of weight PDF in the vicinity of $y^*$.\par

\begin{figure}[!t]
    \centering\includegraphics[scale = 0.3]{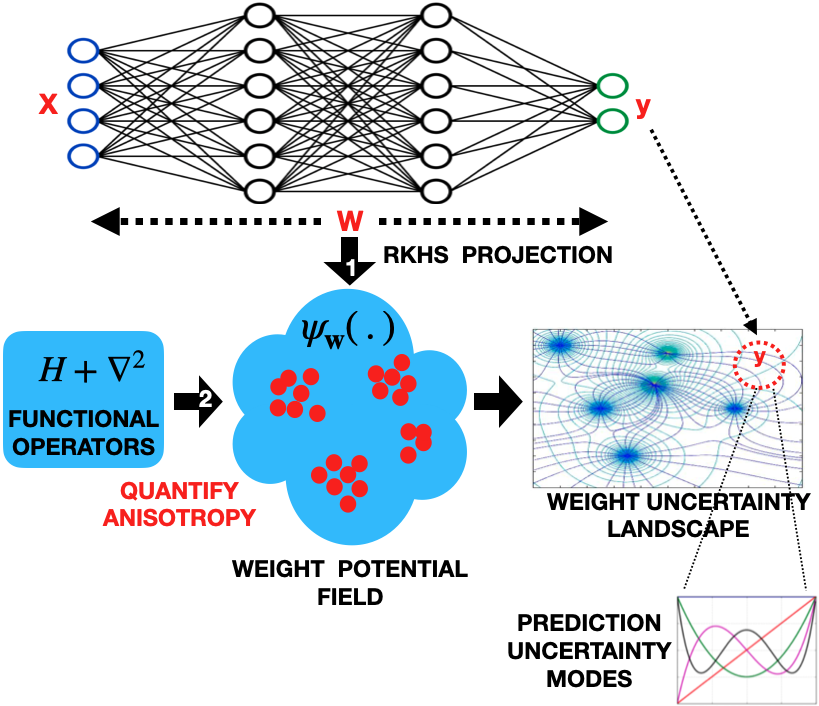}
    \caption{Proposed approach for neural network uncertainty quantification: (1) RKHS projection of model weights ($\psi_\mathbf{w}(.)$) creates a potential field quantifying its PDF. (2) Functional operators act on $\psi_\mathbf{w}(.)$ to create a multi-moment uncertainty landscape/function which when evaluated on a prediction quantifies its uncertainty.}
    \label{fm}
\end{figure}

The theory of operators in the Hilbert space has been well developed in quantum mathematics. Taking inspiration from it, we specifically leverage the spatial Laplacian operator for quantifying the local variability (or anisotropy) of the RKHS based weight PDF, $\psi_\mathbf{w}(.)$. Such an approach to link operator theory with the RKHS is elegant because of the advantage that solutions in the RKHS are computed in the input space (kernel trick). The weight PDF Laplacian also allows for an orthogonal local mode decomposition, through perturbation theory, so that the higher order modes are automatically centered in regions of the weight space that have high anisotropy. This methodology to evaluate the dynamics of the local density of samples (model weights in this case) is very different from conventional statistical analysis.


The proposed methodology looks at a machine learning model in the inverse direction, i.e. it makes it possible to evaluate the change in the overall weight behavior (or regularization from Bayesian viewpoint) of a trained model when $y^*$ location is slightly changed in the space of samples. This is different from evaluating how $y^*$ behaves when a subset of the weights are changed (marginalized over, dropped or fitted with a Gaussian distribution) which is done implicitly in many approximate Bayesian methodologies. An arbitrary choice of weights or inefficient marginalization or distribution fitting, in such methods, can lead to unreliable uncertainty estimates, especially when dealing with larger models/datasets. The functional, $\psi_\mathbf{w}(y^*)$, that evaluates the model predictive PDF in the RKHS, therefore makes it possible to follow an intuition of local fit criteria \cite{leo} for uncertainty quantification very efficiently by deeming a prediction $y^*$ reliable (certain) only if $\delta_\mathbf{w}(y^*) = p(y^* + \Delta{y^*}|\mathbf{w}, \mathbf{x}^*) - p(y^*|\mathbf{w}, \mathbf{x}^*) \approx 0$ where $\Delta{y^*}$ is a small perturbation around $y^*$.\par

Furthermore, the proposed functional operator follows the notion of perturbation theory in physics to quantify the local anisotropy of the model's predictive PDF in the space of multiple orthogonal uncertainty moments or functions and gives a high resolution description of $\delta_\mathbf{w}(y^*)$. The following proposition summarizes the goal of deriving uncertainty functions associated with a neural network model.

\begin{proposition}[Uncertainty representation in the RKHS]
Consider a function $y = f(\mathbf{x}; \mathbf{w})$ that represents a neural network parameterized by weights $\mathbf{w} \in W$ for any input point $\mathbf{x} \in X$ and corresponding output $y \in Y$ and optimized over training samples $(\mathbf{x}_1, y_1), ..., (\mathbf{x}_n, y_n) \in X \times Y$, for a $k$-class classifciation problem with $d$-dimensional input vectors so that $Y = \{1,...,k\}$ and $X \subset \mathbb{R}^d$. Let us consider the predictive PDF associated with its weights to be represented by an RKHS function $\psi_\mathbf{w}(.) \in \mathbf{H}_K$, where $\mathbf{H}_K$ is a uniquely determined and possibly infinite dimensional Hilbert space (constructed from model weights) with associated kernel $K: W \times Y \rightarrow \mathbb{R}$. The goal, using perturbation theory, is to derive an operator from the local gradient flow of $\psi_\mathbf{w}(y)$ which would quantify different moments of the model's uncertainty at its prediction location $y \in Y$. This is depicted in Fig. 1 and captured by the operator $\mathcal{U}$ as follows:

\begin{equation}
    \mathcal{U}(\psi_\mathbf{w}(y)) = [U_1(y), U_2(y), ..., U_m(y)]
\end{equation}

where $U_j(y)$ is the $j^{th}$ uncertainty moment associated with a prediction $y$ and $\mathcal{U}: \mathbb{R} \rightarrow \mathbb{R}^m$ denotes an operator acting on $\psi_\mathbf{w}(y)$ that decomposes it into $m$ orthogonal functions quantifying different degrees of heterogeneity or local anisotropy of $\psi_\mathbf{w}(y)$ around $y$.
\end{proposition}

This framework of uncertainty moment decomposition was first introduced in \cite{stw, stm} as the QIPF (quantum information potential field) and showed promising results in model uncertainty quantification for regression problems \cite{stw} and in time series analysis \cite{stm}. The methodology enjoys the following advantages over Bayesian methods.

\begin{itemize}
\item The QIPF is significantly simpler to compute than the highly iterative Bayesian and ensemble approaches, it is non-intrusive to the training process of the model and enables a \textit{single-shot} estimation of model uncertainty at each test instance.
\item It utilizes the Gaussian RKHS whose mathematical properties, specifically the reproducing property \cite{smola} and the kernel mean embedding theory \cite{emb}, makes it a universal injective function for estimating the data PDF mean without making any underlying distribution assumptions while being efficient in situations involving limited number of samples. This makes it a significantly more reliable quantifier of the effect of model weight PDF than stochastic methods.
\item Its principled physics based moment decomposition formulation (made possible through the functional representation of each point in the data space) is able to provide a multi-scale description of the local PDF dynamics that focuses on the heterogeneous regions of the PDF thereby providing a very accurate description of uncertain regions in the data (or model weight) space.
\end{itemize}

The main contributions of this paper are three-fold:
\begin{itemize}
\item We provide a systematic description of the QIPF framework from the perspective of functional analysis in the RKHS and use perturbation theory to derive the functional operator for uncertainty quantification (UQ). The paper therefore provides a new intuition into how a potential field viewpoint of the model weight space can be used to describe the degree of uncertainty of trained weights in the local neighborhood specified by the output in terms of multiple moments.
\item We specifically evaluate the performance of the framework in an important and unsolved problem of model uncertainty quantification in situations involving \textit{covariate shift} in test data.
\item We analyze the performance of the QIPF against baselines of multiple UQ approaches using diverse models trained on benchmark datasets. We use both accuracy as well as calibration metrics to evaluate performance.
\end{itemize}

The rest of the paper is organized as follows. Related work in the field of uncertainty quantification in machine learning are summarized in section 1. Problem formulation and some technical background are provided in section 2. Proposed approach along with important related definitions are presented in section 3. Section 4 describes perturbation theory and section 5 describes how we use it for the derivation of our formulation of uncertainty quantification. Some implementation examples of the proposed framework are presented in section 6 to illustrate its advantages. Section 7 describes the experimental setup and methods used for comparisons. Experimental results and comparisons are presented in section 8 followed by a brief discussion on computation cost of our framework in section 9 along with the conclusion in section 10.

\section{Related Work}
Existing methods for model uncertainty quantification can be broadly classified into Bayesian \cite{mack, neal, bishop} and non-Bayesian techniques \cite{tib, osb, pearce}. Early Bayesian methods involving inference over model weights include Markov-chain Monte Carlo based Bayesian neural networks \cite{bishop}, Hamiltonian Monte Carlo \cite{neal} and Laplacian approximation \cite{mack}. Although mathematically grounded, their applicability towards large datasets and model architectures becomes restrictive due to huge computational overheads involved. Faster variational inference based methods were developed later on to increase the training speed and efficiency of Bayesian neural networks \cite{graves, jord, hof}. However, such methods still suffer from high dimensionality and high complexity of weight associations making them infeasible for efficient learning of parameter dependencies \cite{projbnn}. Deterministic ReLU networks (which are very popular today) on the other hand suffer from arbitrarily high confidence in predictions outside training PDF \cite{hein2, guo}. Therefore, some recent approaches attempt to approximate the posterior of deterministic networks in a \textit{partially} Bayesian manner. Examples of such methods are stochastic weight averaging (SWA) \cite{wilson1}, construction of a Gaussian distribution over the weights \cite{hein} and recent popular method of fitting a Gaussian distribution over weights by combining SWA solution with diagonal covariance from SGD iterates \cite{wilson2}. \cite{khan} attempts to approximate Bayesian uncertainty of models by leveraging perturbation of weights during SGD. Other partial Bayesian approaches include \cite{gales, light}. Another category of uncertainty estimation methods include post-hoc calibration of networks to improve uncertainty estimation \cite{guo, kul}. However, their performance range outside of training PDF is limited. Non-Bayesian techniques include ensemble based methods that aggregate outputs of multiple differently initialized models to give probabilistic outputs. A significant work in this line is that of \cite{laks}, where authors use deep ensemble neural networks to quantify output uncertainty. This method often produces the best results in many applications. \cite{sha} attempts to simulate ensemble models by integrating over a constructed Gaussian distribution of models. Another work attempts to quantify uncertainty using an ensemble of optimal loss surface paths of a neural network \cite{loss}. A popular work by \cite{gal} suggest to use Monte Carlo Dropout during model testing as an efficient way of extracting output uncertainty information.\par


 \section{Problem Formulation and Background}
 
 \textbf{Setup}: The problem setup is described as follows. Let us assume that $\mathbf{x}_i \in \mathbb{R}^d$ represents $d$-dimensional input features and $y_i \in \{1...k\}$ represents the target labels for $k$-class classification problem where $i$ represents the sample number. The training dataset, denoted as $\mathbf{D}$, consists of $N$ i.i.d samples so that $\mathbf{D} = \{\mathbf{x}_{i}, y_{i}\}_{i=1}^N \sim \mathbf{p}(\mathbf{x},y)$. It is standard to treat a neural network as a probabilistic model trained on $\mathbf{D}$ and parameterized by $\mathbf{w}$ to model the conditional distribution $p(y|\mathbf{x}, \mathbf{w})$ over the prediction $y$ for an input $\mathbf{x}$, where $\mathbf{w}$ is learnt from the training set. More specifically, in this paper, we assume the predictive PDF of the model $p(y|\mathbf{x}, \mathbf{w})$ to define the confidence at $\mathbf{x}$ as the maximum predictive probability that can be written as $\max_{i \in \{1,..,k\}}p(y=i|\mathbf{x},\mathbf{w})$ for a $k$-class classification problem.
 
 \textbf{Covariate Shift}: Let us specifically assume the training input samples to be governed by a distribution $p_{trn}(\mathbf{x})$. The goal is to quantify the predictive uncertainty of the trained neural network model on a test-set whose governing PDF has undergone a shift from $p_{trn}(\mathbf{x})$ to become another unknown distribution $p_{tst}(\mathbf{x})$ in the test-set so that $p_{trn}(\mathbf{x}) \neq p_{tst}(\mathbf{x})$, while the target conditional distributions remain the same, i.e. $p_{trn}(y|\mathbf{x}) = p_{tst}(y|\mathbf{x})$. This common practical phenomenon is also known as \textit{covariate shift} or \textit{sample selection bias} \cite{sug}.
 
 \textbf{Bayesian Formulation}: Epistemic uncertainty in neural networks, especially in regions outside of the training data PDF, depends on the degree of its associated bias-variance trade-off and therefore on how optimally a network’s weights have been regularized for a given set of training samples. For a network that learns without any regularization, the optimization problem corresponds to a maximum log-likelihood estimation problem given by $p(\mathbf{D}|\mathbf{w}) = \Pi_ip(y_i|\mathbf{x}_i, \mathbf{w})$. For regularized training of a network, the optimization corresponds to a maximum a-posteriori estimation problem of the weights given by $p(\mathbf{w|\mathbf{D}}) \propto p(\mathbf{D}|\mathbf{w})p(\mathbf{w})$, where the prior over weights $p(\mathbf{w})$ introduces regularization. Both cases still involve point prediction neural networks. The optimal way of regularizing a network (in the Bayesian sense) is by solving an inference problem given by $p(\mathbf{w}|\mathbf{D}) = \frac{p(\mathbf{D}|\mathbf{w})p(\mathbf{w})}{\int{p(\mathbf{D}|\mathbf{w})p(\mathbf{w})d\mathbf{w}}}$ through a marginalization over $\mathbf{w}$. The posterior predictive distribution corresponding to any test-set data pair $\{ \mathbf{x}^*, y^*\}$ then becomes:
 
 \begin{equation}
     p(y^*|\mathbf{x}^*, \mathbf{D}) = \int{p(y^*|\mathbf{x}^*, \mathbf{w})p(\mathbf{w}|\mathbf{D})d\mathbf{w}}
 \end{equation}
 
 Although such a posterior is a theoretically elegant solution, it is intractable in most modern applications because of the difficulty in computing $p(\mathbf{w}|\mathbf{D})$, due to the difficulty in computing its evidence likelihood. Variational inference and Markov chain Monte Carlo (MCMC) type approaches and their faster variants are typically implemented to obtain approximate solutions. Variational inference posits a variational family of distributions (using a latent variable $v$), $q(\mathbf{w}; v)$ of weights and approximates the posterior $p(\mathbf{w}|\mathbf{D})$ by minimizing the Kullback-Liebler divergence between $q(\mathbf{w}; v)$ and $p(\mathbf{w}|\mathbf{D})$, thus turning the posterior estimation into an optimization problem.\par


 \section{Approach and Definitions}
The proposed approach for model uncertainty quantification relies on measuring the variability of model weight PDF around a particular prediction sample. The intuition here is that a lack of training samples corresponding to the local space around a prediction point will lead to high variability in the predictions within that space because of the sparseness in corresponding weights. The goal is therefore to quantify the \textit{local gradient flow} (or heterogeneity) of $\psi_{\mathbf{w}}(y^*)$ (which represents $p(y^*|\mathbf{x}^* \mathbf{w})$) with respect to $y^*$. As far as we know, there is no other statistical procedure that is capable of estimating heterogeneity in the PDF over the sample space. To this end, the first step is to use an RKHS based function called the \textit{information potential field} \cite{prin} to estimate $\psi_{\mathbf{w}}(y^*)$. It is defined as follows:

\begin{definition}[Information Potential Field]
Given a non-empty set $X = \Omega \subset \mathbb{R}^d$ representing an input random variable and a symmetric non-negative definite kernel function $K: X \times X \rightarrow \mathbb{R}$, the information potential field (IPF) is an estimator of the PDF of $X$ in the RKHS from $n$ samples ($\mathbf{x}_1, \mathbf{x}_2, ..., \mathbf{x}_n$) where $\mathbf{x} \in X$ as

  \begin{equation}
 \psi_{X}(\mathbf{x}) = \frac{1}{n}\sum_{t=1}^{n}K(\mathbf{x}_t, \mathbf{x}).
 \label{ipff}
  \end{equation}
  

\end{definition}

At its core, the IPF is the RKHS equivalent of the Parzen's window method \cite{parz} which is a non-parametric estimator of a continuous density function $f_X(\mathbf{x})$ in an asymptotically unbiased form directly from data (i.e. satisfying the condition that $\lim_{n\to\infty} E[f_X(\mathbf{x}_n)] = f_X(\mathbf{x})$, where $n$ is the number of samples). 



We choose $K$ as the Gaussian kernel in the IPF formulation without loss of generality towards other symmetric non-negative definite functions and henceforth we will denote the kernel as $G(.,.)$. The IPF is therefore an efficient RKHS based asymptotically unbiased estimator of the input data PDF defined empirically by samples. In the RKHS, it is a functional that exists over the projected sample space, in the form of $\psi_X(.)$ which gets estimated at any point $\mathbf{x}$ in the input space through kernel evaluations becoming the scalar $\psi_X(\mathbf{x})$. To simplify the notation, $\psi_X(.)$ will be used herein. The potential field analogy of IPF is simple to explain: If the projected Gaussian function centered at a data sample is associated with a unit mass ``information particle'', then it is easy to prove that the interactions of samples projected in RKHS define a potential field resembling a gravitational field specified by the inner product. So, from a physics point of view the IPF, $\psi_X(.)$, can be thought of as a wave-function created by the density of the projected data samples.\par

An important application of the IPF is in the estimation of R\'enyi's quadratic entropy:

\begin{equation}
    H_2(X) = -\log{\int{f^2_X(\mathbf{x})d\mathbf{x}}}
    \label{ren}
\end{equation}

The argument of the logarithm in (\ref{ren}),  $\Psi_X(X)= \int{f^2_X(\mathbf{x})d\mathbf{x}} = E[f_X(\mathbf{x})]$, is the first moment of the PDF and is called the \textit{information potential} (IP) \cite{prin}. This clearly sows that the IPF is an estimator for $f_X(\mathbf{x})$. This leads to an empirical estimate of the information potential given by:

\begin{equation}
\Psi_X(X) = \bigg\langle\psi_X(.), \psi_X(.)\bigg\rangle  = \frac{1}{n^2}\sum_{j=1}^n\sum_{i=1}^{n}G\big(\mathbf{x}_i, \mathbf{x}_j\big)
\label{tip}
\end{equation}

Hence the IPF plays a crucial role in many information theoretic concepts \cite{prin}. The properties of information potential and its estimate have been studied in \cite{erdog}. A lower bound for the information potential as a statistical estimator is derived in \cite{xu2} and also included in section A (appendix) solidifying its theoretical role in statistics. Another interesting connection of the IPF is the kernel mean embedding theory \cite{embor, emb}, which further proves its robustness as a functional PDF estimator. We delve into this connection in section B (appendix).\par



For our purpose of quantifying $p(y^*|\mathbf{x}^*, \mathbf{w})$, we generalize the IPF constructing it from the model weights and evaluating it at any model prediction output, $y^*$, to quantify the weight PDF at $y^*$. The generalization of (\ref{ipff}) is therefore given by:

\begin{equation}
    \psi_\mathbf{w}(y^*) = \frac{1}{n}\sum_{t=1}^{n}G(\mathbf{w}_t, y^*)
    \label{wipf}
\end{equation}

where $n$ is the number of weights. One can notice here that the data space in which the IPF is constructed (model weights) is different from the data space where it is evaluated (model prediction output space) and hence is an extension to the formulation of IPF (\ref{ipff}) in definition 1. We name this extension as the \textit{cross-information potential field} (CIPF) and define it as follows:

\begin{definition}[Cross-Information Potential Field]
Given two non-empty sets $X$ and $Y$ representing random variables and a symmetric non-negative definite kernel function $G: X \times Y \rightarrow \mathbb{R}$, the cross-information potential field (CIPF) constructed by samples $x \in X$ and evaluated at any point in the space $y \in Y$ is defined as

  \begin{equation}
 \psi_X(y) = \frac{1}{n}\sum_{t=1}^{n}G(x_t, y).
 \label{cipff}
  \end{equation}
  
 where $n$ is the total number of samples in $X$.
\end{definition}

The CIPF estimates the similarity between the PDF of one random variable $X$ with respect to another variable $Y$ locally (at each point) in $X$. The CIPF is therefore also a functional in RKHS which can be computed by simple kernel evaluations in the spaces of the two random variables (provided the same kernel is used to project each random variable). Analogous to the formulation of information potential in (\ref{tip}), the total cross-information potential (CIP) between two random variables (say $X$ and $Y$) is a scalar given by:

\begin{equation}
    \Psi_X(Y) = \bigg\langle\psi_X(.), \psi_Y(.)\bigg\rangle  = \frac{1}{nm}\sum_{j=1}^m\sum_{i=1}^{n}G\big(\mathbf{x}_i, y_j\big)
\label{tcip}
\end{equation}

where $n$ and $m$ are number of samples of $X$ and $Y$ respectively. The cross-information potential therefore measures the total overlap between the PDFs of two random variables. Moreover it resembles Bhattacharya distance and other distance metrics \cite{gok} and also appears in both Euclidean and Cauchy-Schwartz divergence measures based on Renyi's quadratic entropy \cite{prin}. The weight-IPF, $\psi_\mathbf{w}(y^*)$, which is used as an estimation for $p(y^*|\mathbf{x}^*, \mathbf{w})$, therefore measures the local overlap of the two densities (the weight PDF and the prediction space PDF) at the point indicated by the prediction output, $y^*$. It is therefore a value that depends on the relative overlap of the two densities.\par

The proposed methodology calls for a more specific information, the variability of $\psi_\mathbf{w}(.)$ around $y^*$, which unfortunately is not readily quantified in probability theory. A useful operator for this goal is the Laplacian operator which is a spatial high-pass filter and formally measures the flow of the gradient in a potential field \cite{huang}. Our next step is therefore to utilize the Laplacian operator to measure the local gradient flow of $\psi_\mathbf{w}(.)$ across the prediction space. Using a Schr\"odinger's equation (from quantum mathematics) with a Laplacian based wave-function, we show how one can obtain an elegantly normalized and scaled form of $\nabla_y^2\psi_\mathbf{w}(.)$, i.e. local gradient flow of $\psi_\mathbf{w}(.)$. First, the general form of the Schr\"odinger time-independent equation is stated as follows \cite{sch}.


\begin{theorem}[Time-independent Schr\"odinger equation]
Consider a particle within a potential field created by a physical system having an associated wave-function $\psi(.)$, so that the square of $\psi(x)$ gives the probability of the particle being at $x$. The compact representation of the Schr\"odinger time-independent equation describing the system dynamics is given by:

\begin{equation}
    H\psi(.) = E(.)\psi(.)
    \label{shro9}
\end{equation}

where $E(.)$ is a function describing the total energy of the system and $H$ is called the Hamiltonian which is an operator that expresses the \textit{rate of change} of $\psi(.)$ while preserving the total energy, $E(.)$.
\end{theorem}

 Solution of the Schr\"odinger equation has been fundamental to quantify multi-scale uncertainty in quantum physics in terms of multiple local moments. Its solution, through the moment expansion of $\psi(.)$ (and consequently $E(.)$), quantifies the system in terms of multiple moments of Hamiltonian $H$ (or multiple degrees of local rate of change) thereby yielding multiple moments of system uncertainty.\par
 
In the RKHS, the IPF $\psi_\mathbf{w}(.)$ is a wave-function containing the inherent uncertainty of the weight vector solution associated with the limited amount of data and intrinsic random nature of the input data. This makes a Schr\"odinger type formulation appropriate. Furthermore, unlike quantum mechanics that utilizes a Hilbert space, we will be estimating the solution in a RKHS, with the great advantage that we can use the kernel trick to compute the solution in the input space, directly from samples. To this end, we construct (\ref{shro9}) as follows:

\begin{equation}
    H_\mathbf{w}\psi_\mathbf{w}(.) = E_\mathbf{w}(.)\psi_\mathbf{w}(.)
    \label{se}
\end{equation}

where $H_\mathbf{w}$ is a Hamiltonian operator measuring the local rate of change of $\psi_\mathbf{w}(.)$ (weight-IPF) to quantify its uncertainty. We use a Laplacian based Hamiltonian operator to describe the local dynamics (rate of change) of $\psi_\mathbf{w}(.)$ which we call the \textit{quantum information potential field} (QIPF). sections 4 and 5 derive the QIPF expression from (\ref{se}), but we formally define the QIPF expression for a neural network model as follows:

\begin{definition}[Model Quantum Information Potential Field]
Consider a neural network trained on a dataset $\mathbb{D} = \{x_i, y_i\}_{i=1}^N$ with trained weights $\mathbf{w}$ so that its weight-PDF evaluated at any prediction point $y^*$ is given by $\psi_\mathbf{w}(y^*) = \frac{1}{n}\sum_{t=1}^{n}G_\sigma(w_t, y^*)$, where $n$ is the number of weights, $G$ denotes Gaussian kernel with its kernel width $\sigma$. The Schr\"odinger equation associated with $\psi_\mathbf{w}(.)$ is given as

\begin{equation}
    H_\mathbf{w}\psi_\mathbf{w}(.) = \bigg(H_0(.) - \frac{\sigma^2}{2}\nabla_{y}^2\bigg)\psi_\mathbf{w}(.) = E_\mathbf{w}(.)\psi_\mathbf{w}(.)
    \label{sch2}
\end{equation}

where $H_0(.)$ is the potential energy component of the Hamiltonian and $-\frac{\sigma^2}{2}\nabla_y^2$ is the gradient flow operator component of the Hamiltonian consisting of the Laplacian operator $\nabla_y^2$ which acts on $\psi_\mathbf{w}(.)$ with respect to model prediction space $y$. Upon rearranging the middle and last terms, we obtain a normalized expression for the potential energy function at any prediction ouput $y^*$ given by

\begin{equation}
    H_0(y^*) = E_\mathbf{w}(y^*) + (\sigma^2/2)\frac{\nabla_y^2\psi_\mathbf{w}(y^*)}{\psi_\mathbf{w}(y^*)}
\end{equation}

which is called the quantum information potential field (QIPF) of weights. Here, $E_\mathbf{w}(.) = -min(\sigma^2/2)\frac{\nabla_y^2\psi_\mathbf{w}(.)}{\psi_\mathbf{w}(.)}$ and acts as a lower bound to ensure $H_0(.)$ is always positive.

\end{definition}
The QIPF is a therefore a functional operator in the RKHS that takes the normalized Laplacian of the IPF to describe the local dynamics (rate of change) of $\psi_\mathbf{w}(.)$. It has a small value in parts of the RKHS that have high density of samples and increases its value in parts of the space that are sparse in projected data. The QIPF operator is the RKHS version of the energy based learning methods proposed by LeCun \cite{lecun2006}, which  avoids the problems associated with estimating the normalization constant in probabilistic models. A close analysis of the mathematical form of the QIPF operator shows its similarity with the integrand of Fisher's information \cite{flego}. For uncertainty quantification, we are much more interested in the low density regions of the PDF. Hence, just like in quantum mechanics, a local decomposition of the QIPF will be very beneficial for uncertainty quantification. Our final step therefore is to solve (\ref{sch2}) in a way that induces orthogonal moment expansion of the potential function $H_0(.)$, given by $H_0^k(.)$, corresponding to moments of the wave-function $\psi_\mathbf{w}^k(.)$ and total energy at each moment $E_\mathbf{w}^k(.)$ where $k = 0, 1, 2, ...$ is the moment number. $H_0^k(.)$, $\psi_\mathbf{w}^k(.)$ and $E_\mathbf{w}^k(.)$ preserve the same meaning as $H_0(.)$, $\psi_\mathbf{w}(.)$ and $E_\mathbf{w}(.)$ respectively, but for different corresponding moments as denoted by $k$, each of which specializes in a different part of weight PDF potential field. Formally, this decomposition of $H_\mathbf{w}$ is defined as follows.


\begin{definition}[Model QIPF Decomposition]
Consider a neural network trained on a dataset $\mathbb{D} = \{x_i, y_i\}_{i=1}^N$ and characterized with trained weights $\mathbf{w}$ so that its weight PDF is represented by a Gaussian RKHS embedding called the information potential field given by $\psi_\mathbf{w}^0(y^*) = \frac{1}{n}\sum_{t=1}^{n}G_\sigma(w_t, y^*)$, where $n$ is the number of weights, $G$ denotes Gaussian kernel with its kernel width $\sigma$. The model's uncertainty associated with a particular prediction $y^*$ corresponding to a test-input sample $\mathbf{x}^*$ can be described by an ordered set of $m$ decomposed weight-QIPF moments $\{H_0^1(y^*), H_0^2(y^*), ..., H_0^m(y^*)\}$, the $k^{th}$ moment of which is given by the expression:

\begin{equation}
    H_0^k(y^*) = E_\mathbf{w}^k(y^*) + (\sigma^2/2)\frac{\nabla_y^2\psi_\mathbf{w}^k(y^*)}{\psi_\mathbf{w}^k(y^*)}
    \label{final}
\end{equation}
where, \\
$\psi_\mathbf{w}^k(.)$: $k^{th}$ order Hermite function (normalized) projection of the information potential field $\psi_\mathbf{w}(.)$. \\
$\nabla_y^2$: Laplacian operator acting with respect to model prediction space $y$.\\
$E_\mathbf{w}^k(.) = -min(\sigma^2/2)\frac{\nabla_y^2\psi_\mathbf{w}^k(.)}{\psi_\mathbf{w}^k(.)}$: lower bound to ensure each moment is positive.
\end{definition}

$H_0^0(y^*)$, $H_0^1(y^*)$, $H_0^2(y^*)$ ... are thus the successive local moments of model uncertainty (called QIPF moments) that induce anisotropy in the predictive PDF space and represent functional measurements at $y^*$ corresponding to \textit{different degrees of heterogeneity of model PDF} at $y^*$. Definitions 1-4 therefore fully describe the proposed uncertainty decomposition framework for neural network models.\par

\begin{figure}[!t]
    \centering\includegraphics[scale = 0.20]{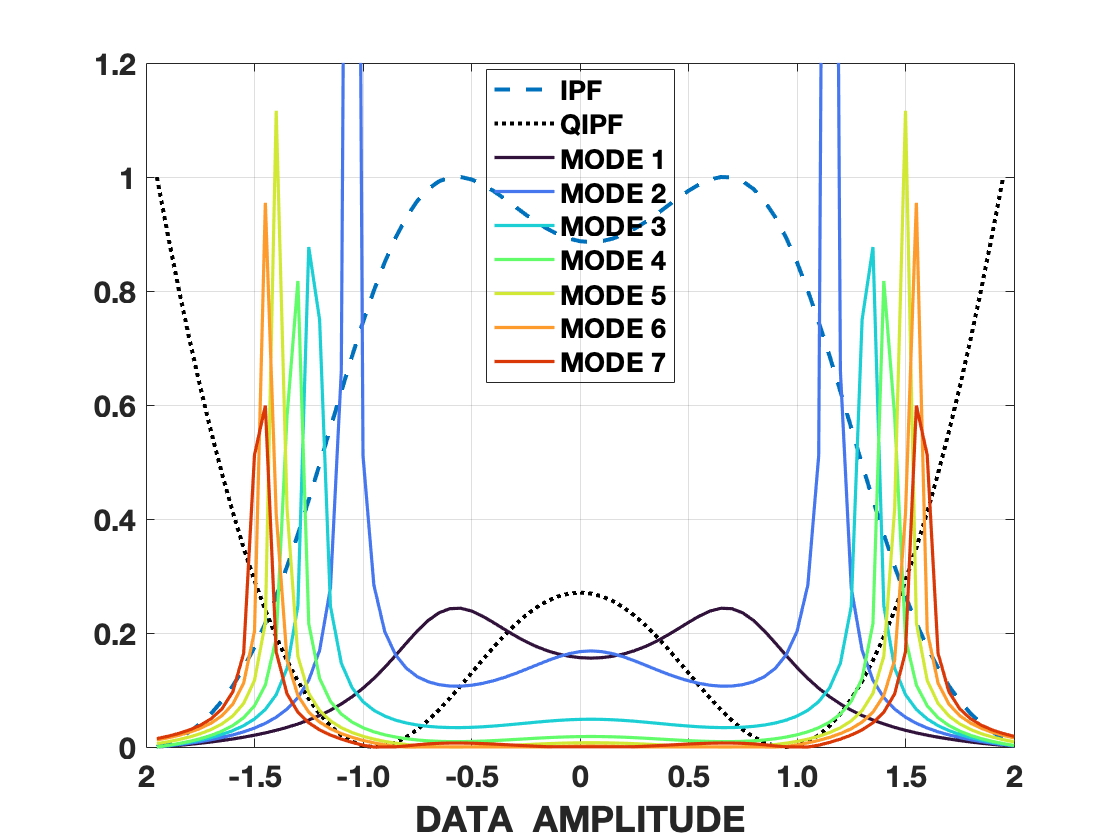}
    \caption{Sine-wave QIPF: Kernel width = 0.5}
    \label{sp1}
\end{figure}

As a pedagogical example, we demonstrate the QIPF decomposition of a simple sine-wave signal in Fig. \ref{sp1} to show how the different moments of the QIPF behave in the space of the IPF of the signal. The IPF (dashed blue line) conforms to the PDF of a sinusoidal signal (the peaks correspond to the positive and negative peaks of the sine-wave, where the rate of change is the slowest, so there are more samples). One thing to note is that the signal amplitude is between -1 and 1, but the IPF starts being positive outside the signal range because of the Gaussian kernel. The QIPF (dotted black line) is the mirror image of the IPF, i.e. it is small when the density of samples is high, and increases when the density of samples decreases. Specifically we see a very large increase when the IPF goes to zero. The Hermite modes are automatically located in the regions where the density of samples varies. Since the Hermite polynomials are orthogonal, the moments always work with the residue between the QIPF and previous modes. Hence they are located across the range of the random variable with a large number automatically placed on the tails of the PDF. The location of the modes depends on the kernel size as discussed in \cite{stw}.\par

The following sections (4 and 5) provide the derivation and intuition behind the QIPF expression and its decomposition which were stated in definitions 3 and 4 respectively.

 \section{Perturbation Theory}
Perturbation theory is well known in quantum mechanical applications. It involves approximation methods to describe a complicated system as a perturbed version of a similar but simpler system whose mathematical solution is well known \cite{qtlan}. In doing so, the various physical quantities associated with the complicated system (energy levels and eigenstates) get expressed as \textit{corrections} to the known quantities of the unperturbed system. Another way of interpreting the theory could be that it is a method to extract more intrinsic dynamical information of a system by observing its behavior upon perturbing it. We formally define the time-independent perturbation theory \cite{qtlan2} as follows.
 
  \begin{theorem}[Rayleigh-Schr\"{o}dinger perturbation theory]
  Suppose that the goal is to find approximate solutions of a time-independent system with Hamiltonian $H$ for which it is difficult to find exact solutions (wavefunction $\psi(.)$ and energy $E(.)$):
  
    \begin{equation}
      H\psi(.) = E(.)\psi(.)
  \end{equation}
  
  Let us assume that the exact solutions of $\psi^{0}(.)$ and $E^0(.)$ of a simpler system with Hamiltonian $H_0$ are known and arise from the time-independent Schr\"odinger equation as follows:
  
  \begin{equation}
      H_0\psi^{0}(.) = E^{0}(.)\psi^{0}(.)
  \end{equation}

Here the superscript 0 denotes the quantities to be associated with an unperturbed system.

Let $H_p$ be a Hamiltonian representing a weak physical disturbance and let $\lambda$ be a dimensionless parameter ranging between 0 (no perturbation) and 1 (full perturbation) so that the difference between $H$ and $H_0$ is merely seen as perturbation on $H_0$ by $\lambda H_p$. This leads to an expression of $H$ as a perturbed Hamiltonian given by $H = H_0 + \lambda H_p$. One therefore requires to find approximate solutions to the following Schr\"odinger's equation of the perturbed Hamiltonian (in order to solve $H$):

\begin{equation}
   H\psi(.) = (H_0 + \lambda H_p)\psi(.)  = E(.)\psi(.)
\end{equation}

Assuming $\lambda$ to be sufficiently weak, one can express $E$ and $\psi$ in terms of a power series starting from the unperturbed quantities leading to the following approximate solutions: 
\begin{equation}
\begin{split}
E(.) = E^0(.) + \lambda E^1(.) + \lambda^2 E^2(.) + ...  \\
\psi(.) = \psi^0(.) + \lambda\psi^{1}(.) + \lambda^2\psi^{2}(.) + ...
\end{split}
\label{exp}
\end{equation}
where $E^k(.)$ and $\psi^k(.)$ are $k^{th}$ order corrections to energy and wavefunction respectively.



 \end{theorem}
 
 \section{Deriving Model Uncertainty: Perturbation in Weight Potential Field}
 The proposed approach uses perturbation theory on the RKHS field representation of the model weights to systematically extract its uncertainty information in the form of successive moments. This approach can also be interpreted as perturbing the model weight potential field (or PDF) to quantify its variability in the local vicinity of its prediction in terms of successive orders of corrections (moments) induced to the original weight PDF. The idea here is that weight PDF would vary more around uncertain predictions and hence uncertainty would be characterized with more \textit{corrections} to the original (unperturbed) weight field. To this end, the model becomes an uncertain system of its weight vector and its ideal description can be provided by the following Schr\"odinger's equation:

\begin{equation}
    H_\mathbf{w}\psi_\mathbf{w}(.) = E_\mathbf{w}(.)\psi_\mathbf{w}(.)
    \label{h}
\end{equation}

where $H_\mathbf{w}$ is a Hamiltonian operator measuring the local variability of wave-function associated with the weights $\psi_\mathbf{w}(.)$. Since the quantities $H_\mathbf{w}$, $\psi_\mathbf{w}(.)$ and $E_\mathbf{w}(.)$ are unknown, let us solve them using perturbation theory through the following three steps.




 \textbf{Base System of Model Weights}: Following the intuition of perturbation theory, we begin with a simpler system of weights (assumed to be close to (\ref{h})) described by a base-system Hamiltonian $H_0(.)$ (which is normally taken as a potential energy function) having a known solution, $\psi_\mathbf{w}^0(.)$, which for our case is the information potential field of weights (\ref{wipf}). Its evaluation at any prediction $y^*$ is given by:

\begin{equation}
    \psi_\mathbf{w}^0(y^*) = \frac{1}{n}\sum_{t=1}^{n}G_\sigma(w_t, y^*)
    \label{vs}
\end{equation}

Here, $G$ is the Gaussian kernel, $\sigma$ is the kernel width and $w_t$ denoted $t^{th}$ model weight (as described in definitions 3 and 4). The Schr\"odinger's equation corresponding to the base-system Hamiltonian $H_0(.)$ with the known solution $\psi_\mathbf{w}^0(.)$ becomes:


\begin{equation}
    H_0(.)\psi_\mathbf{w}^0(.) = E_\mathbf{w}^0(.)\psi_\mathbf{w}^0(.)
    \label{h0}
\end{equation}


 \begin{figure*}[!t]
  \begin{subfigure}{0.3\linewidth}
    \centering\includegraphics[scale = 0.25]{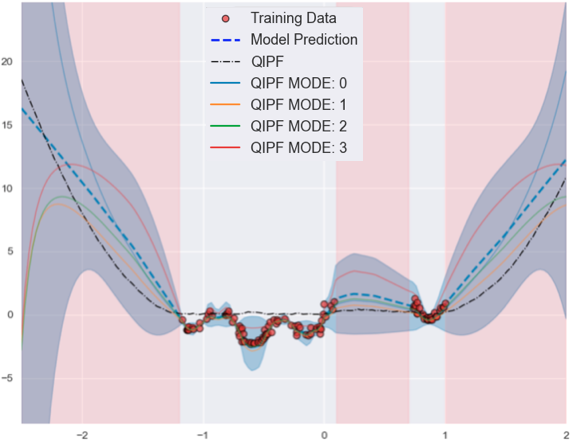}
    \caption{QIPF}
  \end{subfigure}
     \begin{subfigure}{0.22\linewidth}
    \centering\includegraphics[scale = 0.19]{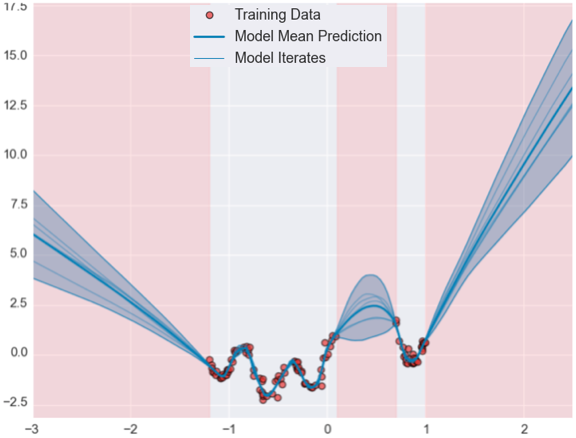}
    \caption{SGD}
  \end{subfigure}
  \begin{subfigure}{0.22\linewidth}
    \centering\includegraphics[scale = 0.19]{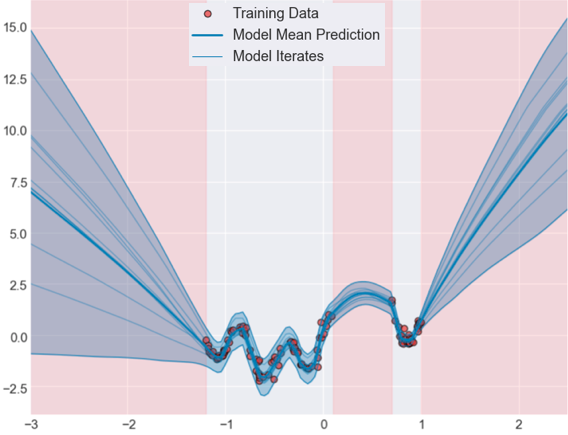}
    \caption{SWAG}
  \end{subfigure}
  \begin{subfigure}{0.22\linewidth}
    \centering\includegraphics[scale = 0.19]{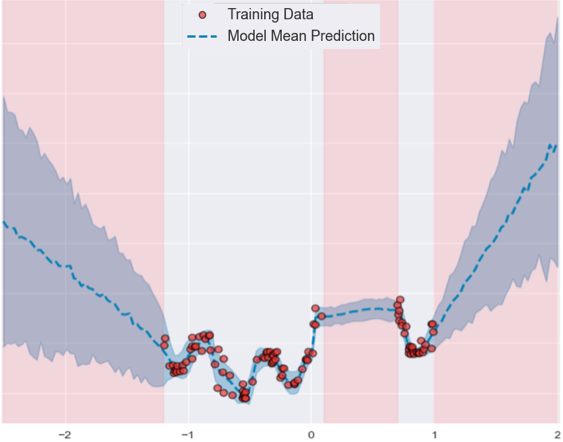}
    \caption{MC Dropout}
  \end{subfigure}
  
  \begin{subfigure}{0.24\linewidth}
    \centering\includegraphics[scale = 0.2]{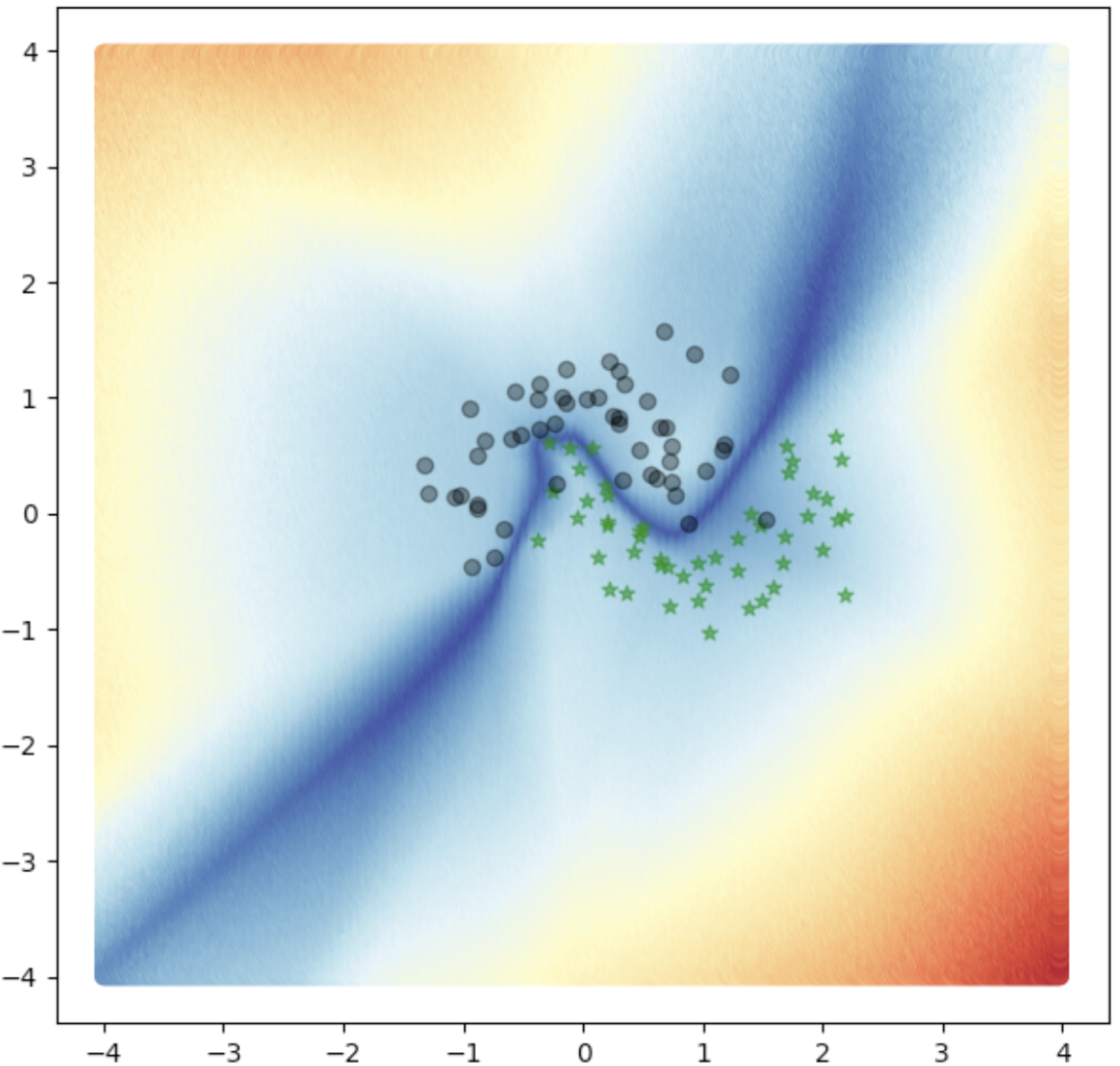}
    \caption{MC Dropout}
  \end{subfigure}
    \begin{subfigure}{0.24\linewidth}
    \centering\includegraphics[scale = 0.2]{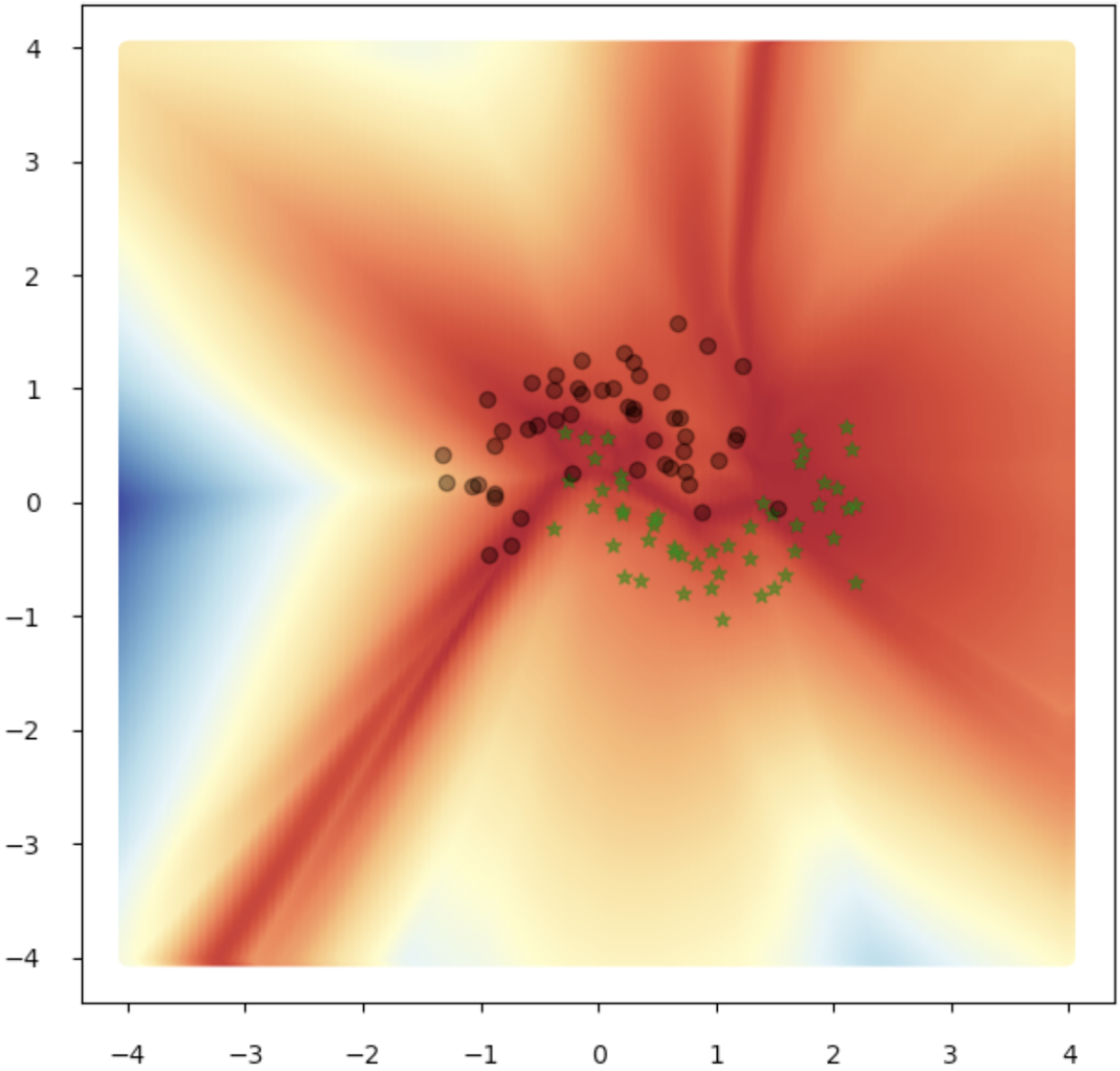}
    \caption{SGD}
  \end{subfigure}
    \begin{subfigure}{0.24\linewidth}
    \centering\includegraphics[scale = 0.2]{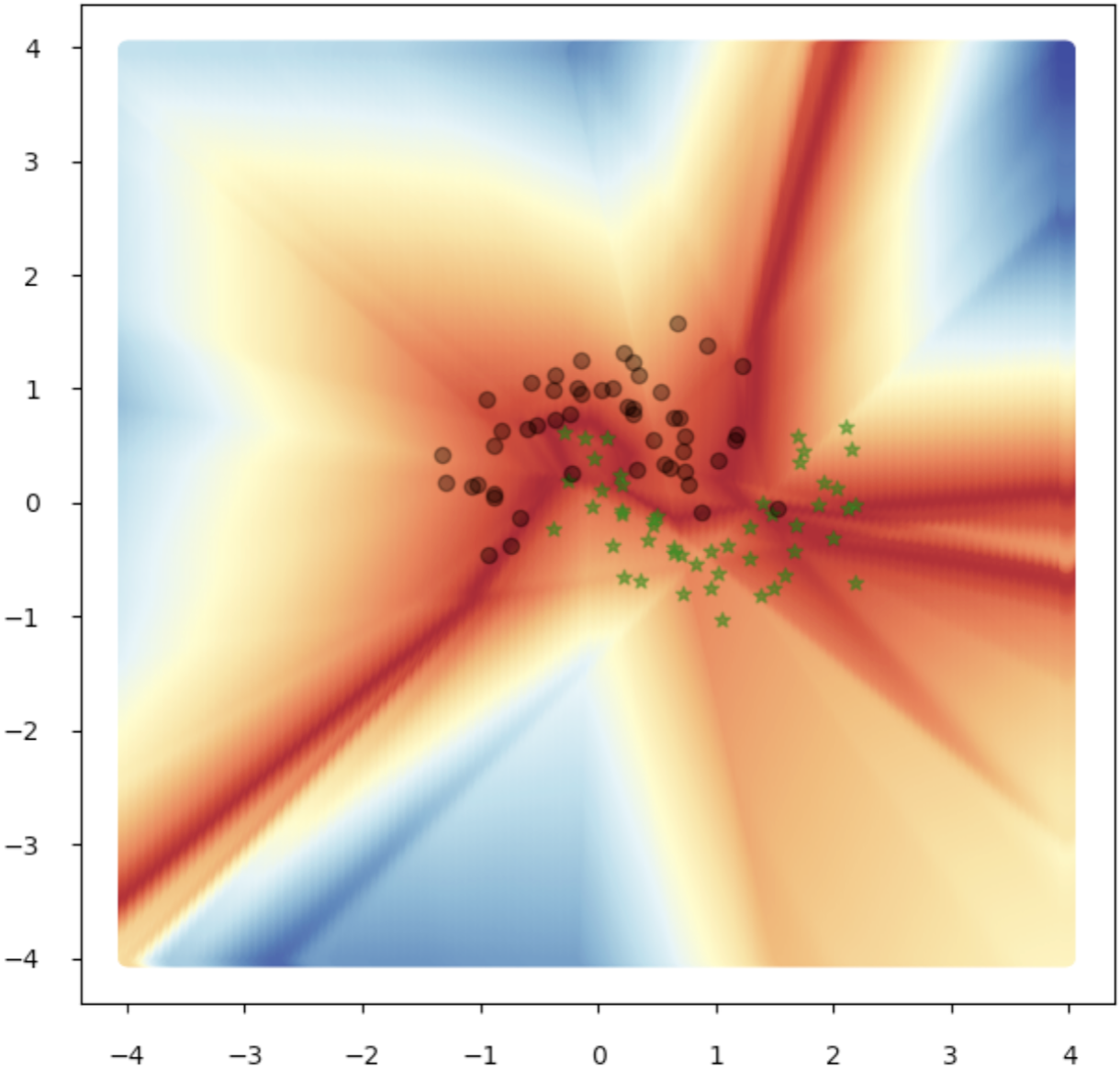}
    \caption{SWAG}
  \end{subfigure}
    \begin{subfigure}{0.24\linewidth}
    \centering\includegraphics[scale = 0.2]{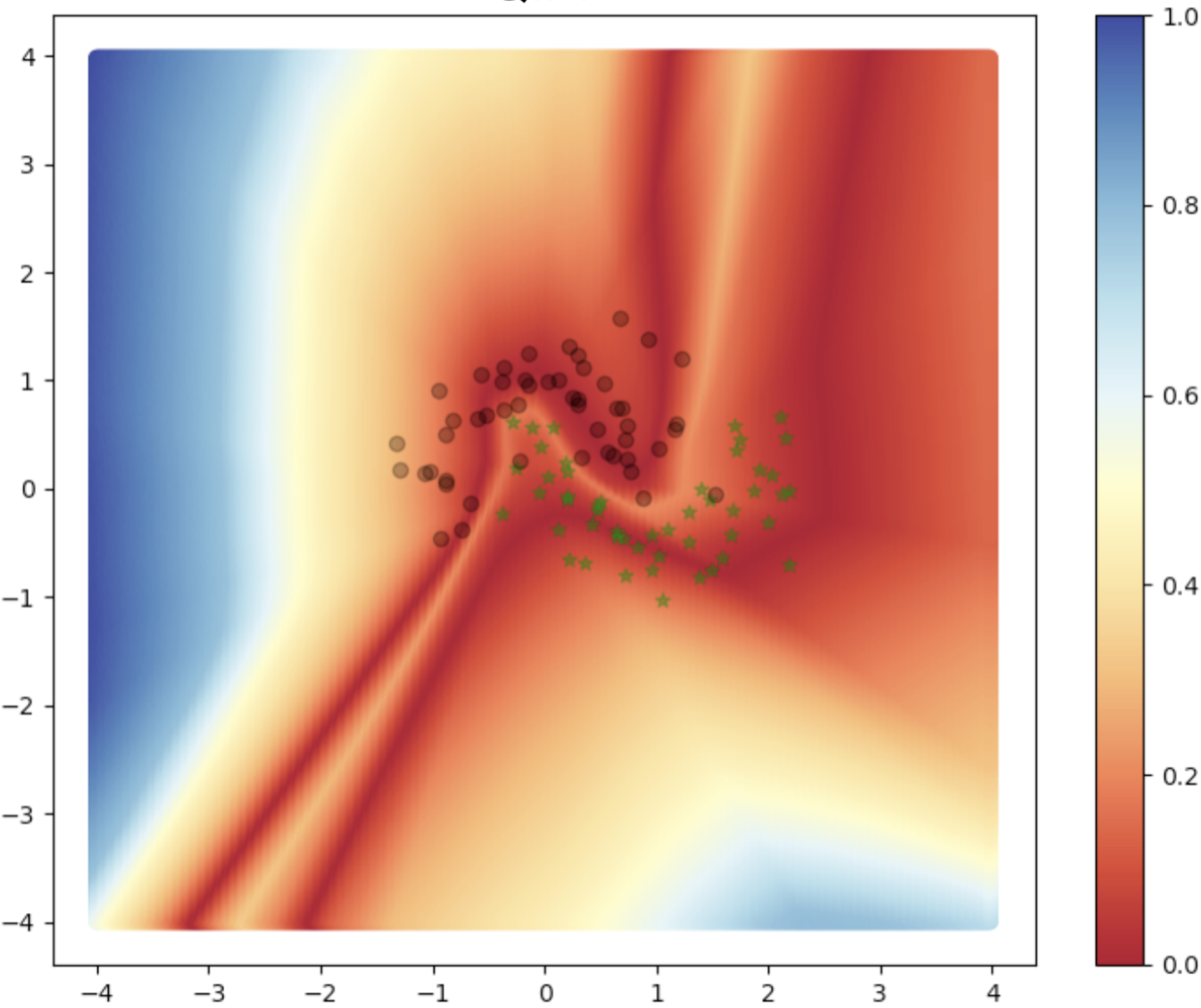}
    \caption{QIPF}
  \end{subfigure}
    \caption{(a-d): Model uncertainty quantified by different methods for a regression problem (white regions: seen data regions, pink regions: unseen data regions). QIPF uncertainty can be observed to be more sensitive in discriminating between seen and unseen regions, while also being able to quantify sample uncertainty within the training region. (e-h): Model uncertainty quantified by different methods for a classification problem (noise added two-moons dataset). QIPF uncertainty can be seen to be significantly more precise showing low uncertainty mostly in regions where training samples are present and high uncertainties at the decision boundary and outside of the data regions.}
  \label{sp}
  \end{figure*}

 \textbf{Perturbing the Base Hamiltonian}: Following perturbation theory to obtain $H_\mathbf{w}$ from $H_0(.)$, a perturbation Hamiltonian is added to $H_0(.)$ that shifts its evaluation from $y^*$ to $y^* + \Delta y^*$. This step implicitly determines how regularized the system of weights is around $y^*$. A big change in system dynamics due to the perturbation would mean the weights are not regularized around $y^*$ are hence the model uncertainty at $y^*$ is high. Therefore a good choice for the perturbation Hamiltonian $H_p$ is a Laplacian operator (which is a local gradient based operator) that operates on $y^*$ and is denoted as $\nabla_{y^*}^2$. The intensity of the perturbation controls the size of neighborhood around $y^*$ where the analysis is being made and hence is directly dependent on the kernel width selected in the IPF formulation $\psi_\mathbf{w}^0(.)$ for defining the weight PDF function in the unperturbed Hamiltonian $H_0(.)$. Therefore the perturbation Hamiltonian becomes $H_p = -\frac{\sigma^2}{2}\nabla_{y^*}^2$ so that $H_\mathbf{w} = H_0(.) - \frac{\sigma^2}{2}\nabla_{y^*}^2$. This is the same as the part of the Hamiltonian referred to as the gradient flow operator in the QIPF formulation in \cite{stw}. Therefore the Schr\"odinger equation corresponding to the desired system Hamiltonian can be expressed as:

\begin{equation}
    H_\mathbf{w}\psi_\mathbf{w}(.) = \bigg(H_0(.) - \frac{\sigma^2}{2}\nabla_{y^*}^2\bigg)\psi_\mathbf{w}(.) = E_\mathbf{w}(.)\psi_\mathbf{w}(.)
    \label{s}
\end{equation}

Rearranging the terms of (\ref{s}), we obtain the following:
\begin{equation}
   H_0(.) = E_\mathbf{w}(.) + (\sigma^2/2)\frac{\nabla_y^2\psi_\mathbf{w}(.)}{\psi_\mathbf{w}(.)}
    \label{s2}
\end{equation}
Here $E_\mathbf{w}(.)$ is considered to be a normalizing constant (lower bound) given by $E_\mathbf{w}(.) = -min(\sigma^2/2)\frac{\nabla_y^2\psi_\mathbf{w}(.)}{\psi_\mathbf{w}(.)}$ so that $H_0(.)$ is always positive. $H_0(.)$ is called the quantum information potential field (QIPF) and is the same expression as described in definition 3.

 \textbf{Moment Decomposition of the QIPF}: The final step involves expressing $\psi_\mathbf{w}(.)$ in (\ref{s}) as a power series expansion starting with the base-system (unperturbed) Hamiltonian solution $\psi_\mathbf{w}^0(.)$ (similar to (\ref{exp}) in perturbation theory) as follows:

\begin{equation}
    \psi_\mathbf{w}(.) = \psi_\mathbf{w}^0(.) + \lambda\psi_\mathbf{w}^{1}(.) + \lambda^2\psi_\mathbf{w}^{2}(.) + ...
    \label{mom}
\end{equation}

We take the perturbation intensities associated with $\lambda$ as simply the kernel width term $\frac{\sigma^2}{2}$. Therefore, $ \psi_\mathbf{w}^{1}(.), \psi_\mathbf{w}^{2}(.) ...$ are the successive orders (moments) of corrections to $\psi_\mathbf{w}^0(.)$ in order to obtain $\psi_\mathbf{w}(.)$. The moments are estimated by projecting $\psi_\mathbf{w}^0(.)$ to the successive orders of normalized Hermite functions. Substituting (\ref{s2}) in (\ref{mom}) then leads to a moment expansion of the QIPF, $H_0(.)$ as follows:

\begin{equation}
\begin{split}
H_0^k(.) = E_\mathbf{w}^k(.) + (\sigma^2/2)\frac{\nabla_y^2\mathscr{P}_k\big(\psi_\mathbf{w}^0(.)\big)}{\mathscr{P}_k\big(\psi_\mathbf{w}^0(.)\big)} \\
   = E_\mathbf{w}^k(.) + (\sigma^2/2)\frac{\nabla_y^2\psi_\mathbf{w}^k(.)}{\psi_\mathbf{w}^k(.)}
    \label{sh}
\end{split}
\end{equation}

Here, $\mathscr{P}_k\big(\psi_\mathbf{w}(.)\big) = \psi_\mathbf{w}^k(.)$ is the projection of $\psi_\mathbf{w}(.)$ in the $k^{th}$ order Hermite function to obtain an approximation of $\psi_\mathbf{w}^k(.)$. Instead of doing a moment expansion of $E_\mathbf{w}(.)$ here, for practical considerations, we simply let it be a lower bound corresponding to each wave-function moment given by $E_\mathbf{w}^k(.) = -min(\sigma^2/2)\frac{\nabla_y^2\psi_\mathbf{w}^k(.)}{\psi_\mathbf{w}^k(.)}$ to enforce the condition that the corresponding $H_0^k(.)$ is always positive. Note that (\ref{sh}) is the expression for moment decomposition of weight-QIPF as described in definition 4. Here, the computation of the desired system Hamiltonian $H_\mathbf{w}$ induces successive corrections of the base Hamiltonian $H_0(.)$ given by (\ref{sh}) so that  $H_\mathbf{w} = \sum_{k=0}^{k=\infty}H_0^k(.)$. These corrections quantify successive moments of model uncertainty.\par


We posit that the moment expanded expression series in (\ref{sh}) will be naturally decaying, i.e. the amplitudes of higher order moments will successively decrease because of the low values of IPF (wave-function) in regions of large heterogeneity. Since more uncertain model predictions will correspond to more heterogeneous regions of the model weight PDF, the successive order moments of QIPF represented by $H_0^k(y^*)$ provide different resolutions of certainty of the model in its prediction at $y^*$.\par

The pseudo-code of QIPF implementation is presented in algorithm 1. As can be observed, the implementation is fairly simple involving only three main steps:
\begin{itemize}
    \item Evaluation of the model weight PDF in the Gaussian RKHS given by the cross-information potential field which we call the weight-IPF:     $\psi_\mathbf{w}^0(y^*)$, given by (\ref{vs}).
    \item Projection of the weight-IPF into successive orders of normalized Hermite function spaces: $\mathscr{P}_k(\psi_\mathbf{w}^0)= \psi_\mathbf{w}^k$.
    \item Evaluation of the decomposed uncertainty moments of the QIPF: $H_0^k(y)$, given by (\ref{final}).
\end{itemize}



  \section{Illustrative Examples}
  

 The difference in the properties of various uncertainty quantification approaches are illustrated in the case of regression (Fig. \ref{sp}(a-d)). An over-parameterized and fully connected multi-layer perceptron consisting of 3 hidden layers with 100 neurons in each layer is trained on just 120 data pairs to learn a weighted sine-wave function with added noise. The training is carried out using training samples that were generated only in the white regions, with the training samples shown as red balls. Pink regions represent regions with no training data. The model was tested in the entire data region [-2, 2.5] and different UQ methods were implemented. The QIPF implementation extracts 4 modes (after fixing the kernel size as 100 times the Silverman bandwidth) using all the weights. For Monte Carlo (MC) dropout method, we added a dropout layer after each hidden layer with a small dropout rate of 0.1 during training and quantified uncertainty using 100 iterations at each test sample. We implement SGD iterates \cite{sgd} as a baseline as well as the Stochastic Weight Averaging - Gaussian (SWAG) method \cite{wilson2}, which is a recent popular baseline in the class of approximate Bayesian methods. We used publicly available code \footnote{\url{https://github.com/wjmaddox/swa_gaussian}} for their implementations. We took three times the standard deviation of the modes (in the case of QIPF) and of the test iterations/samples (of other methods) as the corresponding uncertainty ranges. The weight-QIPF modes were centered with the prediction for easier visualization. As can be seen from Fig. \ref{sp}a, the QIPF uncertainty immediately increases outside of the regions with samples (white), with the higher order moments successively being the more sensitive (as is expected). The QIPF uncertainty can be seen to discriminate better (in terms of uncertainty) the seen and unseen regions of the data than other methods, while also being sensitive towards uncertainties within the training region. MC dropout and SWAG extrapolate well at the boundaries outside of the seen regions but fail to effectively capture unseen region in the middle. SGD is able to capture that region well but fails to extrapolate outside effectively and also fails to quantify uncertainty within the training regions.\par
 
\begin{algorithm}[!t]
\caption{Model Weight-QIPF Decomposition}\label{euclid}
\begin{algorithmic}
\State \textbf{Input:} $w$: Model weights, $\sigma$: Kernel width.
\State $m$: Number of quantum modes.
\State $y^*$: point in pre-softmax prediction space being evaluated.
\State \textbf{Initialization:}
\State $\psi_\mathbf{w}$: Weight-IPF (Wave-function), G: Gaussian kernel.
\State $\psi^1, \psi^2, ... , \psi^m$: Wave-function Hermite Function Projections.
\State $H_0^1, H_0^2, ... , H_0^m$: weight-QIPF moments.
\State $E^1, E^2, ... , E^m$: Lower bound associated with each mode.
\State \textbf{Computations:}
\State $\psi_\mathbf{w}=0$
\For {$i = $ 1 to length(w)}
\State $\psi_\mathbf{w} \gets \psi_\mathbf{w} + G(w_i, y^*)$ 
\EndFor
\State $\psi_\mathbf{w} \gets \sqrt{mean(\psi_\mathbf{w})}$
\State 
\State $[\psi_\mathbf{w}^1, \psi_\mathbf{w}^2, ... , \psi_\mathbf{w}^m] \gets Hermite  Projections(\psi_\mathbf{w})$
\State $[{\nabla_y^2}\psi_\mathbf{w}^1, ... , {\nabla_y^2}\psi_\mathbf{w}^m] \gets Laplacians$ 
\State 
\For {each mode $k$}
\State $E_\mathbf{w}^k = -\min_{y_1,y_2,...,y_q}\frac{\sigma^2/2{\nabla_y^2}\psi_\mathbf{w}^k(y_1,y_2,...,y_q)}{\psi_\mathbf{w}^k(y_1,y_2,...,y_q)}$
\State
\State $H_{0}^k(y^*) = E_\mathbf{w}^k(y^*) + \frac{\sigma^2/2{\nabla_y^2}\psi_\mathbf{w}^k(y^*)}{\psi_\mathbf{w}^k(y^*)}$
\EndFor
\end{algorithmic}
\end{algorithm}

 In another example (Fig. \ref{sp}(e-h)), we compare the different methods in a classification problem involving a noise added two-moons dataset. Here we also used a fully connected multi-layer perceptron consisting of 3 hidden layers but only with 20 neurons in each hidden layer. Training was carried out using only the samples (shown in the figure) and evaluated over a larger area during testing. We observe many interesting characteristics of the different methods. MC dropout can be seen to be quite effective in quantifying the uncertainty at the decision boundary, which also extrapolates well. However, the method shows the model to be increasing in its confidence (decreasing uncertainty) as one moves away from the training regions, which is unrealistic. SGD and SWAG methods seem to fix this problem and show more realistic (increasing) uncertainty levels outside of the training regions. However, they are not effective in capturing the uncertainty at the decision boundary. QIPF (Fig. \ref{sp}(h)), on the other hand, shows very desirable uncertainty characteristics. It shows the model to be confident only in the tightly bounded regions where samples are present. It shows sharp uncertainty at the decision boundary which extrapolates in an elegant way similar to MC dropout, while also showing increasing uncertainty of the model as one moves away from the training regions.\par
  
  We also provide additional analysis of how QIPF decomposition behaves with regularization of the model in section C (appendix).


    \begin{figure*}[!t]
  \begin{subfigure}{0.48\linewidth}
    \centering\includegraphics[scale = 0.25]{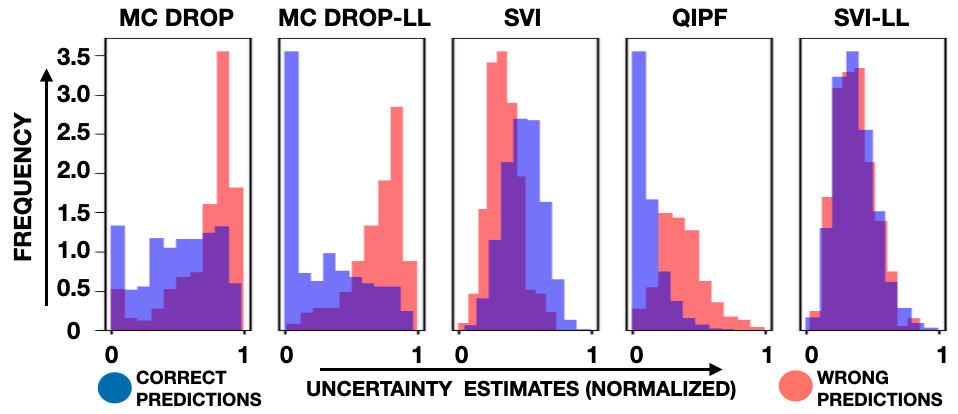}
    \caption{Histogram of Uncertainty Scores}
  \end{subfigure}
  \begin{subfigure}{0.24\linewidth}
    \centering\includegraphics[scale = 0.3]{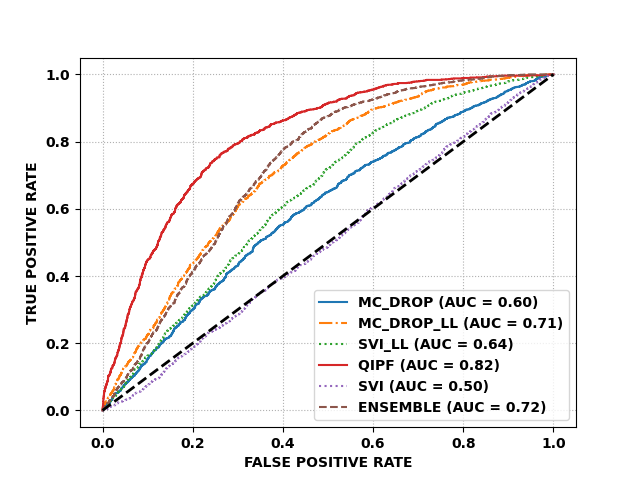}
    \caption{ROC Curves}
  \end{subfigure}
  \begin{subfigure}{0.24\linewidth}
    \centering\includegraphics[scale = 0.3]{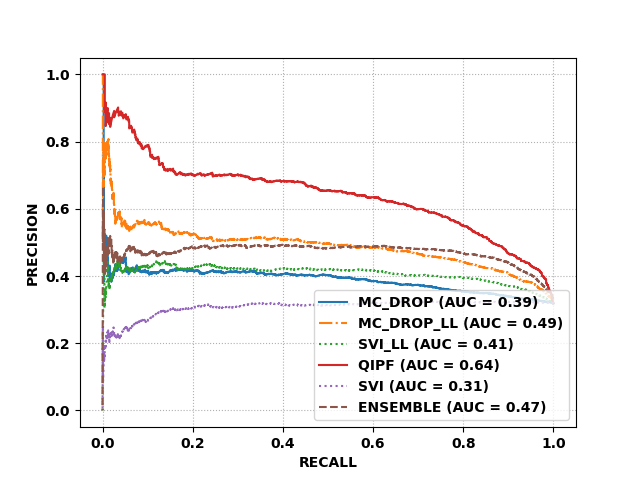}
    \caption{Precision-recall curves}
  \end{subfigure}
    \begin{subfigure}{0.24\linewidth}
    \centering\includegraphics[scale = 0.19]{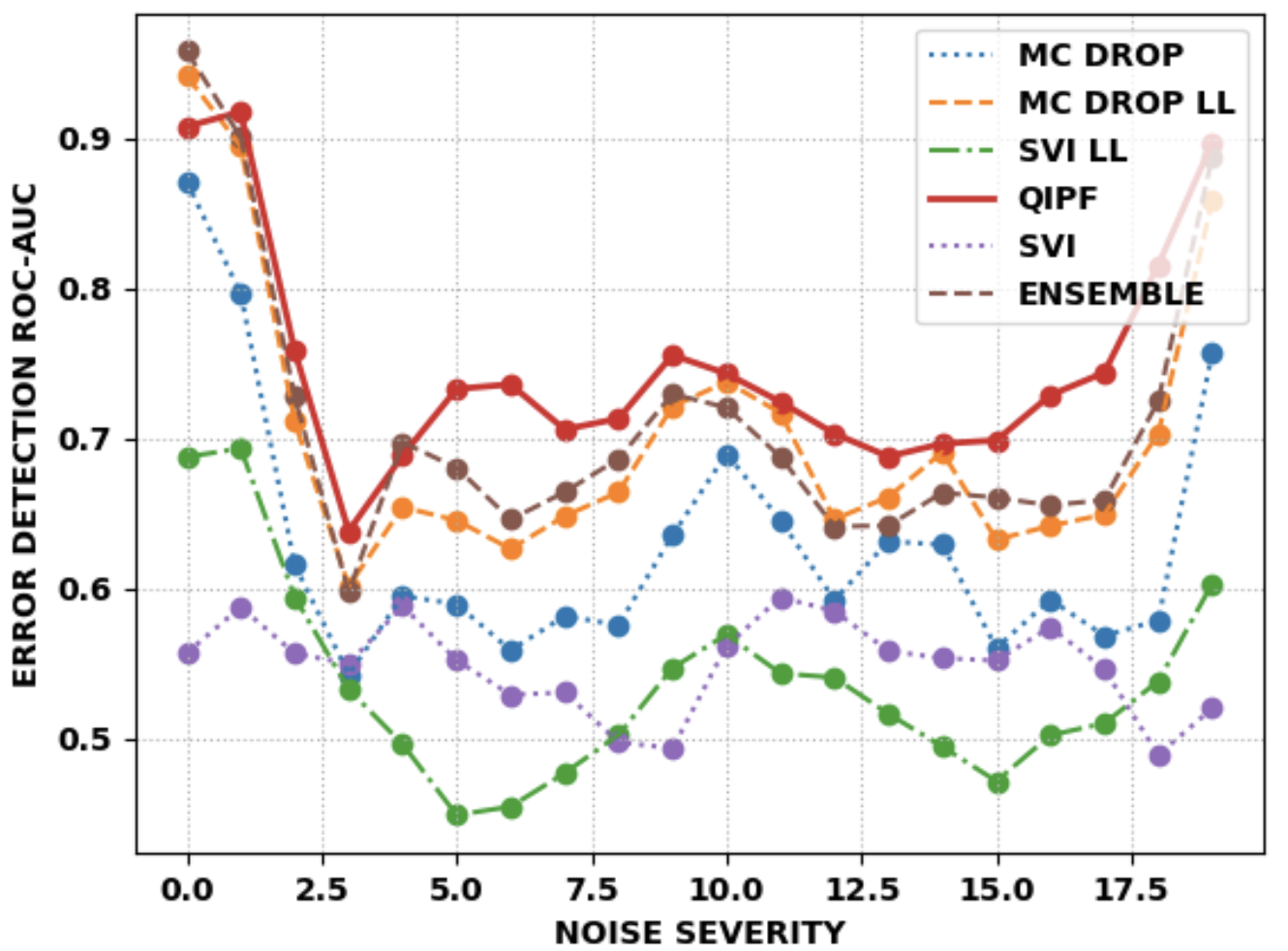}
    \caption\centering{ROC-AUC vs noise severity}
  \end{subfigure}
    \begin{subfigure}{0.24\linewidth}
    \centering\includegraphics[scale = 0.19]{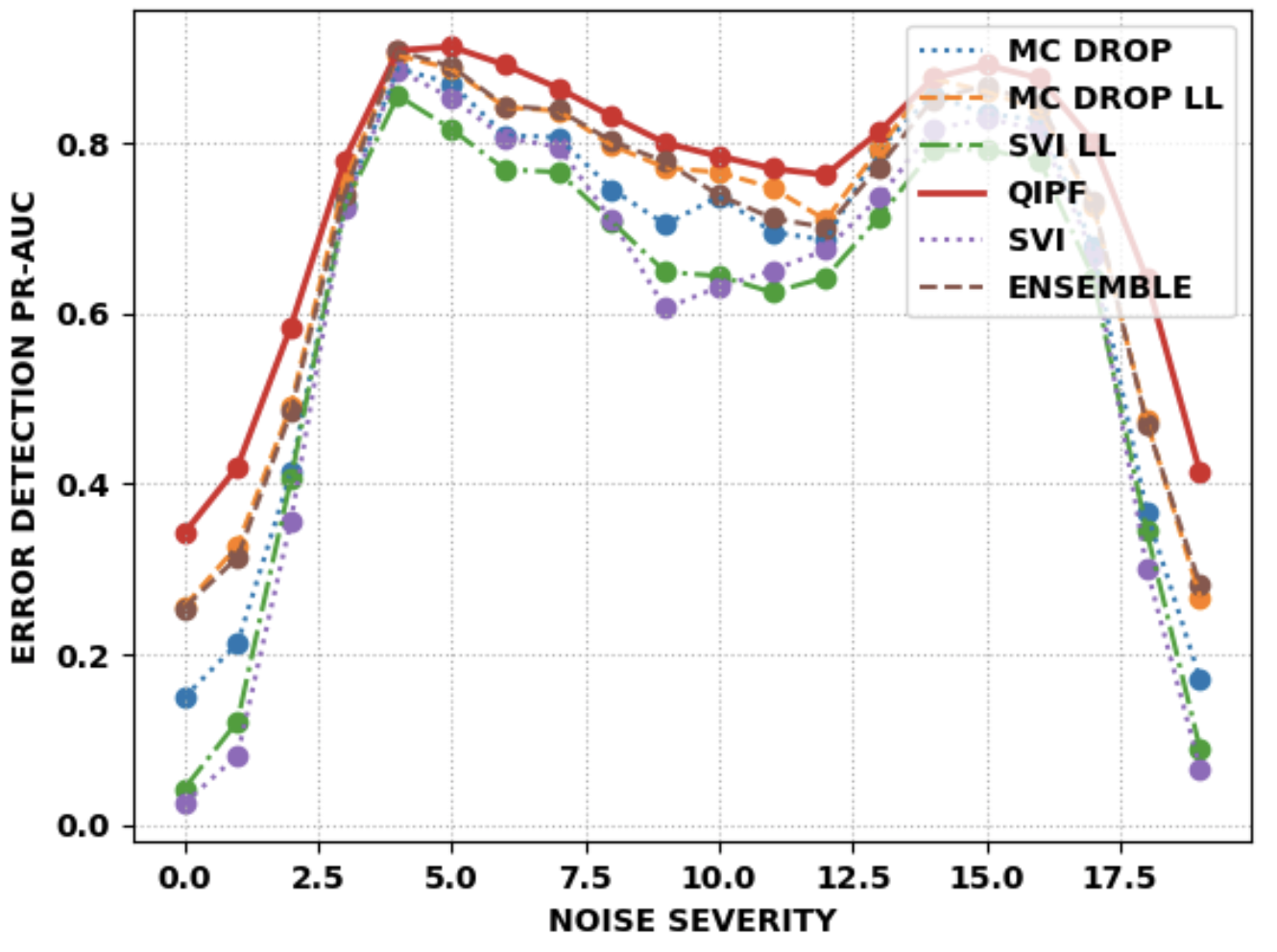}
    \caption\centering{PR-AUC vs noise severity}
  \end{subfigure}
  \begin{subfigure}{0.24\linewidth}
    \centering\includegraphics[scale = 0.19]{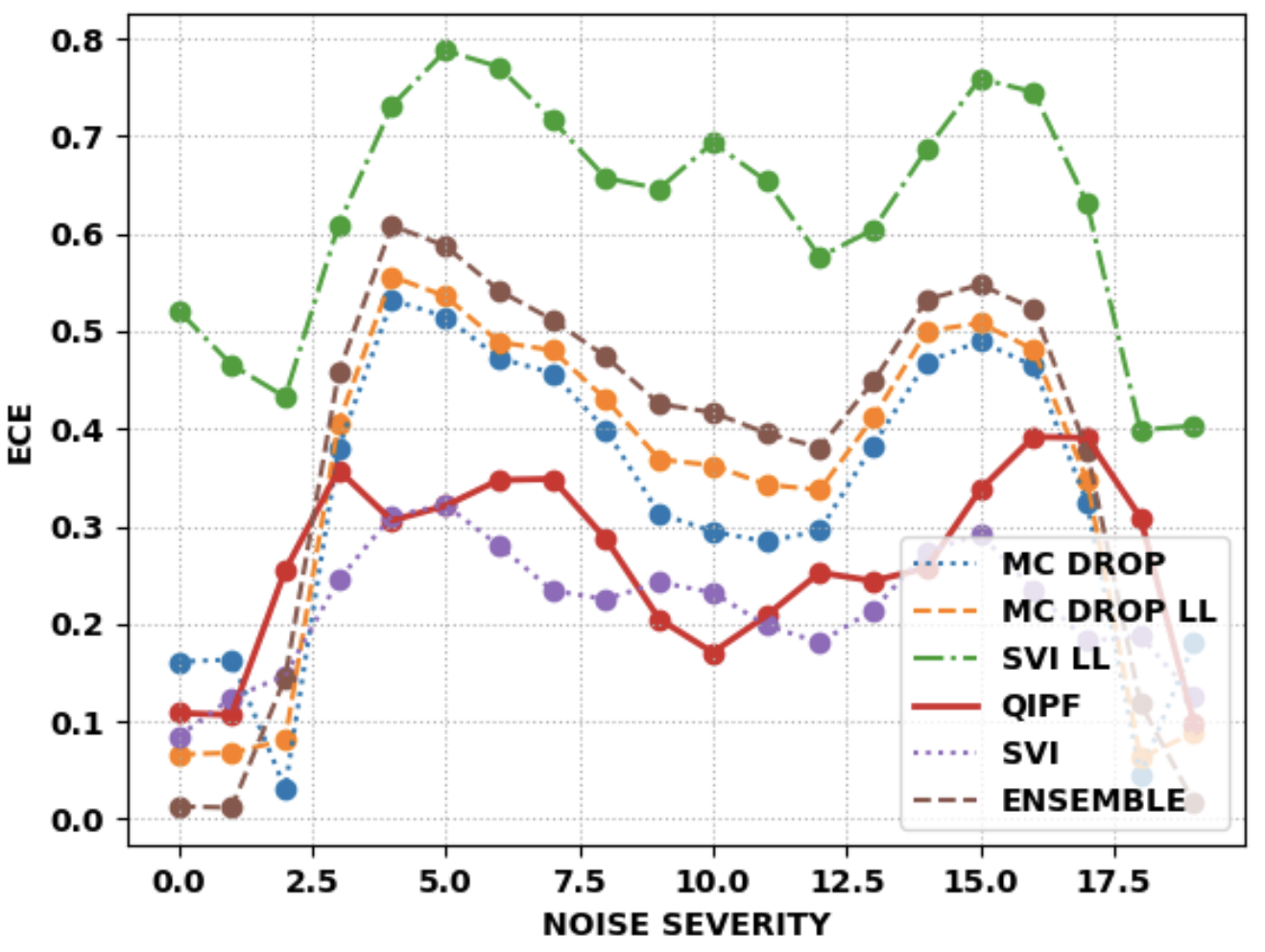}
    \caption{Exp. Calibration Error vs noise severity}
  \end{subfigure}
      \begin{subfigure}{0.24\linewidth}
    \centering\includegraphics[scale = 0.19]{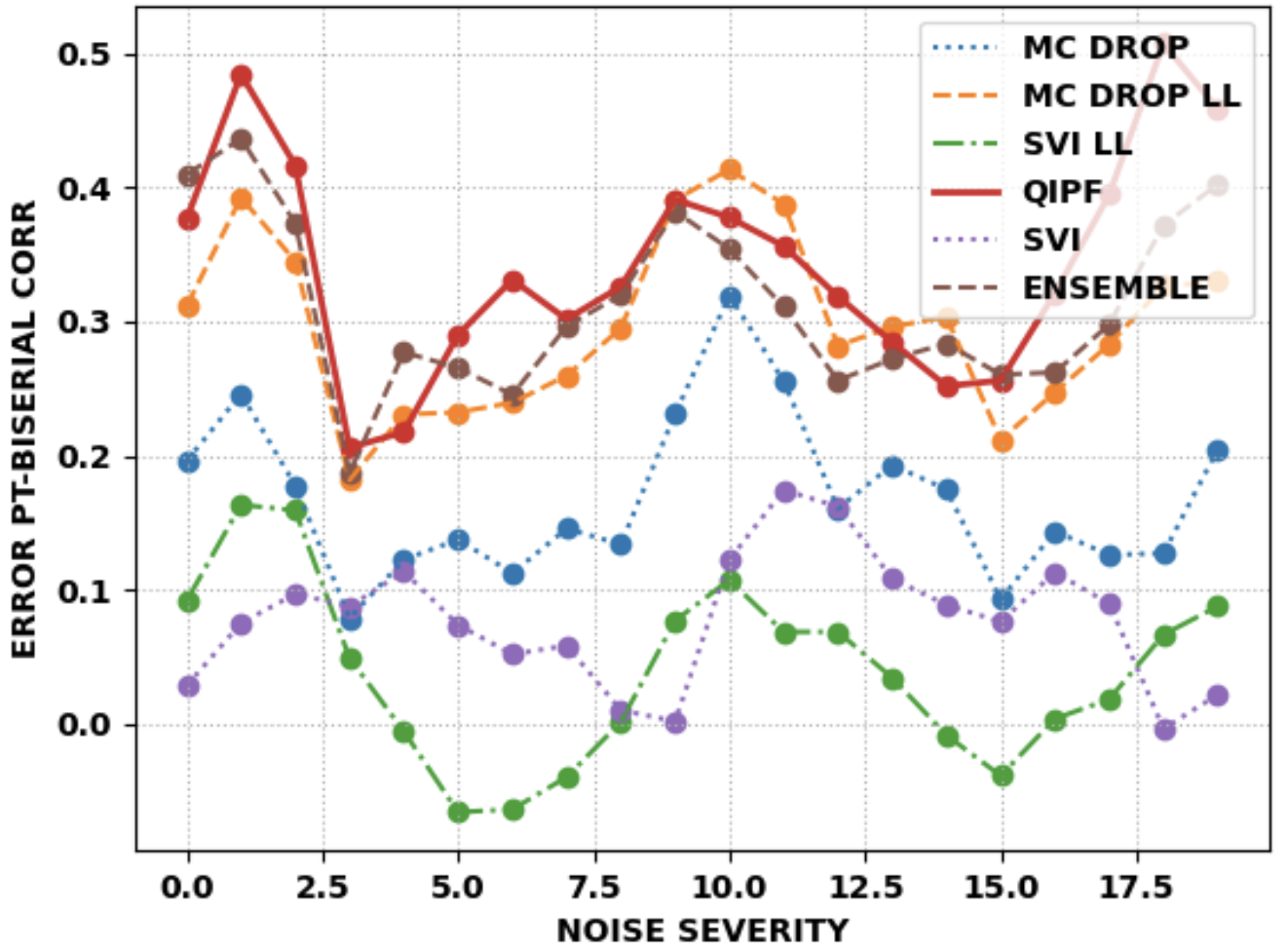}
    \caption{Point-Biserial Error Correlation vs noise severity}
  \end{subfigure}
    \caption{MNIST example: (a) shows the histograms of uncertainty estimates where QIPF can be seen to achieve better class-separation between correct and wrong predictions (very low overlap). (b) and (c) show corresponding error detection ROC and precision-recall curves, (d), (e), (f) and (g) show the graphs of ROC-AUC, PR-AUC, expected calibration error vs noise severity and PT-Biserial error correlation vs noise intensity. QIPF framework can be seen to have better performance in all of the metrics.}
  \label{sc1}
  \end{figure*}

 \section{Experimental Methods and Metrics}
 For comparisons, the experimental setup in \cite{laks2} is followed and the choice of methodologies focus on QIPF. Established comparison benchmarks that do not use prior information about the nature or intensity of test-set corruption to quantify model uncertainty under covariate shift. The selection includes:
 \begin{itemize}
 	\item \textbf{MC Dropout:} Monte Carlo implementation of dropout during model testing \cite{gal, sriv2}.
 	\item \textbf{MC Dropout LL:} MC dropout implemented only over the last model layer \cite{riq}.
 	\item \textbf{SVI:} Stochastic Variational Bayesian Inference \cite{graves, blun}.
 	\item \textbf{SVI LL:} Stochastic Variational Bayesian Inference applied only to the last model layer \cite{riq}.
 	\item \textbf{Ensemble:} Ensembles of T ($=10$) networks independently trained with different random initializations \cite{laks}.
 \end{itemize}

The evaluations of different methods test their ability to detect classification errors of the model in real time when fed with corrupted test data. We measure this ability in terms of areas under ROC and precision-recall (PR) curves associated with the error detection problem. We also evaluate calibration metrics of Brier score and expected calibration error, along with point biserial and Spearman correlation scores between quantified uncertainties and errors. These metrics are described in section D (appendix).

 \section{Results}
 Python 3.7 with tensorflow library is employed to perform all simulations and the datasets are MNIST, K-MNIST (Kuzushiji-MNIST, a more challenging alternative to MNIST) \cite{cla} and CIFAR-10. For MNIST and K-MNIST, we used the regular 2-convolutional layer LeNet network architecture  with standard training procedure and dataset split. Two different networks for modeling CIFAR-10 are implemented. One of them is a VGG-3 architecture consisting of 3 VGG convolution blocks (each consists of two convolution layers followed by batch normalization, max pooling and dropout layers) and the other is ResNet-18. The successive blocks in VGG-3 consisted of 32, 64 and 128 filters respectively in each of their two layers. This was followed by two dense layers (with dropout in the middle) consisting of 128 and 10 nodes respectively. ReLu activation function was used after each layer except the last, where Softmax was used. All networks were trained on uncorrupted datasets using Adam optimizer and categorical cross-entropy cost function. We achieved testing accuracies (on uncorrupted test-set) of 98.6\% for MNIST, 91.9\% for K-MNIST, 87.5\% for CIFAR-10 (using VGG-3) and 91.5\% for CIFAR-10 (using ResNet-18) thereby representing a diverse group of models.\par

\begin{table*}[!h]
\centering
\resizebox{\textwidth}{!}{%
\begin{tabular}{|l|l|llllll|llllll|}
\hline
\multicolumn{1}{|c|}{\multirow{2}{*}{\begin{tabular}[c]{@{}c@{}}CORRUPTION\\ TYPE\end{tabular}}} & \multirow{2}{*}{DATASET}                                    & \multicolumn{6}{c|}{ROC-AUC}                                                                                                                                                                                                                         & \multicolumn{6}{c|}{PR-AUC}                                                                                                                                                                                                                          \\ \cline{3-14} 
\multicolumn{1}{|c|}{}                                                                           &                                                             & \multicolumn{1}{c}{MC-DROP}         & \multicolumn{1}{c}{MC-DROP-LL}      & \multicolumn{1}{c}{SVI}             & \multicolumn{1}{c}{SVI-LL}                   & \multicolumn{1}{c}{ENSEMBLE}        & \multicolumn{1}{c|}{QIPF}                     & \multicolumn{1}{c}{MC-DROP}         & \multicolumn{1}{c}{MC-DROP-LL}      & \multicolumn{1}{c}{SVI}             & \multicolumn{1}{c}{SVI-LL}                   & \multicolumn{1}{c}{ENSEMBLE}        & \multicolumn{1}{c|}{QIPF}                     \\ \hline
\multirow{4}{*}{ROTATION}                                                                        & MNIST                                                       & \multicolumn{1}{c}{0.63 $\pm$ 0.08} & \multicolumn{1}{c}{0.70 $\pm$ 0.09} & \multicolumn{1}{c}{0.55 $\pm$ 0.03} & \multicolumn{1}{c}{0.66 $\pm$ 0.06}          & \multicolumn{1}{c}{\color{blue}{\textbf{0.71 $\pm$ 0.09}}} & \multicolumn{1}{c|}{\color{red}{\textbf{0.75 $\pm$ 0.07}}} & \multicolumn{1}{c}{0.65 $\pm$ 0.23} & \multicolumn{1}{c}{0.69 $\pm$ 0.20} & \multicolumn{1}{c}{0.60 $\pm$ 0.27} & \multicolumn{1}{c}{0.67 $\pm$ 0.24}          & \multicolumn{1}{c}{\color{blue}{\textbf{0.69 $\pm$ 0.20}}} & \multicolumn{1}{c|}{\color{red}{\textbf{0.75 $\pm$ 0.17}}} \\
                                                                                                 & K-MNIST                                                     & 0.49 $\pm$ 0.02                     & 0.52 $\pm$ 0.06                     & 0.52 $\pm$ 0.01                     & 0.54 $\pm$ 0.07                              & \color{blue}{\textbf{0.56 $\pm$ 0.11}}                     & \color{red}{\textbf{0.58 $\pm$ 0.11}}                     & 0.77 $\pm$ 0.22                     & 0.79 $\pm$ 0.19                     & 0.78 $\pm$ 0.21                     & 0.80 $\pm$ 0.18                              & \color{blue}{\textbf{0.81 $\pm$ 0.14}}                     & \color{red}{\textbf{0.84 $\pm$ 0.11}}                      \\
                                                                                                 & \begin{tabular}[c]{@{}l@{}}CIFAR-10\\ (VGG-3)\end{tabular}  & \multicolumn{1}{c}{0.58 $\pm$ 0.07} & \multicolumn{1}{c}{0.61 $\pm$ 0.08} & \multicolumn{1}{c}{0.53 $\pm$ 0.01} & \multicolumn{1}{c}{\color{blue}{\textbf{0.63 $\pm$ 0.05}}}          & \multicolumn{1}{c}{0.57 $\pm$ 0.07} & \multicolumn{1}{c|}{\color{red}{\textbf{0.66 $\pm$ 0.04}}} & \multicolumn{1}{c}{0.68 $\pm$ 0.12} & \multicolumn{1}{c}{0.70 $\pm$ 0.11} & \multicolumn{1}{c}{0.65 $\pm$ 0.18} & \multicolumn{1}{c}{\color{blue}{\textbf{0.72 $\pm$ 0.15}}}          & \multicolumn{1}{c}{0.66 $\pm$ 0.14} & \multicolumn{1}{c|}{\color{red}{\textbf{0.73 $\pm$ 0.17}}} \\ 
                                                                                                 & \begin{tabular}[c]{@{}l@{}}CIFAR-10\\ (RESNET-18)\end{tabular}  & \multicolumn{1}{c}{0.49 $\pm$ 0.02} & \multicolumn{1}{c}{\color{blue}{\textbf{0.62 $\pm$ 0.07}}} & \multicolumn{1}{c}{0.53 $\pm$ 0.01} & \multicolumn{1}{c}{0.60 $\pm$ 0.05} & \multicolumn{1}{c}{0.59 $\pm$ 0.06} & \multicolumn{1}{c|}{\color{red}{\textbf{0.64 $\pm$ 0.06}}}          & \multicolumn{1}{c}{0.67 $\pm$ 0.14} & \multicolumn{1}{c}{\color{blue}{\textbf{0.75 $\pm$ 0.08}}} & \multicolumn{1}{c}{0.71 $\pm$ 0.15} & \multicolumn{1}{c}{0.74 $\pm$ 0.10} & \multicolumn{1}{c}{0.72 $\pm$ 0.11} & \multicolumn{1}{c|}{\color{red}{\textbf{0.76 $\pm$ 0.10}}}          \\ \hline
\multirow{4}{*}{BRIGHTNESS}                                                                      & MNIST                                                       & 0.78 $\pm$ 0.09                     & 0.87 $\pm$ 0.08                     & 0.55 $\pm$ 0.04                     & 0.78 $\pm$ 0.02                              & \color{red}{\textbf{0.90 $\pm$ 0.06}}            & \color{blue}{\textbf{0.87 $\pm$ 0.04}}                               & 0.19 $\pm$ 0.12                     & 0.29 $\pm$ 0.11                     & 0.09 $\pm$ 0.13                     & 0.18 $\pm$ 0.18                              & \color{blue}{\textbf{0.30 $\pm$ 0.16}}            & \color{red}{\textbf{0.35 $\pm$ 0.11}}                               \\
                                                                                                 & K-MNIST                                                     & 0.52 $\pm$ 0.04                     & 0.56 $\pm$ 0.11                     & 0.50 $\pm$ 0.02                     & 0.55 $\pm$ 0.10                              & \color{blue}{\textbf{0.59 $\pm$ 0.16}}                     & \color{red}{\textbf{0.60 $\pm$ 0.16}}                      & 0.70 $\pm$ 0.34                     & \color{red}{\textbf{0.73 $\pm$ 0.30}}                     & 0.70 $\pm$ 0.34                     & \color{blue}{\textbf{0.72 $\pm$ 0.31}}                              & 0.12 $\pm$ 0.11                     & 0.14 $\pm$ 0.19                      \\
                                                                                                 & \begin{tabular}[c]{@{}l@{}}CIFAR-10\\ (VGG-3)\end{tabular}  & 0.69 $\pm$ 0.03                     & \color{blue}{\textbf{0.75 $\pm$ 0.11}}                     & 0.52 $\pm$ 0.03                     & 0.69 $\pm$ 0.07                              & 0.69 $\pm$ 0.11                     & \color{red}{\textbf{0.80 $\pm$ 0.11}}                      & 0.44 $\pm$ 0.16                     & \color{blue}{\textbf{0.50 $\pm$ 0.16}}                     & 0.32 $\pm$ 0.24                     & 0.44 $\pm$ 0.19                              & 0.41 $\pm$ 0.19                     & \color{red}{\textbf{0.61 $\pm$ 0.12}}               \\       
                                                                                                 & \begin{tabular}[c]{@{}l@{}}CIFAR-10\\ (RESNET-18)\end{tabular}  & 0.47 $\pm$ 0.01                     & 0.46 $\pm$ 0.00                     & \multicolumn{1}{c}{0.51 $\pm$ 0.02}                     & \color{red}{\textbf{\textbf{0.57 $\pm$ 0.01}}}                              & \color{blue}{\textbf{0.52 $\pm$ 0.01}}                     & 0.50 $\pm$ 0.02                      & 0.84 $\pm$ 0.01                     & 0.84 $\pm$ 0.01                     & \multicolumn{1}{c}{0.85 $\pm$ 0.11}                    & \color{blue}{\textbf{0.89 $\pm$ 0.01}}                              & \color{red}{\textbf{0.89 $\pm$ 0.01}}                     & 0.87 $\pm$ 0.01                      \\ \hline
\multirow{4}{*}{SHEAR}                                                                           & MNIST                                                       & 0.62 $\pm$ 0.10                     & 0.68 $\pm$ 0.11                     & 0.56 $\pm$ 0.03                     & 0.63 $\pm$ 0.09                              & \color{blue}{\textbf{0.69 $\pm$ 0.11}}                     & \color{red}{\textbf{0.70 $\pm$ 0.09}}                      & 0.64 $\pm$ 0.23                     & 0.68 $\pm$ 0.21                     & 0.59 $\pm$ 0.27                     & 0.64 $\pm$ 0.24                              & \color{blue}{\textbf{0.68 $\pm$ 0.22}}                     & \color{red}{\textbf{0.71 $\pm$ 0.17}}                      \\
                                                                                                 & K-MNIST                                                     & 0.49 $\pm$ 0.03                     & 0.55 $\pm$ 0.08                     & 0.51 $\pm$ 0.02                     & 0.59 $\pm$ 0.06                              & \color{blue}{\textbf{0.61 $\pm$ 0.12}}                     & \color{red}{\textbf{0.64 $\pm$ 0.11}}                      & 0.65 $\pm$ 0.25                     & 0.69 $\pm$ 0.22                     & 0.67 $\pm$ 0.26                     & 0.71 $\pm$ 0.22                              & \color{blue}{\textbf{0.73 $\pm$ 0.17}}                     & \color{red}{\textbf{0.77 $\pm$ 0.14}}                      \\
                                                                                                 & \begin{tabular}[c]{@{}l@{}}CIFAR-10\\ (VGG-3)\end{tabular}  & 0.63 $\pm$ 0.08                     & 0.66 $\pm$ 0.09                     & 0.57 $\pm$ 0.03                     & \color{blue}{\textbf{0.66 $\pm$ 0.04}}                              & 0.62 $\pm$ 0.08                     & \color{red}{\textbf{0.69 $\pm$ 0.02}}                      & 0.66 $\pm$ 0.18                     & 0.68 $\pm$ 0.16                     & 0.62 $\pm$ 0.25                     & \color{blue}{\textbf{0.68 $\pm$ 0.20}}                              & 0.64 $\pm$ 0.20                      & \color{red}{\textbf{0.69 $\pm$ 0.24}} \\                      
                                                                                                & \begin{tabular}[c]{@{}l@{}}CIFAR-10 \\ (RESNET-18)\end{tabular} & 0.52 $\pm$ 0.02                     & \color{blue}{\textbf{0.66 $\pm$ 0.08}}                     & \multicolumn{1}{c}{0.55 $\pm$ 0.02}                    & 0.64 $\pm$ 0.05                              & 0.64 $\pm$ 0.06                     & \color{red}{\textbf{0.68 $\pm$ 0.07}}                      & 0.59 $\pm$ 0.17                     & \color{blue}{\textbf{0.71 $\pm$ 0.10}}                     & \multicolumn{1}{c}{0.69 $\pm$ 0.22}                     & 0.68 $\pm$ 0.14                              & 0.67 $\pm$ 0.14                   &  \color{red}{\textbf{0.71 $\pm$ 0.10}} 
                                                                                                \\ \hline
\multirow{4}{*}{ZOOM}                                                                            & MNIST                                                       & 0.53 $\pm$ 0.12                     & 0.58 $\pm$ 0.15                     & 0.51 $\pm$ 0.02                     & 0.58 $\pm$ 0.11                              & \color{blue}{\textbf{0.63 $\pm$ 0.14}}                     & \color{red}{\textbf{0.70 $\pm$ 0.13}}                      & 0.56 $\pm$ 0.27                     & 0.60 $\pm$ 0.24                     & 0.56 $\pm$ 0.30                     & 0.60 $\pm$ 0.26                              & \color{blue}{\textbf{0.63 $\pm$ 0.24}}                     & \color{red}{\textbf{0.72 $\pm$ 0.20}}                      \\
                                                                                                 & K-MNIST                                                     & 0.49 $\pm$ 0.04                     & 0.54 $\pm$ 0.08                     & 0.51 $\pm$ 0.02                     & 0.58 $\pm$ 0.08                              & \color{blue}{\textbf{0.61 $\pm$ 0.13}}                     & \color{red}{\textbf{0.65 $\pm$ 0.11}}                      & 0.62 $\pm$ 0.26                     & 0.65 $\pm$ 0.23                     & 0.63 $\pm$ 0.26                     & 0.68 $\pm$ 0.22                              & \color{blue}{\textbf{0.71 $\pm$ 0.17}}                     & \color{red}{\textbf{0.75 $\pm$ 0.15}}                      \\
                                                                                                 & \begin{tabular}[c]{@{}l@{}}CIFAR-10\\ (VGG-3)\end{tabular}  & 0.59 $\pm$ 0.08                     & \color{blue}{\textbf{0.63 $\pm$ 0.09}}                     & 0.56 $\pm$ 0.03                     & 0.60 $\pm$ 0.08                              & 0.58 $\pm$ 0.08                     & \color{red}{\textbf{0.64 $\pm$ 0.08}}                      & 0.67 $\pm$ 0.17                     & \color{blue}{\textbf{0.70 $\pm$ 0.15}}                     & 0.65 $\pm$ 0.23                     & 0.68 $\pm$ 0.17                              & 0.65 $\pm$ 0.18                     & \color{red}{\textbf{0.71 $\pm$ 0.17}}      \\                
                                                                                                 & \begin{tabular}[c]{@{}l@{}}CIFAR-10\\ (RESNET-18)\end{tabular}  & 0.52 $\pm$ 0.01                     & \color{blue}{\textbf{0.64 $\pm$ 0.07}}                     & \multicolumn{1}{c}{0.58 $\pm$ 0.02}                     & 0.63 $\pm$ 0.04                              & 0.61 $\pm$ 0.06                     & \color{red}{\textbf{0.65 $\pm$ 0.07}}                      & 0.65 $\pm$ 0.17                     & \color{blue}{\textbf{0.73 $\pm$ 0.01}}                     & \multicolumn{1}{c}{0.68 $\pm$ 0.20}                     & 0.72 $\pm$ 0.14                              & 0.70 $\pm$ 0.13                     & \color{red}{\textbf{0.74 $\pm$ 0.12}}                      \\ \hline
\multirow{4}{*}{SHIFT}                                                                            & MNIST                                                       & 0.65 $\pm$ 0.13                     & \color{blue}{\textbf{0.73 $\pm$ 0.16}}                     & 0.53 $\pm$ 0.03                     & 0.67 $\pm$ 0.12                              & 0.71 $\pm$ 0.18                     & \color{red}{\textbf{0.77 $\pm$ 0.14}}                      & 0.42 $\pm$ 0.28                     & \color{blue}{\textbf{0.49 $\pm$ 0.23}}                     & 0.36 $\pm$ 0.32                     & 0.43 $\pm$ 0.28                              & 0.46 $\pm$ 0.24                     & \color{red}{\textbf{0.59 $\pm$ 0.18}}                      \\
                                                                                                 & K-MNIST                                                     & 0.53 $\pm$ 0.05                     & 0.66 $\pm$ 0.08                     & 0.53 $\pm$ 0.01                     & 0.69 $\pm$ 0.05                              & \color{blue}{\textbf{0.78 $\pm$ 0.10}}                     & \color{red}{\textbf{0.81 $\pm$ 0.07}}                      & 0.27 $\pm$ 0.17                     & 0.36 $\pm$ 0.15                     & 0.28 $\pm$ 0.18                     & 0.38 $\pm$ 0.18                              & \color{blue}{\textbf{0.46 $\pm$ 0.11}}                     & \color{red}{\textbf{0.57 $\pm$ 0.10}}                      \\
                                                                                                 & \begin{tabular}[c]{@{}l@{}}CIFAR-10\\ (VGG-3)\end{tabular}  & 0.75 $\pm$ 0.04                     & \color{red}{\textbf{0.79 $\pm$ 0.03}}                     & 0.53 $\pm$ 0.01                     & 0.72 $\pm$ 0.02                              & 0.72 $\pm$ 0.04                     & \color{blue}{\textbf{0.75 $\pm$ 0.02}}                      & 0.37 $\pm$ 0.03                     & \color{red}{\textbf{0.43 $\pm$ 0.04}}                     & 0.24 $\pm$ 0.05                     & 0.37 $\pm$ 0.05                              & 0.34 $\pm$ 0.03                     & \color{blue}{\textbf{0.38 $\pm$ 0.10}}      \\                
                                                                                                 & \begin{tabular}[c]{@{}l@{}}CIFAR-10\\ (RESNET-18)\end{tabular}  & 0.56 $\pm$ 0.01                     & \color{red}{\textbf{0.79 $\pm$ 0.01}}                     & \multicolumn{1}{c}{0.53 $\pm$ 0.02}                     & 0.71 $\pm$ 0.01                              & 0.74 $\pm$ 0.01                     & \color{blue}{\textbf{0.78 $\pm$ 0.01}}                      & 0.31 $\pm$ 0.28                     & \color{red}{\textbf{0.54 $\pm$ 0.01}}                     & \multicolumn{1}{c}{0.32 $\pm$ 0.10}                     & 0.45 $\pm$ 0.01                              & 0.44 $\pm$ 0.01                     & \color{blue}{\textbf{0.50 $\pm$ 0.01}}                      \\ \hline
\end{tabular}%
}
\caption{Average ROC-AUC values (left) and average PR-AUC values (right) of different methods for all datasets and corruption types over all corruption intensities. Best values are boldened in red and second best in blue.}
\label{t1}
\end{table*}

\begin{table*}[!h]
\centering
\resizebox{\textwidth}{!}{%
\begin{tabular}{|l|l|llllll|llllll|}
\hline
\multicolumn{1}{|c|}{\multirow{2}{*}{\begin{tabular}[c]{@{}c@{}}CORRUPTION\\ TYPE\end{tabular}}} & \multirow{2}{*}{DATASET}                                    & \multicolumn{6}{c|}{EXPECTED CALIBRATION ERROR}                                                                                                                                                                                                                         & \multicolumn{6}{c|}{BRIER SCORE}                                                                                                                                                                                                                          \\ \cline{3-14} 
\multicolumn{1}{|c|}{}                                                                           &                                                             & \multicolumn{1}{c}{MC-DROP}         & \multicolumn{1}{c}{MC-DROP-LL}      & \multicolumn{1}{c}{SVI}             & \multicolumn{1}{c}{SVI-LL}                   & \multicolumn{1}{c}{ENSEMBLE}        & \multicolumn{1}{c|}{QIPF}                     & \multicolumn{1}{c}{MC-DROP}         & \multicolumn{1}{c}{MC-DROP-LL}      & \multicolumn{1}{c}{SVI}             & \multicolumn{1}{c}{SVI-LL}                   & \multicolumn{1}{c}{ENSEMBLE}        & \multicolumn{1}{c|}{QIPF}                     \\ \hline
\multirow{4}{*}{ROTATION}                                                                        & MNIST                                                       & \multicolumn{1}{c}{0.33 $\pm$ 0.15} & \multicolumn{1}{c}{0.39 $\pm$ 0.13} & \multicolumn{1}{c}{\color{red}{\textbf{0.22 $\pm$ 0.06}}} & \multicolumn{1}{c}{0.65 $\pm$ 0.10}          & \multicolumn{1}{c}{0.38 $\pm$ 0.30} & \multicolumn{1}{c|}{\color{blue}{\textbf{0.26 $\pm$ 0.09}}} & \multicolumn{1}{c}{0.34 $\pm$ 0.11} & \multicolumn{1}{c}{\color{red}{\textbf{0.32 $\pm$ 0.12}}} & \multicolumn{1}{c}{\color{blue}{\textbf{0.34 $\pm$ 0.05}}} & \multicolumn{1}{c}{0.81 $\pm$ 0.16}          & \multicolumn{1}{c}{0.37 $\pm$ 0.14} & \multicolumn{1}{c|}{0.35 $\pm$ 0.10} \\
                                                                                                 & K-MNIST                                                     & 0.58 $\pm$ 0.18                     & 0.47 $\pm$ 0.16                     & \color{blue}{\textbf{0.15 $\pm$ 0.03}}                     & 0.79 $\pm$ 0.12                              & 0.45 $\pm$ 0.16                     & \color{red}{\textbf{0.09 $\pm$ 0.02}}                      & 0.40 $\pm$ 0.09                     & 0.39 $\pm$ 0.08                     & \color{blue}{\textbf{0.30 $\pm$ 0.02}}                     & 0.98 $\pm$ 0.17                              & 0.37 $\pm$ 0.09                     & \color{red}{\textbf{0.22 $\pm$ 0.02}}                      \\
                                                                                                 & \begin{tabular}[c]{@{}l@{}}CIFAR-10\\ (VGG-3)\end{tabular}  & \multicolumn{1}{c}{0.33 $\pm$ 0.12} & \multicolumn{1}{c}{0.33 $\pm$ 0.11} & \multicolumn{1}{c}{\color{red}{\textbf{0.14 $\pm$ 0.02}}} & \multicolumn{1}{c}{1.69 $\pm$ 0.43}          & \multicolumn{1}{c}{0.42 $\pm$ 0.16} & \multicolumn{1}{c|}{\color{blue}{\textbf{0.25 $\pm$ 0.07}}} & \multicolumn{1}{c}{0.29 $\pm$ 0.20} & \multicolumn{1}{c}{0.23 $\pm$ 0.13} & \multicolumn{1}{c}{\color{blue}{\textbf{0.23 $\pm$ 0.12}}} & \multicolumn{1}{c}{2.90 $\pm$ 4.09}          & \multicolumn{1}{c}{0.54 $\pm$ 0.22} & \multicolumn{1}{c|}{\color{red}{\textbf{0.18 $\pm$ 0.15}}} \\ 
                                                                                                 & \begin{tabular}[c]{@{}l@{}}CIFAR-10\\ (RESNET-18)\end{tabular}  & \multicolumn{1}{c}{0.36 $\pm$ 0.1} & \multicolumn{1}{c}{0.58 $\pm$ 0.15} & \multicolumn{1}{c}{\color{blue}{\textbf{0.20 $\pm$ 0.03}}} & \multicolumn{1}{c}{0.58 $\pm$ 0.02} & \multicolumn{1}{c}{0.33 $\pm$ 0.11} & \multicolumn{1}{c|}{\color{red}{\textbf{0.18 $\pm$ 0.15}}}          & \multicolumn{1}{c}{0.32 $\pm$ 0.05} & \multicolumn{1}{c}{0.48 $\pm$ 0.10} & \multicolumn{1}{c}{\color{blue}{\textbf{0.24 $\pm$ 0.11}}} & \multicolumn{1}{c}{0.51 $\pm$ 0.11} & \multicolumn{1}{c}{0.25 $\pm$ 0.05} & \multicolumn{1}{c|}{\color{red}{\textbf{0.18 $\pm$ 0.10}}}          \\ \hline
\multirow{4}{*}{SHEAR}                                                                           & MNIST                                                       & 0.32 $\pm$ 0.14                     & 0.33 $\pm$ 0.18                     & \color{red}{\textbf{0.17 $\pm$ 0.04}}                     & 0.59 $\pm$ 0.09                              & 0.36 $\pm$ 0.20                     & \color{blue}{\textbf{0.24 $\pm$ 0.08}}                      & 0.29 $\pm$ 0.11                     & 0.29 $\pm$ 0.13                     & \color{red}{\textbf{0.28 $\pm$ 0.04}}                     & 0.72 $\pm$ 0.13                              & 0.32 $\pm$ 0.15                     & \color{blue}{\textbf{0.30 $\pm$ 0.09}}                      \\
                                                                                                 & K-MNIST                                                     & 0.40 $\pm$ 0.15                     & 0.38 $\pm$ 0.17                     & \color{blue}{\textbf{0.17 $\pm$ 0.03}}                     & 0.72 $\pm$ 0.10                              & 0.38 $\pm$ 0.18                     & \color{red}{\textbf{0.10 $\pm$ 0.02}}                      & 0.36 $\pm$ 0.10                     & 0.35 $\pm$ 0.2                     & \color{blue}{\textbf{0.31 $\pm$ 0.02}}                     & 0.88 $\pm$ 0.15                              & 0.33 $\pm$ 0.11                     & \color{red}{\textbf{0.21 $\pm$ 0.03}}                      \\
                                                                                                 & \begin{tabular}[c]{@{}l@{}}CIFAR-10\\ (VGG-3)\end{tabular}  & 0.33 $\pm$ 0.18                     & 0.36 $\pm$ 0.18                     & \color{red}{\textbf{0.16 $\pm$ 0.02}}                     & 4.24 $\pm$ 0.98                              & 0.36 $\pm$ 0.19                     & \color{blue}{\textbf{0.32 $\pm$ 0.18}}                      & 0.42 $\pm$ 0.19                     & 0.30 $\pm$ 0.15                     & \color{red}{\textbf{0.07 $\pm$ 0.06}}                     & 2.56 $\pm$ 1.28                              & 0.09 $\pm$ 0.03                      & \color{blue}{\textbf{0.08 $\pm$ 0.24}} \\                      
                                                                                                & \begin{tabular}[c]{@{}l@{}}CIFAR-10 \\ (RESNET-18)\end{tabular} & 0.25 $\pm$ 0.12                     & 0.49 $\pm$ 0.18                     & \multicolumn{1}{c}{\color{red}{\textbf{0.17 $\pm$ 0.02}}}                     & 0.62 $\pm$ 0.19                              & 0.26 $\pm$ 0.15                     & \color{blue}{\textbf{0.24 $\pm$ 0.19}}                      & 0.41 $\pm$ 0.07                     & 0.55 $\pm$ 0.16                     & \multicolumn{1}{c}{\color{red}{\textbf{0.30 $\pm$ 0.05}}}                     & 0.68 $\pm$ 0.10                              & 0.37 $\pm$ 0.07                   &  \color{blue}{\textbf{0.36 $\pm$ 0.12}} 
                                                                                                \\ \hline
\multirow{4}{*}{ZOOM}                                                                            & MNIST                                                       & 0.37 $\pm$ 0.18                     & \color{blue}{\textbf{0.35 $\pm$ 0.22}}                     & \color{red}{\textbf{0.19 $\pm$ 0.06}}                     & 0.70 $\pm$ 0.16                              & 0.36 $\pm$ 0.25                     & 0.73 $\pm$ 0.44                      & 0.31 $\pm$ 0.14                     & \color{blue}{\textbf{0.30 $\pm$ 0.16}}                     & \color{red}{\textbf{0.29 $\pm$ 0.05}}                     & 0.84 $\pm$ 0.21                              & 0.32 $\pm$ 0.18                     & 0.97 $\pm$ 0.63                      \\
                                                                                                 & K-MNIST                                                     & 0.39 $\pm$ 0.17                     & 0.37 $\pm$ 0.18                     & \color{blue}{\textbf{0.13 $\pm$ 0.07}}                     & 0.67 $\pm$ 0.11                              & 0.35 $\pm$ 0.19                     & \color{red}{\textbf{0.12 $\pm$ 0.02}}                      & 0.35 $\pm$ 0.11                     & 0.33 $\pm$ 0.11                     & \color{blue}{\textbf{0.25 $\pm$ 0.05}}                     & 0.81 $\pm$ 0.15                              & 0.30 $\pm$ 0.12                     & \color{red}{\textbf{0.18 $\pm$ 0.04}}                      \\
                                                                                                 & \begin{tabular}[c]{@{}l@{}}CIFAR-10\\ (VGG-3)\end{tabular}  & 0.34 $\pm$ 0.21                     & 0.40 $\pm$ 0.18                     & \color{red}{\textbf{0.16 $\pm$ 0.02}}                     & 1.44 $\pm$ 0.80                              & 0.41 $\pm$ 0.22                     & \color{blue}{\textbf{0.25 $\pm$ 0.17}}                      & 0.52 $\pm$ 0.10                     & 0.53 $\pm$ 0.12                     & \color{red}{\textbf{0.26 $\pm$ 0.07}}                     & 1.27 $\pm$ 0.84                              & 0.46 $\pm$ 0.18                     & \color{blue}{\textbf{0.27 $\pm$ 0.07}}      \\                
                                                                                                 & \begin{tabular}[c]{@{}l@{}}CIFAR-10\\ (RESNET-18)\end{tabular}  & 0.28 $\pm$ 0.13                     & 0.54 $\pm$ 0.18                     & \multicolumn{1}{c}{\color{blue}{\textbf{0.19 $\pm$ 0.05}}}                     & 1.08 $\pm$ 0.11                              & 0.30 $\pm$ 0.17                     & \color{red}{\textbf{0.18 $\pm$ 0.18}}                      & 0.35 $\pm$ 0.05                     & 0.52 $\pm$ 0.17                     & \multicolumn{1}{c}{\color{blue}{\textbf{0.27 $\pm$ 0.08}}}                     & 1.05 $\pm$ 0.81                              & 0.34 $\pm$ 0.08                     & \color{red}{\textbf{0.25 $\pm$ 0.14}}                     \\ \hline
\multirow{4}{*}{SHIFT}                                                                            & MNIST                                                       & 0.23 $\pm$ 0.15                     & 0.20 $\pm$ 0.19                     & \color{red}{\textbf{0.15 $\pm$ 0.07}}                     & 0.59 $\pm$ 0.10                              & \color{blue}{\textbf{0.19 $\pm$ 0.22}}                     & 0.20 $\pm$ 0.10                      & \color{blue}{\textbf{0.22 $\pm$ 0.12}}                     & \color{red}{\textbf{0.19 $\pm$ 0.14}}                     & 0.27 $\pm$ 0.06                     & 0.69 $\pm$ 0.16                              & 0.22 $\pm$ 0.18                     & 0.29 $\pm$ 0.14                      \\
                                                                                                 & K-MNIST                                                     & 0.18 $\pm$ 0.05                     & 0.13 $\pm$ 0.07                     & 0.17 $\pm$ 0.01                     & 0.55 $\pm$ 0.07                              & \color{blue}{\textbf{0.09 $\pm$ 0.09}}                     & \color{red}{\textbf{0.08 $\pm$ 0.02}}                      & 0.21 $\pm$ 0.07                     & 0.18 $\pm$ 0.08                     & 0.28 $\pm$ 0.01                     & 0.64 $\pm$ 0.10                              & \color{red}{\textbf{0.15 $\pm$ 0.09}}                     & \color{blue}{\textbf{0.18 $\pm$ 0.04}}                      \\
                                                                                                 & \begin{tabular}[c]{@{}l@{}}CIFAR-10\\ (VGG-3)\end{tabular}  & 0.07 $\pm$ 0.03                     & 0.09 $\pm$ 0.02                     & 0.17 $\pm$ 0.01                     & 2.38 $\pm$ 0.60                              & \color{blue}{\textbf{0.05 $\pm$ 0.03}}                     & \color{red}{\textbf{0.04 $\pm$ 0.03}}                      & 0.35 $\pm$ 0.18                     & 0.37 $\pm$ 0.14                     & 0.39 $\pm$ 0.05                     & 1.96 $\pm$ 0.92                              & \color{red}{\textbf{0.16 $\pm$ 0.03}}                     & \color{blue}{\textbf{0.22 $\pm$ 0.05}}      \\                
                                                                                                 & \begin{tabular}[c]{@{}l@{}}CIFAR-10\\ (RESNET-18)\end{tabular}  & \color{blue}{\textbf{0.1 $\pm$ 0.01}}                     & 0.21 $\pm$ 0.02                     & \multicolumn{1}{c}{0.17 $\pm$ 0.02}                     & 1.07 $\pm$ 0.08                              & \color{red}{\textbf{0.03 $\pm$ 0.01}}                     & 0.53 $\pm$ 0.02                      & \color{blue}{\textbf{0.31 $\pm$ 0.28}}                     & 0.34 $\pm$ 0.01                     & \multicolumn{1}{c}{0.38 $\pm$ 0.05}                     & 2.41 $\pm$ 0.65                              & \color{red}{\textbf{0.11 $\pm$ 0.02}}                     & 0.44 $\pm$ 0.02                      \\ \hline

\end{tabular}%
}
\caption{Average ECE (left) and average Brier scores (right) of different methods for all datasets and corruption types over all corruption intensities. Best values are boldened in red and second best in blue. Lower values are better.}
\label{t2}
\end{table*}

\begin{table}[]
\begin{tabular}{|l|l|l|l|}
\hline
\multirow{2}{*}{MODEL} & \multicolumn{3}{c|}{TIME (ms) PER TEST SAMPLE} \\ \cline{2-4} 
                       & MC DROP        & MC DROP LL       & QIPF       \\ \hline
LeNeT                  & 3.84           & 3.55             & 0.64       \\ \hline
VGG-3                  & 30.03          & 25.75            & 0.88       \\ \hline
ResNet-18                 & 51.33          & 39.72            & 1.86       \\ \hline
\end{tabular}
\caption{Computational time analysis}
\label{t3}
\end{table}

 To evaluate the UQ methods on the trained models, we corrupted the test-sets of datasets with different intensities using common corruption techniques \cite{hend}. Some of the corruption techniques are illustrated in section E (appendix). For QIPF implementation, we first evaluated the cross - IPF, given by (\ref{vs}) at each model's test prediction (the maximum last layer output value before softmax thresholding) in the field created by model weights. The number of weights used for cross-IPF computation were reduced by average pooling of weights at each layer so that we used only 1022 pooled weight values in LeNet, 2400 in VGG-3 and 3552 in ResNet. The kernel width was set as the bandwidth determined by the Silverman's rule \cite{sil} multiplied by a factor that is determined through cross-validation over a part of the original/uncorrupted training dataset (found to be 80 in all cases). This was followed by computation of uncertainty modes of the QIPF (\ref{sh}). We extracted 4 QIPF modes for LeNet and VGG-3 models and 10 modes for ResNet-18. We took the average value of the modes at any prediction to be its uncertainty score. For other methods, the same implementation strategy as \cite{laks2} is followed and details are presented in section F (appendix). We illustrate the QIPF implementation on MNIST in Fig. \ref{sc1}. Fig \ref{sc1}a shows the histogram plots of uncertainty estimates of the different methods corresponding to the correct test-set predictions (blue) and the wrong predictions (red), with the test-set being corrupted by 270 degrees of rotation. The class-separation ability of the QIPF decomposition framework can be seen to be significantly better than the other methods with the most frequent uncertainty score values for the correct and wrong predictions being further apart from each other when compared to other methods. This is also evident from the ROC and precision-recall (PR) curves associated with prediction error detection of the different uncertainty quantification methods in Fig. \ref{sc1}b and Fig. \ref{sc1}c respectively where the QIPF has the highest area under the curve (AUC) values. Fig. \ref{sc1}d and \ref{sc1}e show the ROC-AUC and PR-AUC associated with the UQ methods for different intensities/severity of rotation corruption. Also shown are the expected calibration errors and point-biserial correlation coefficients in Fig. \ref{sc1}f and \ref{sc1}g respectively between the test-set prediction errors and the uncertainty estimates at each intensity. We observe here that the QIPF decomposition method performs significantly better than all other methods in terms of error detection as well as error correlation and calibration. Similar visualizations for other datasets/models are provided in section G (appendix). Table \ref{t1} summarizes the average ROC-AUC and PR-AUC values over all corruption intensities for different types of test-set corruptions for each dataset/model and UQ method. Best values are boldened in red and second best values in blue. We can see here that the QIPF outperforms other methods in most cases while ensemble method is seen to perform the second best, which is expected as it's been previously reported to be the state-of-the-art \cite{laks2}. Table \ref{t2} similarly summarizes the average expected calibrations errors and average brier scores of the different methods for different models. We see here that QIPF yields the best calibration values in most cases followed by stochastic variational inference. Correlation measures between uncertainty estimates and the prediction errors are given in section H (appendix).

\section{Computational Cost Discussion}
 The only computational parts of QIPF are the cross-IPF ($\psi_\mathbf{w}$) which grows linearly with the number of weights and moment extraction growing linearly with the number of moments. The time complexity of the QIPF for each test iteration therefore becomes $\mathcal{O}(n+m)$, where $n$ and $m$ are the number of weights and number of moments respectively, which is the best computation complexity among all the methods. In spite of this, we see here that we are able to outperform all methods with very low values of $n$ (since we do average pooling of weights at the layers) and low value of $m$ (4 to 10 QIPF modes). This is unlike ensemble or Monte Carlo based methods such as MC dropout where reducing the number of sampling steps or size of ensembles has considerably negative effects on accuracy of uncertainty estimation. The compute times of our implementations of QIPF and MC dropout per test sample is shown in Table \ref{t3}. It can be seen that QIPF is much faster.

\section{Conclusion}
This paper shows that by implementing a functional decomposition of a trained model's weight PDF represented in an RKHS, one can locally measure the interactions between the weights and the model predictions to yield precise estimates of predictive uncertainty in a computationally efficient manner. This is made possible by a set of RKHS operators that quantify local anisotropy over the projected weight space, which to the best of our knowledge is beyond the capability of statistical methods. Moreover, the framework of perturbation theory helps understand the role of the moment decomposition and allows for a possible classification of different types of uncertainty. We applied this operator based approach on a important problem of uncertainty quantification in the presence of covariate shifts in the test data and showed that it outperforms established Bayesian and ensemble approaches across various metrics that measure the quality of uncertainty estimates. Furthermore, it is also significantly faster to compute and scales better than existing approaches because it is free of normalizations as conventional Bayesian approaches. In the future, we plan to extend the analysis of the framework to more large scale applications.
\section{Acknowledgments}
This work was partially supported by DARPA under grant no. FA9453-18-1-0039 and ONR under grant no. N00014-21-1-2345.

\bibliographystyle{IEEEtran}
\bibliography{aaai22}

\onecolumn
\appendix
\section*{A. Lower Bound for Information Potential}
Suppose $\Psi(f,.)$ be a vector in the RKHS $H_\Psi$ induced by the kernel $\Psi$ and $M$ be a subspace of $H_\Psi$ that is spanned by $N$ linearly independent vectors $\Psi(g_1, .), \Psi(g_1, .), ... ,\Psi(g_N, .) \in H_\Psi$. Then,

\begin{equation}
    \int{f(x)^2 dx} = ||\Psi(f,.)||^2 \geq \sum_{i,j=1}^{N}\big<\Psi(f,.),\Psi(g_i,.)\big>_{H_\Psi} \times G^{-1}(i,j) \times \big<\Psi(f,.), \Psi(f,.)\big>_{H_\Psi}.
    \label{lb}
\end{equation}

where $G(i,j)$ is the $N \times N$ Gram matrix whose $(i,j)$ term is $\big<\Psi(f,.), \Psi(g_i, .)\big>_{H_\Psi}$.

\textit{proof}: Using projection theorem \cite{projection}, one can obtain the orthogonal projections of $\Psi(f,.)$ onto the subspace $M$ as

\begin{equation}
    P(\Psi(f,.)|M) = \sum_{i,j=1}^{N}\big<\Psi(f,.),\Psi(f,.)\big>_{H_\psi}G^{-1}(i,j)V(g_j,.).
\end{equation}

The Gram matrix is positive definite for linear independent vectors and therefore the inverse always exists. One can now calculate the norm square of the projected vector \cite{29} as follows:

\begin{equation}
    ||P(\Psi(f,.)|M)||^2 = \big<\Psi(f,.), P(\Psi(f,.)|M)\big>_{H_Psi} = \sum_{i,j=1}^{N}\big< \Psi(f,.),\Psi(g_i,.) \big>_{H_\Psi}G^{-1}(i,j) \times \big< \Psi(f,.),\Psi(g_i,.) \big>
    \label{k1}
\end{equation}

The projection residual, on the other hand, is given in \cite{resd} as:

\begin{equation}
    d^2 = ||\Psi(f,.)||^2 - ||P(\Psi(f,.)|M||^2 \geq 0
    \label{k2}
\end{equation}

Combining (\ref{k1}) and (\ref{k2}), one arrives at the lower bound expression given by (\ref{lb}).

\section*{B. Connection of IPF with Kernel Mean Embedding Theory}
The kernel mean embedding (KME) theory \cite{embor, emb} allows one to non-parametrically and universally quantify a data distribution $\mathbb{P}$ from the input space $x_i \in \mathbb{P}$ as an element of its associated RKHS and is given by:

\begin{equation}
\phi(\mathbb{P}) = \mu_{\mathbb{P}} = \int k(x, .)d\mathbb{P}(x).
\end{equation}

The KME is therefore a functional representation of the data PDF induced by points $x$. It has several useful properties, one of which is the property of injectivity for characteristic kernels, meaning that $\mu_\mathbb{P} = \mu_\mathbb{T}$ if and only if $\mathbb{P} = \mathbb{T}$. This therefore allows for unique characterizations of data/model PDF. In practice, one does not have prior information regarding the specific nature of the data PDF, $\mathbb{P}$. Hence we typically rely on the unbiased empirical estimate of the KME given by $\hat\mu = \frac{1}{n}\sum_{t=1}^{n}k(x_t,.)$, which converges to $\mu$ for $n \rightarrow \infty$ in accordance with the law of large numbers. One can notice that the expression for empirical KME is similar to that of IPF. The IPF is therefore a general PDF estimator evaluated directly on input data samples. Being an evaluation of KME, it ensures unbiased and efficient PDF estimation, with convergence independent of data dimensionality \cite{emb}.

\section*{C. QIPF Uncertainty Behavior with Model Regularization}

  \begin{figure*}[!h]
  \begin{subfigure}{0.5\linewidth}
    \centering\includegraphics[scale = 0.25]{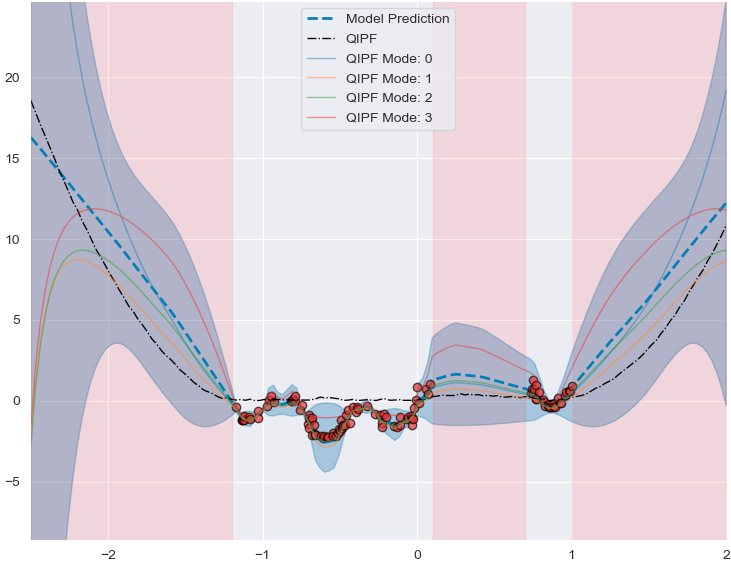}
    \caption\centering{No regularization}
  \end{subfigure}
  \begin{subfigure}{0.5\linewidth}
    \centering\includegraphics[scale = 0.27]{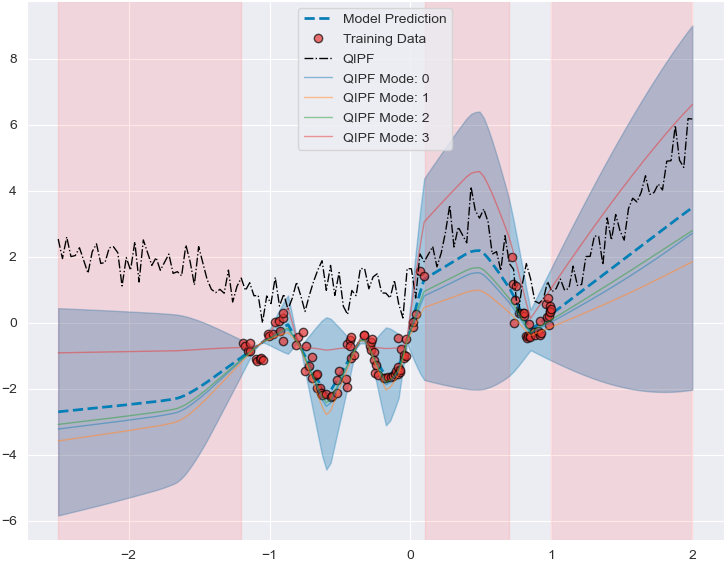}
    \caption\centering{L2 regularization: 0.01}
  \end{subfigure}
  \begin{subfigure}{0.5\linewidth}
    \centering\includegraphics[scale = 0.27]{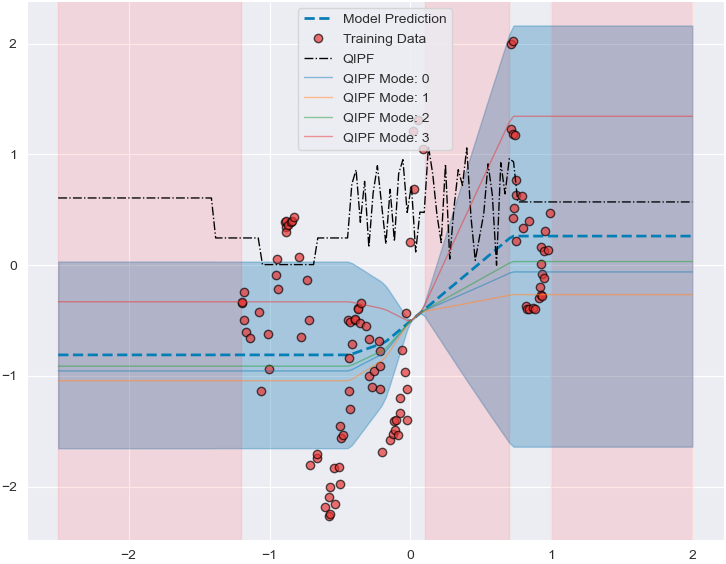}
    \caption\centering{L2 regularization: 0.1}
  \end{subfigure}
    \begin{subfigure}{0.5\linewidth}
    \centering\includegraphics[scale = 0.27]{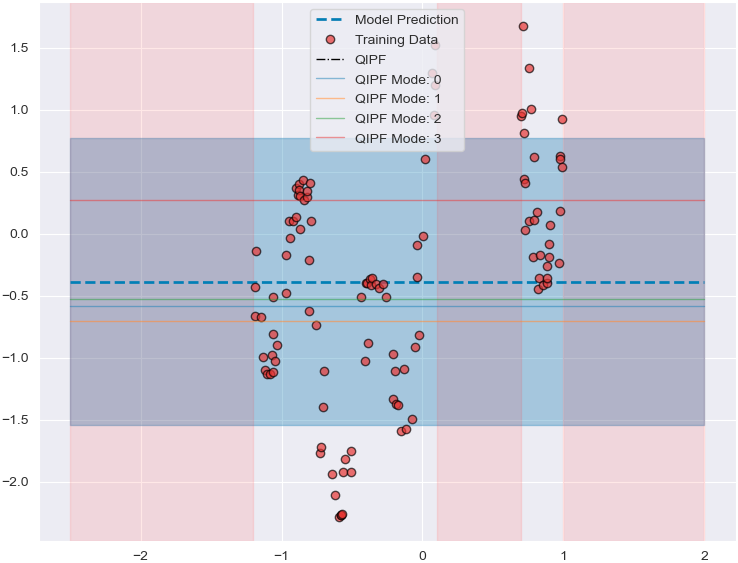}
    \caption\centering{L2 regularization: 0.2}
  \end{subfigure}
    \caption{Model uncertainty quantified by QIPF modes (centered and normalized with respect to model predictions) at different regularization rates at which the model was trained (white regions: seen data regions, pink regions: unseen data regions): (a) shows the QIPF moments to immediately increase outside of the model's training zone hence demonstrating high sensitivity towards the inherent uncertainty of an overfit model. (b-d) shows it to be well calibrated with model's regularization, i.e. QIPF moments decrease outside of the training zone (indicating higher model confidence) while slightly increasing inside the training zone thereby correctly indicating the true nature of the posterior predictive PDF which sacrifices bias for generalization during regularization.}
  \label{sp4}
  \end{figure*}
We implement QIPF decomposition on an over-parameterized and fully connected multi-layer perceptron consisting of 3 hidden layers with 100 neurons in each layer that is trained on just 120 data pairs of a toy regression problem of learning a weighted sine-wave function with added noise. The training is carried out using different L2-regularization intensities (including no regularization that corresponds to a highly overfit model) using training samples that were generated only in the white regions. Pink regions represent regions with no training data. The model was tested in the entire data region [-2, 2.5] and QIPF framework was implemented by extracting 4 modes and taking their average as the uncertainty range. The weight-QIPF modes were centered with the sample amplitude for easier visualization. As can be seen from Fig. \ref{sp4}a, for a fully biased (overfit) model, the QIPF uncertainty immediately increases outside of the seen regions (white) of the model, with fundamental order moment being the highest. This indicates that for an overfit model such as this one, the weights are completely concentrated regions corresponding to the training data and the weight PDF becomes highly peaky in those regions because of which, if one moves even very slightly away from the peaks (towards the regions of initial order QIPF moments), the heterogeneity of the weight PDF becomes very large causing high values of those QIPF moments. Inside the seen region, the QIPF uncertainty can be seen to be very low (model is highly confident in those regions). As regularization is successively increased in Figs. \ref{sp4}b-d, the QIPF uncertainty ranges indicate that the model begins to become more confident outside of the training regions while sacrificing bias (indicated by increased QIPF uncertainty) inside the training zones. This indicates that the QIPF captures true posterior of model.

\section*{D. Metrics for Comparing Uncertainty Measures}
Apart from metrics such as ROC - AUC and PR - AUC for evaluating the capability of detecting prediction errors, we also compute two calibration metrics for the different UQ methods which are the expected calibration error (ECE) and Brier score. Here we measure the calibration of the highest probability value in the prediction layer with respect to model's prediction error. The metrics described as follows:


\subsection*{D.1 Expected Calibration Error}
Expected calibration error measures the difference between a neural network's prediction probability estimate and its accuracy \cite{ece}. Suppose all the confidence scores of the test-set are binned into $K$ distinct bins, then it is the average gap between within bucket accuracy and within bucket predicted confidence (pseudo-probability) given as:

\begin{equation*}
    ECE = \sum_{k=1}^{K}{\frac{|B_k|}{N}}|acc(B_k) - conf (B_k)|
\end{equation*}
where, $acc(B_k) =  |B_k|^{-1}\sum_{k\in B_k}[y_k =\hat{y}_k] $ and $conf(B_k) = |B_k|^{-1}\sum_{k \in B_k}p(\hat{y}_k|x_k, \mathbf{w})$ and $\hat{y}_k = argmax_y p(y|x_k, \mathbf{w}).$

\subsection*{D.2 Brier Score}
Brier score \cite{brier} is strictly proper scoring function (unlike ECE) that measures the accuracy of the predicted probabilities. It is the squared error of the predicted probability $p(\hat{y}|x,\mathbf{w})$  and one-hot encoded true response, $y$ given by:

\begin{equation*}
    BS = |Y|^{-1}\sum_{y\in Y}(p(\hat{y}|x,\mathbf{w}) - \delta(\hat{y}-y))
\end{equation*}

It also has a decomposed interpretation as $BS = uncertainty - resolution + reliability$ \cite{degroot}.

\section*{E. Example Data Corruptions}
\begin{figure*}[!h]
\centering\includegraphics[scale = 0.3]{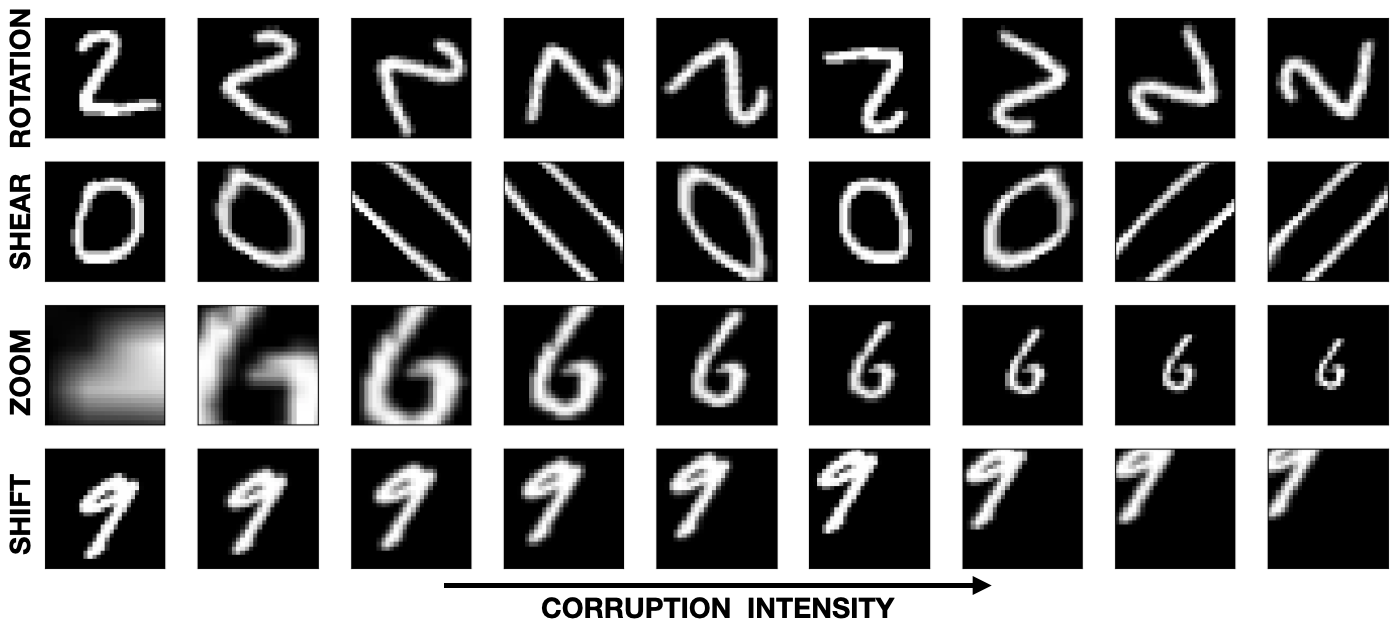}
\caption{Data corruption types at different intensities.}
\label{corr}
\end{figure*}

Some corruption types at different intensities are shown in Fig. \ref{corr} for samples of MNIST dataset.

 \section*{F. Implementation Strategy of Methods}
 \subsection*{F.1. Stochastic Variational Inference}
 For implementing the Stochastic Variational Inference algorithm \cite{blun}, we replaced each dense layer and the last convolutional layer in the network with Flipout layers. We used independent (mean-field) Gaussian distributions for prior initializations of standard deviations and implemented negative log-likelihood Bayesian optimization to obtain the posteriors using 100 epochs.\par
 For implementing SVI-LL (last layer) algorithm, we replaced only the last dense layer with a Flipout layer and implemented the same optimization procedure as before with 50 epochs.
 
  \subsection*{F.2. Monte-Carlo Dropout}
  For implementation of MC-Dropout, we introduced a dropout layer (during testing) after each convolutional block (before the next convolutional layer) and before each dense layer. We implemented 100 stochastic forward runs. The standard deviation of the results of all runs at each prediction was considered to be the uncertainty score. We varied the dropout rates between 0.01 and 0.4 and found that the best results were obtained by using a rate of 0.1 for LeNet models and 0.2 for VGG-3 model, which we used as benchmarks.\par
  
  The same implementation strategy was followed for MC-Dropout-LL (last layer), except that only a single dropout layer was introduced in the network before the last dense layer during testing. It was observed that the dropout rate of 0.2 corresponded to the best result for all models, which we used as benchmarks.
  
  \subsection*{F.3. Ensemble}
  For implemeting the Ensemble methods, we trained 10 models with random initializations for each dataset and computed the standard deviation of results as the uncertainty score.

\section*{G. Additional Result Figures}
\subsection*{G.1. KMNIST}
  \begin{figure*}[!h]
  \begin{subfigure}{0.5\linewidth}
    \centering\includegraphics[scale = 0.29]{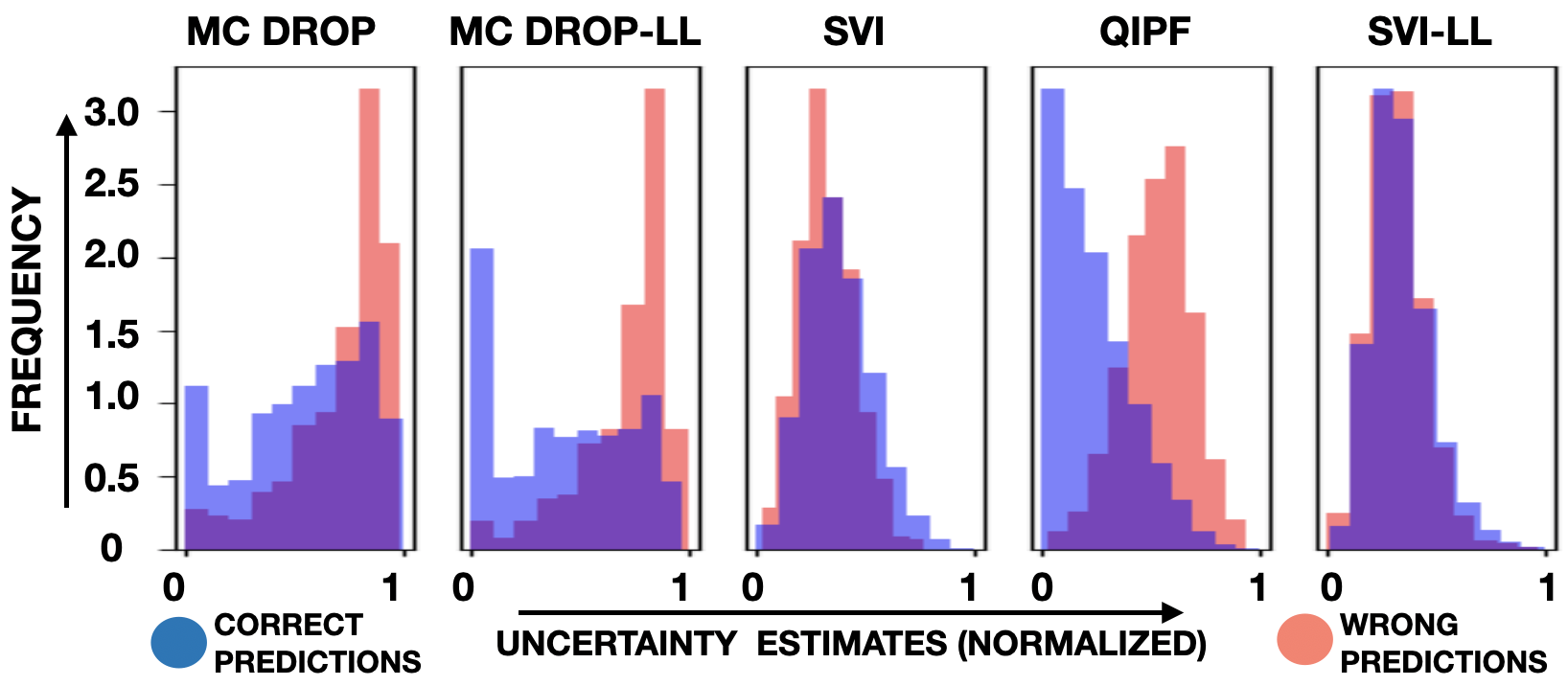}
    \caption\centering{Histogram of Uncertainty Scores}
  \end{subfigure}
  \begin{subfigure}{0.24\linewidth}
    \centering\includegraphics[scale = 0.21]{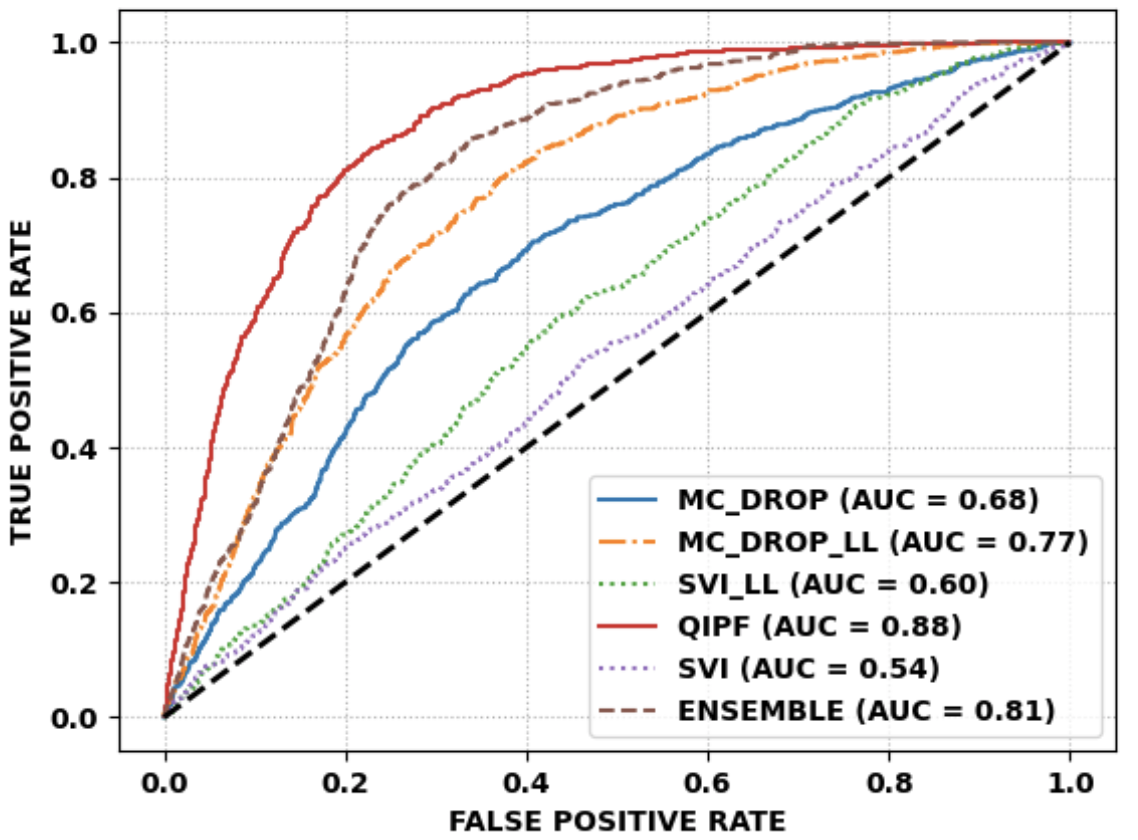}
    \caption\centering{ROC Curves}
  \end{subfigure}
  \begin{subfigure}{0.24\linewidth}
    \centering\includegraphics[scale = 0.22]{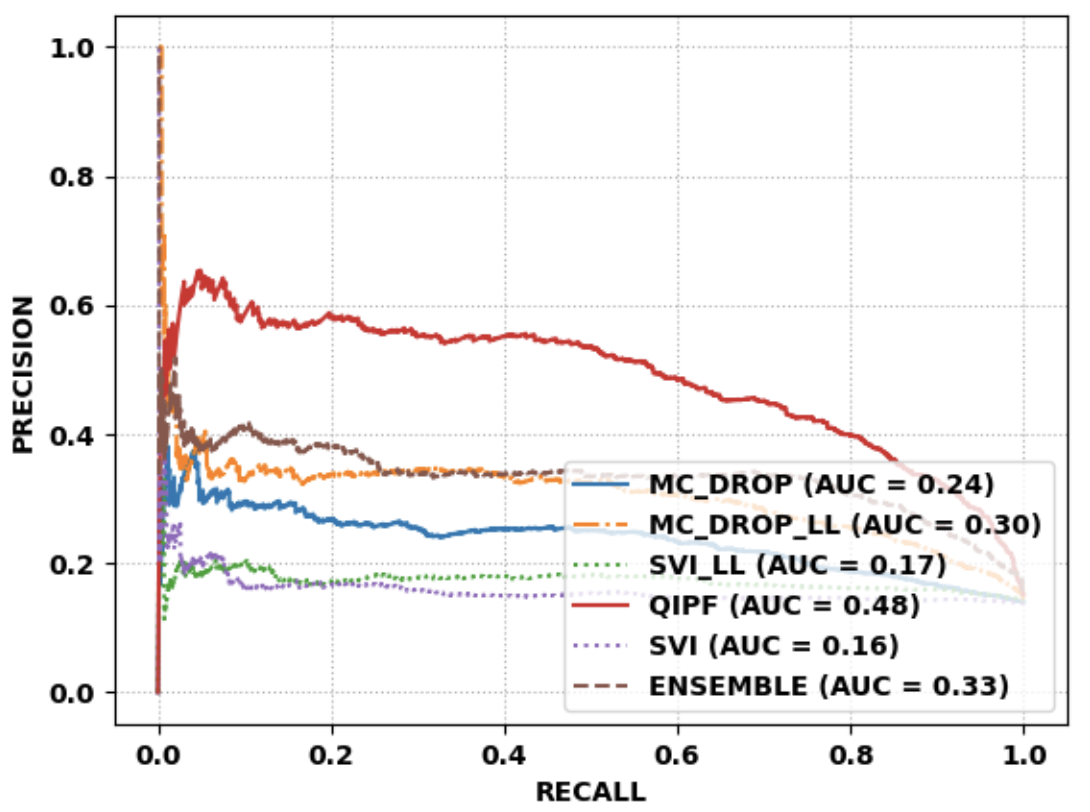}
    \caption\centering{Precision-recall curves}
  \end{subfigure}
  \caption{KMNIST Brightness Corrupted: Intensity = 40\%}
  \end{figure*}
  
     \begin{figure*}[!h]
  \begin{subfigure}{0.5\linewidth}
    \centering\includegraphics[scale = 0.29]{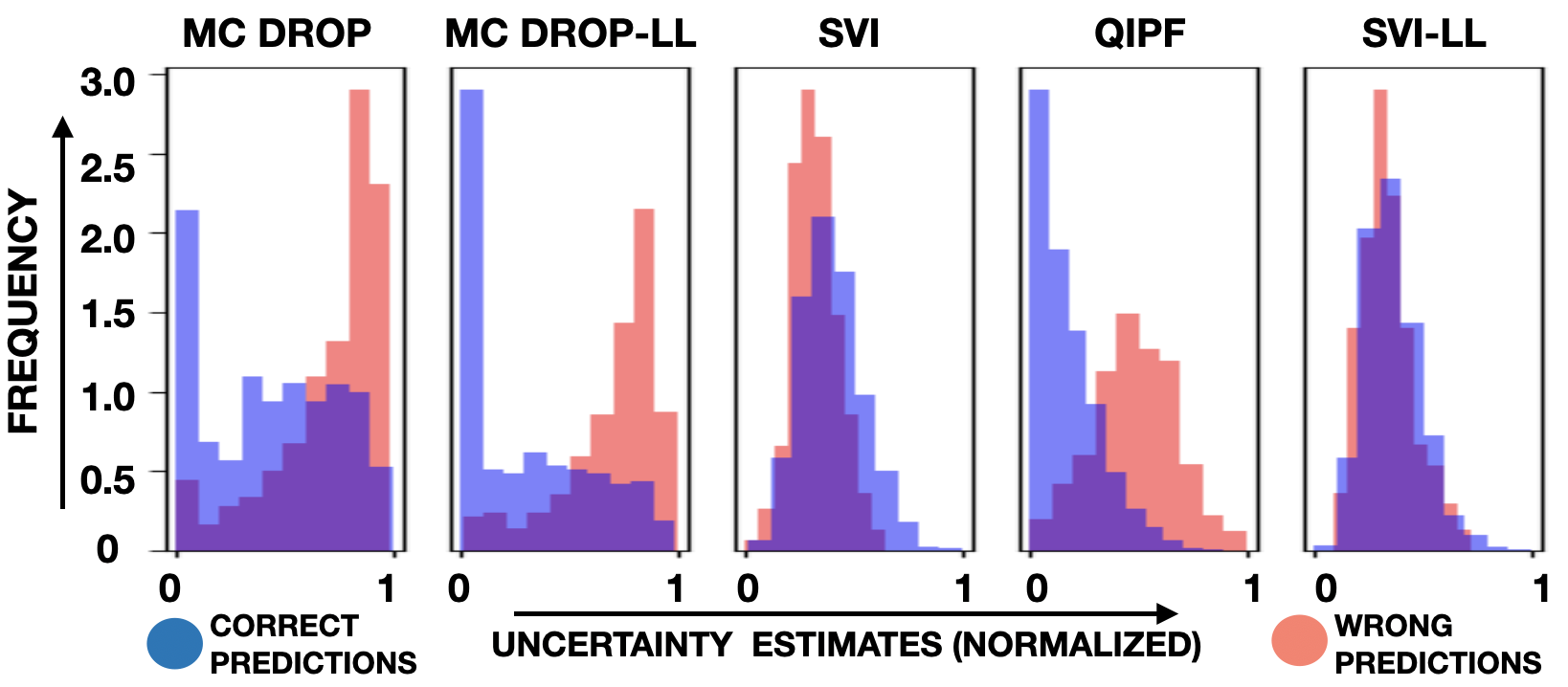}
    \caption\centering{Histogram of Uncertainty Scores}
  \end{subfigure}
  \begin{subfigure}{0.24\linewidth}
    \centering\includegraphics[scale = 0.22]{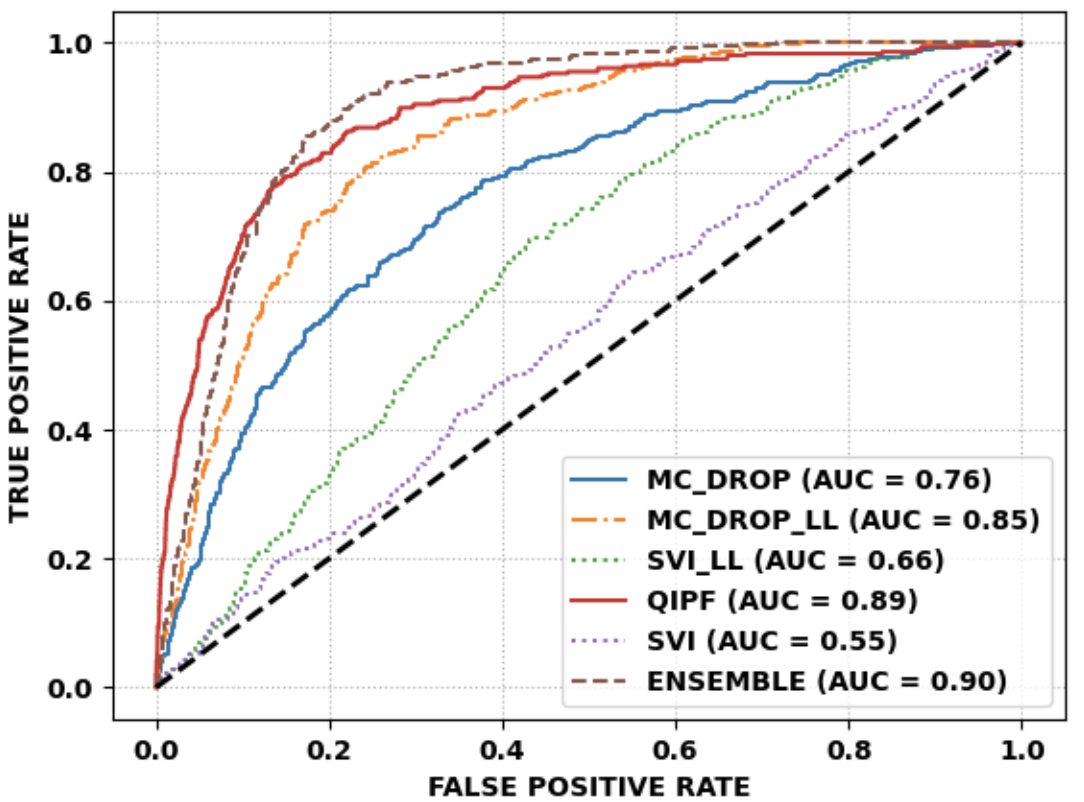}
    \caption\centering{ROC Curves}
  \end{subfigure}
  \begin{subfigure}{0.24\linewidth}
    \centering\includegraphics[scale = 0.22]{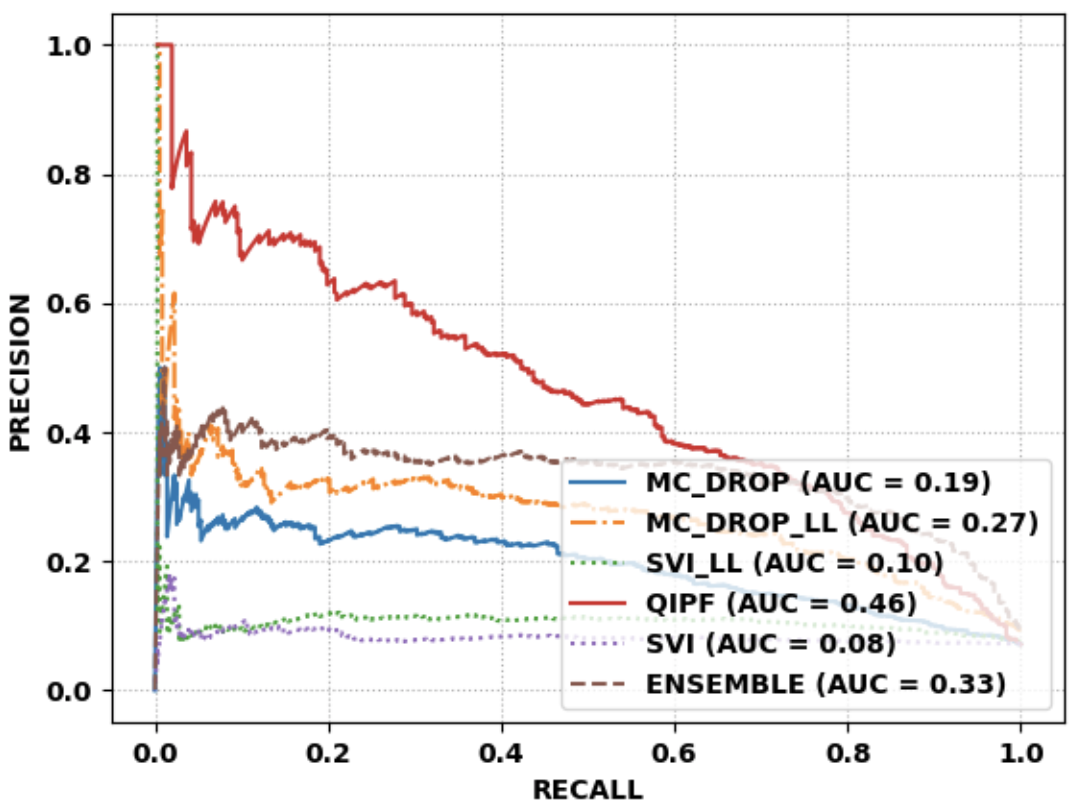}
    \caption\centering{Precision-recall curves}
  \end{subfigure}
  \caption{KMNIST Rotation Corrupted: Intensity = 10\%}
  \end{figure*}
  
      \begin{figure*}[!h]
  \begin{subfigure}{0.5\linewidth}
    \centering\includegraphics[scale = 0.29]{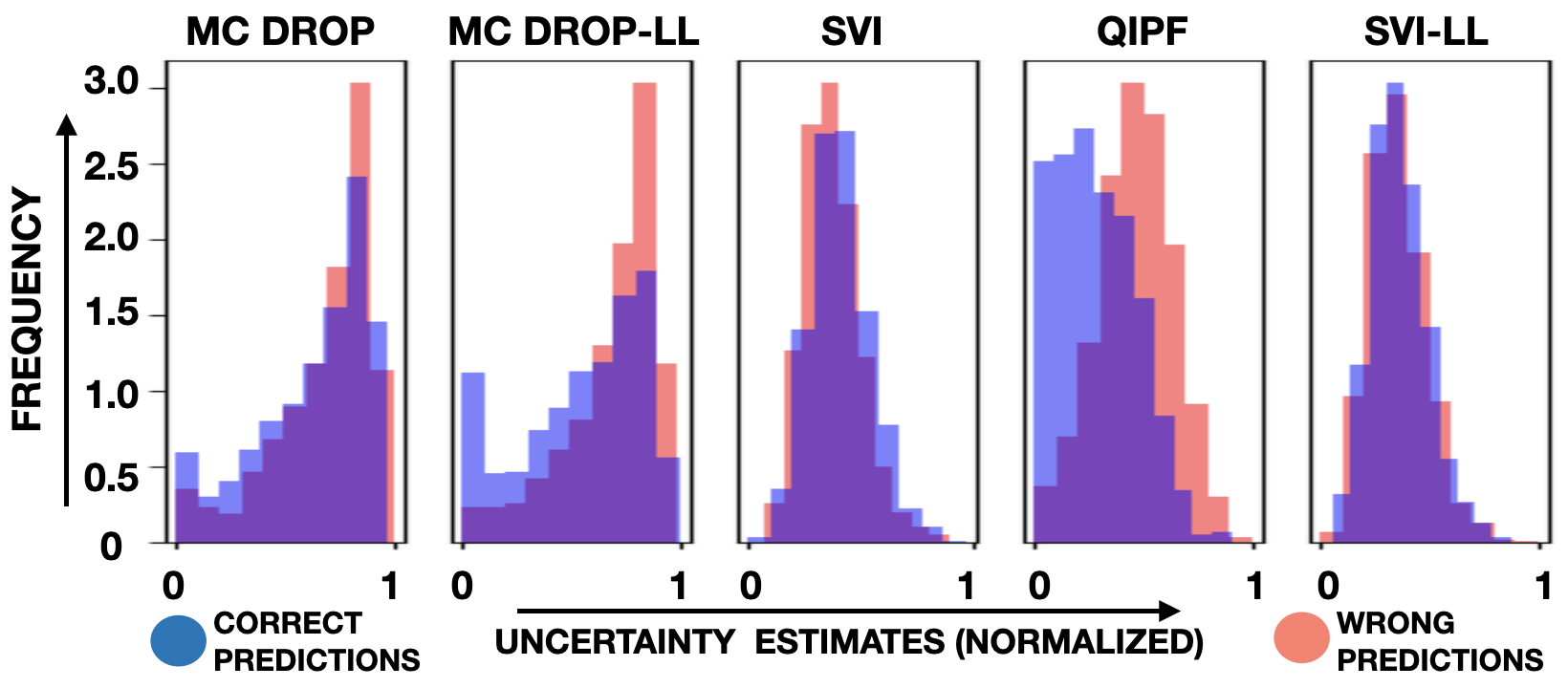}
    \caption\centering{Histogram of Uncertainty Scores}
  \end{subfigure}
  \begin{subfigure}{0.24\linewidth}
    \centering\includegraphics[scale = 0.25]{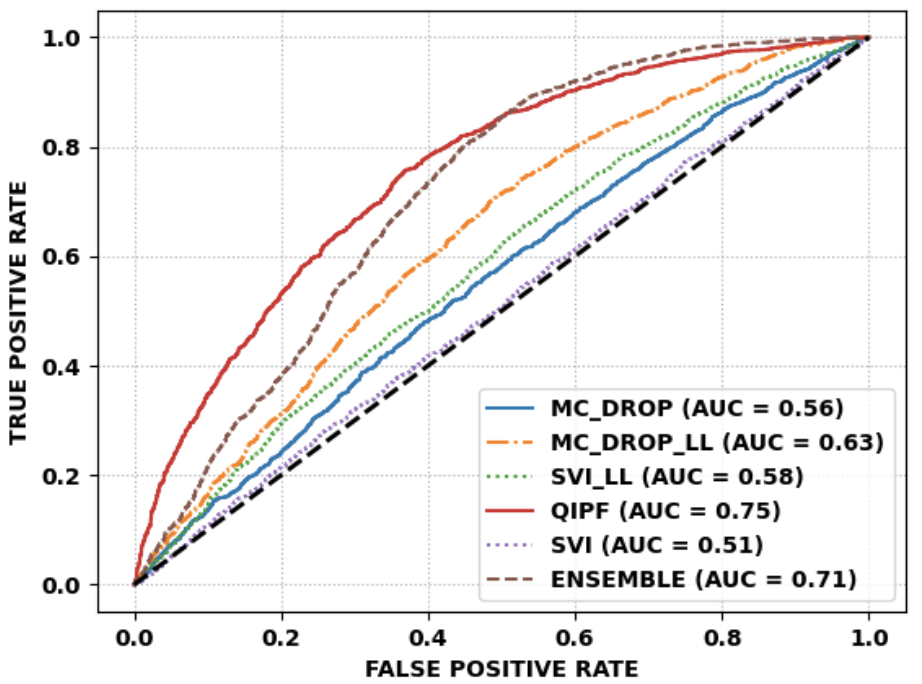}
    \caption\centering{ROC Curves}
  \end{subfigure}
  \begin{subfigure}{0.24\linewidth}
    \centering\includegraphics[scale = 0.25]{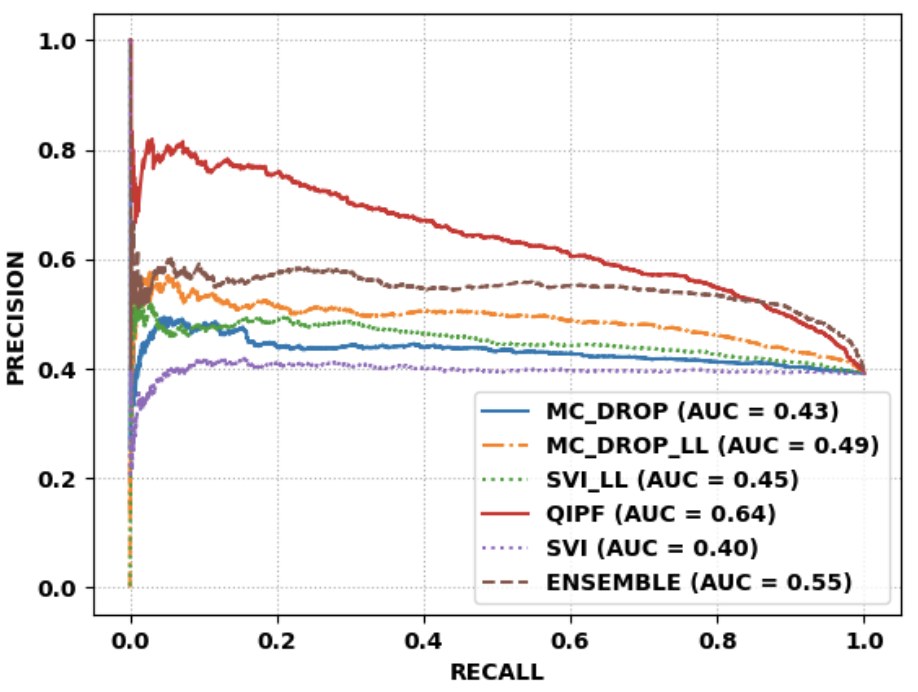}
    \caption\centering{Precision-recall curves}
  \end{subfigure}
  \caption{KMNIST Shear Corrupted: Intensity = 90\%}
  \end{figure*}
  
  \begin{figure*}
      \begin{subfigure}{0.24\linewidth}
    \centering\includegraphics[scale = 0.18]{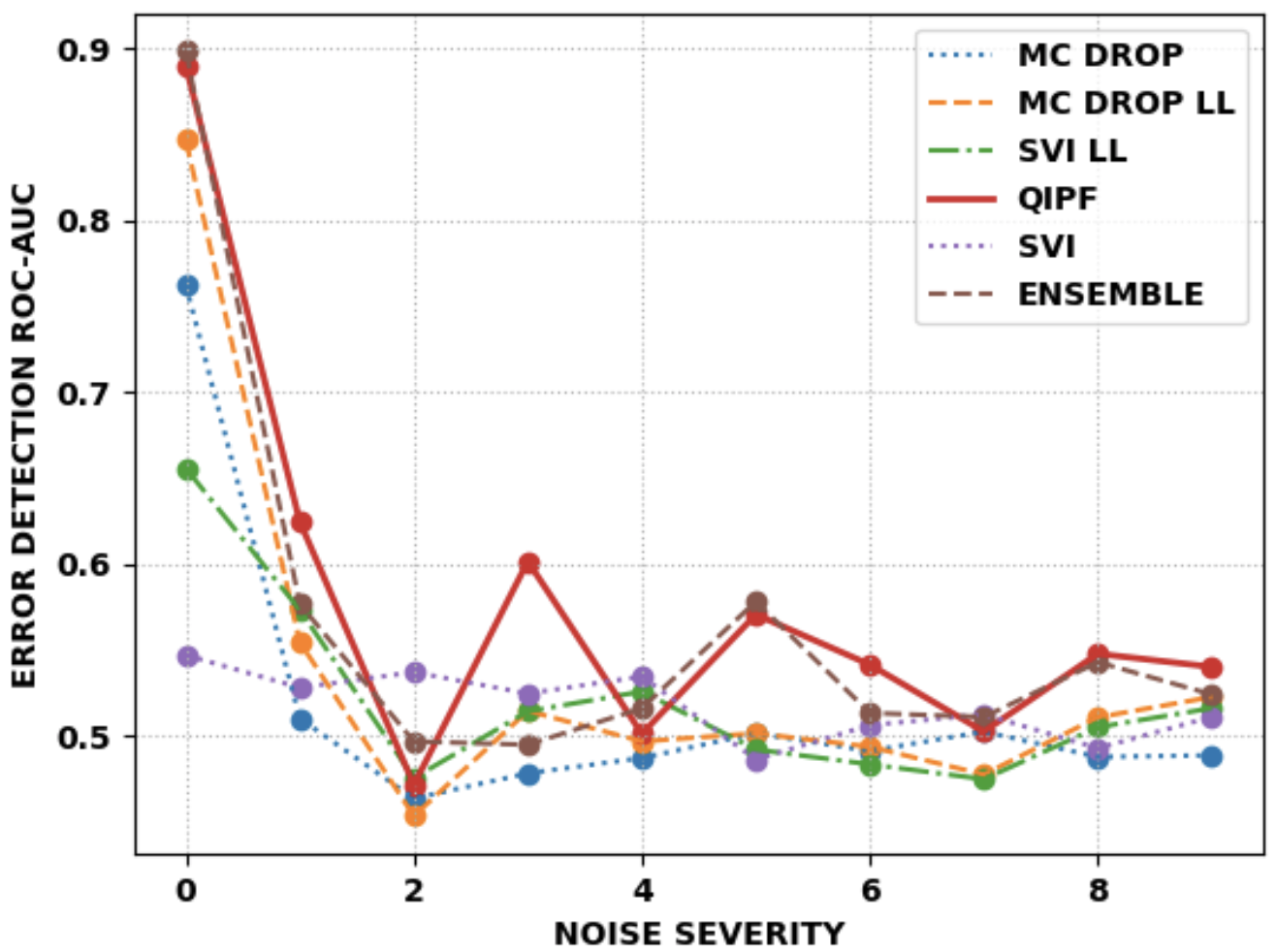}
    \caption\centering{ROC-AUC vs noise severity}
  \end{subfigure}
    \begin{subfigure}{0.24\linewidth}
    \centering\includegraphics[scale = 0.18]{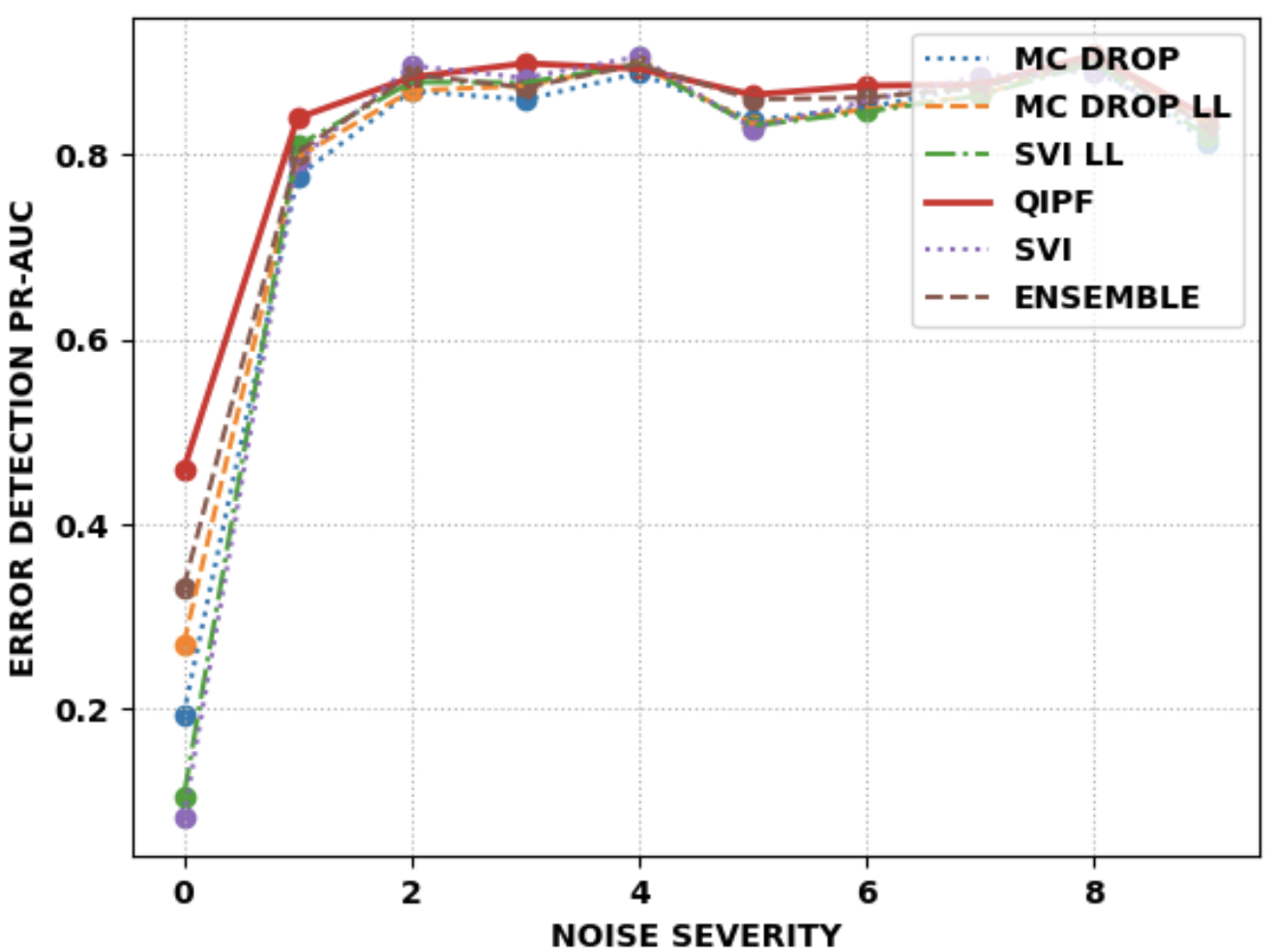}
    \caption\centering{PR-AUC vs noise severity}
  \end{subfigure}
  \begin{subfigure}{0.24\linewidth}
    \centering\includegraphics[scale = 0.18]{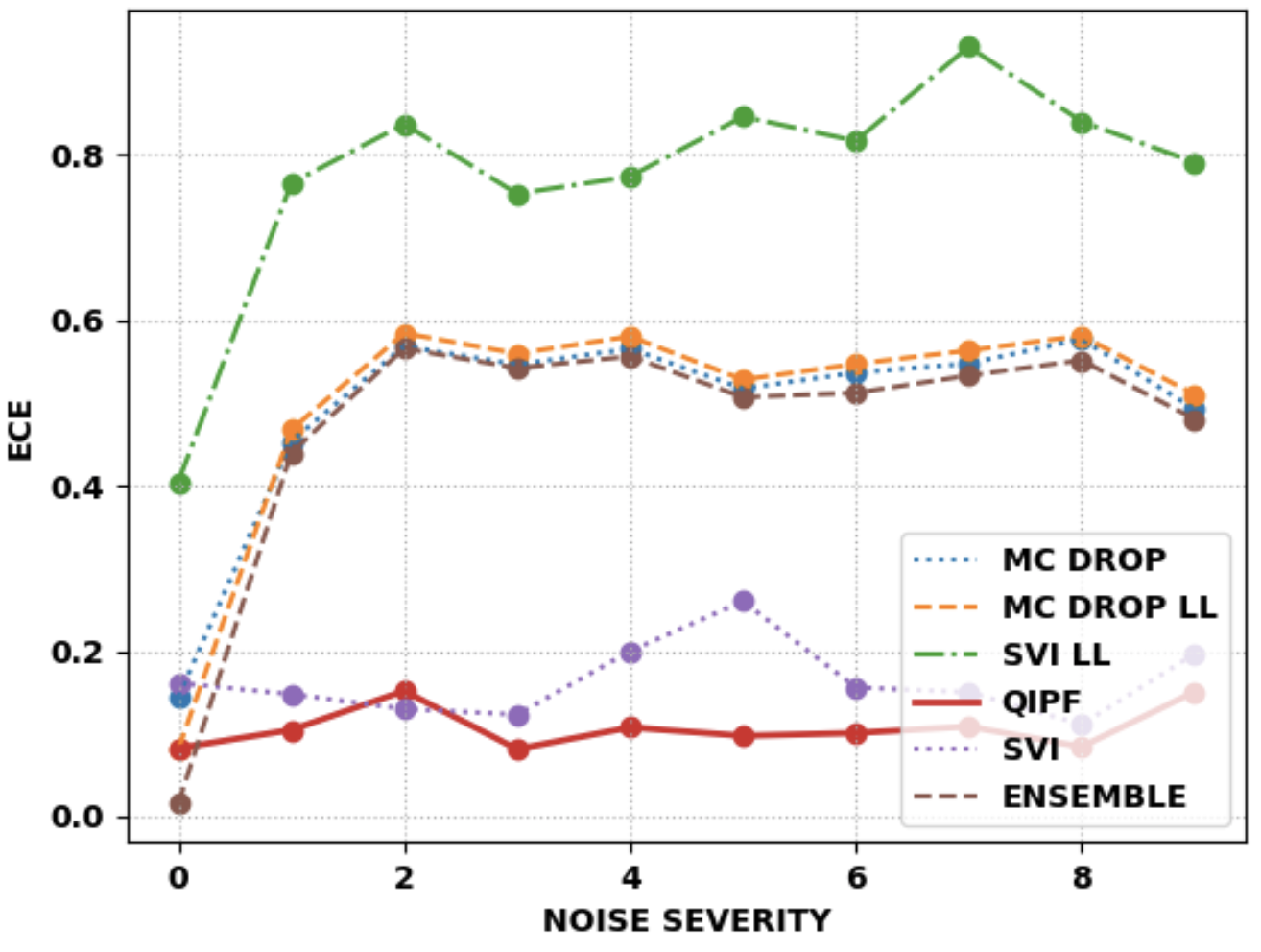}
    \caption\centering{Expected Calibration Error vs noise severity}
  \end{subfigure}
      \begin{subfigure}{0.24\linewidth}
    \centering\includegraphics[scale = 0.18]{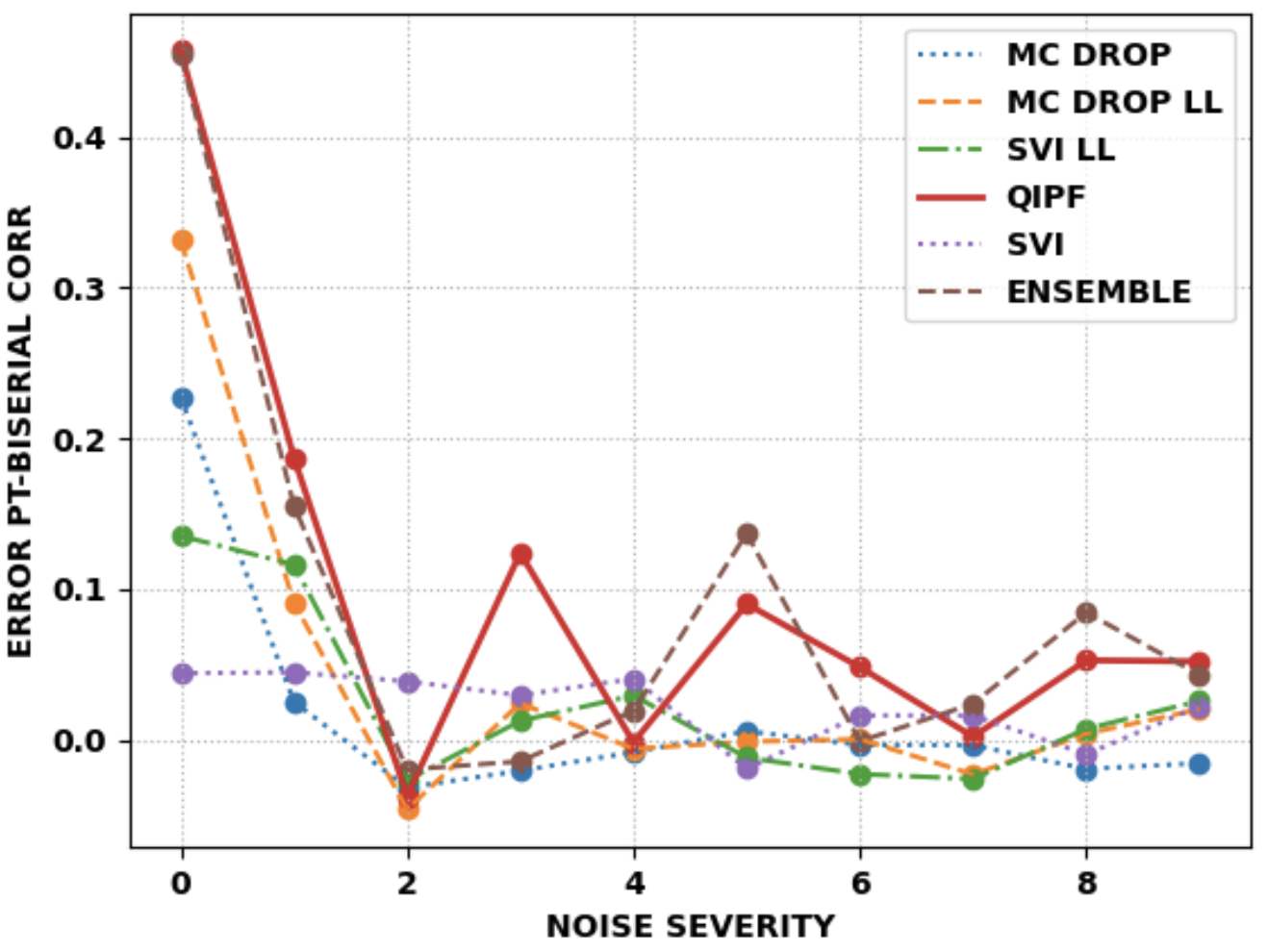}
    \caption\centering{Point-Biserial Error Correlation vs noise severity}
  \end{subfigure}
    \caption{KMNIST: Rotation Corruption (All intensities)}
  \end{figure*}
  
    \begin{figure*}
      \begin{subfigure}{0.24\linewidth}
    \centering\includegraphics[scale = 0.18]{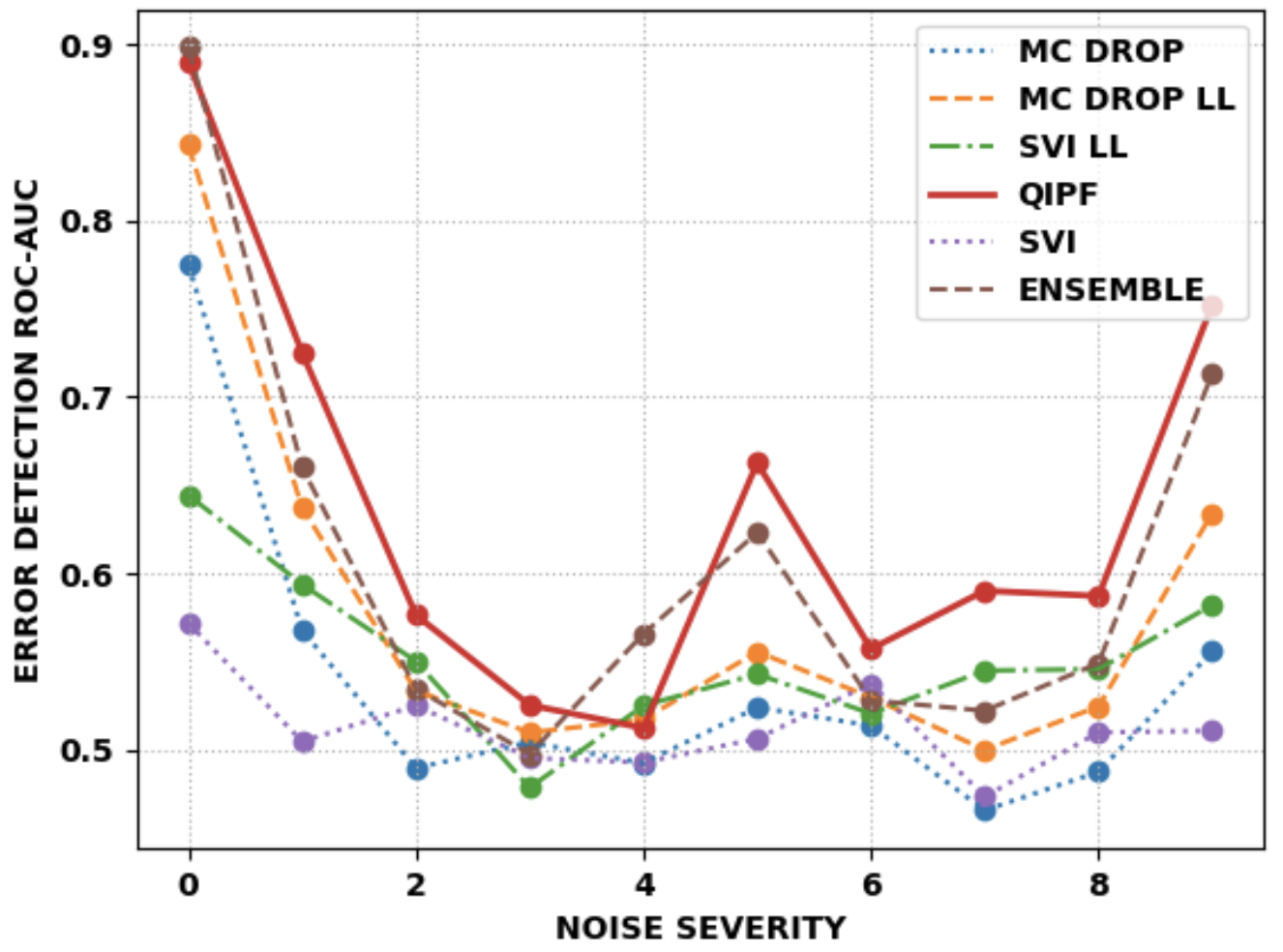}
    \caption\centering{ROC-AUC vs noise severity}
  \end{subfigure}
    \begin{subfigure}{0.24\linewidth}
    \centering\includegraphics[scale = 0.18]{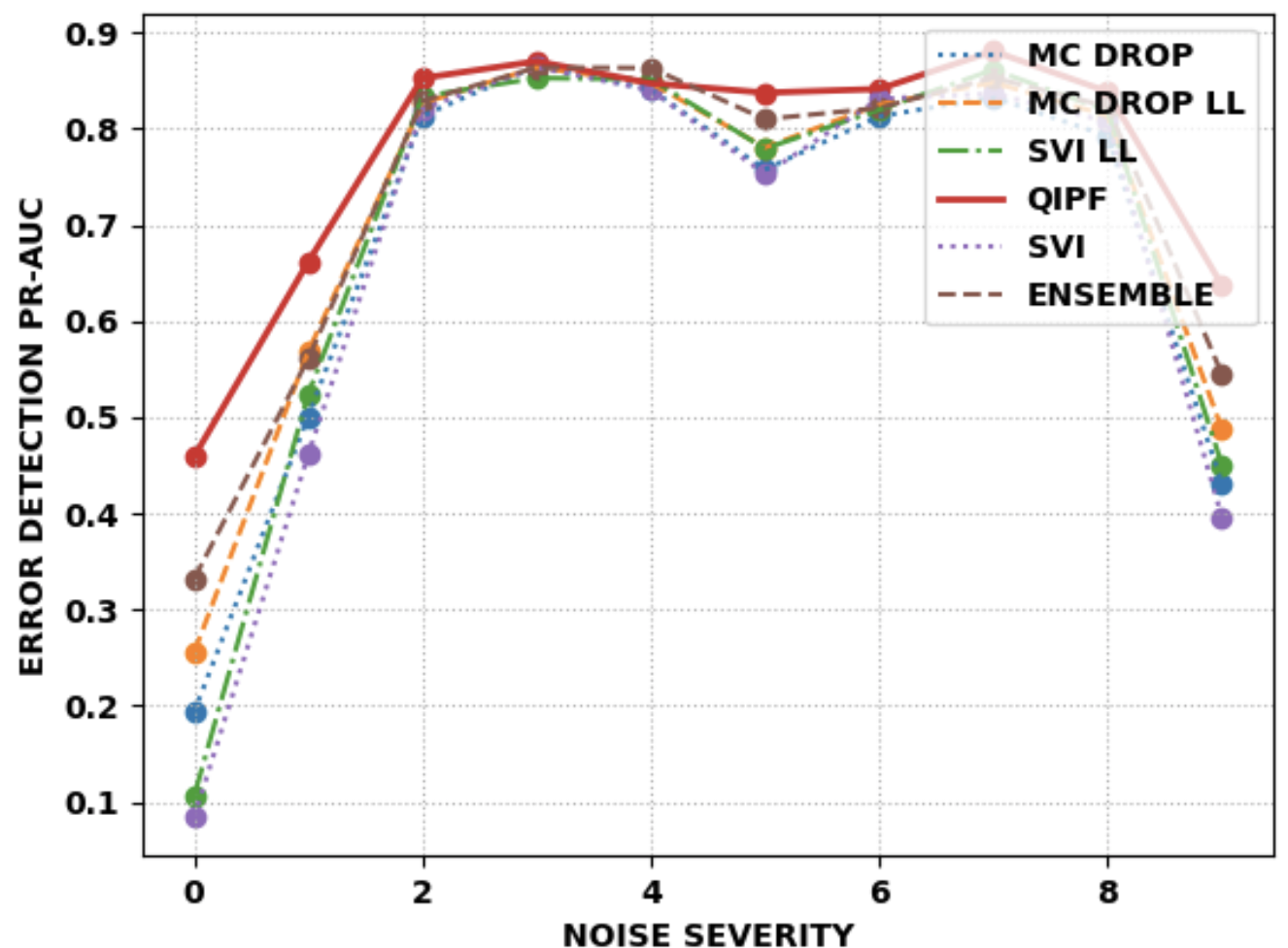}
    \caption\centering{PR-AUC vs noise severity}
  \end{subfigure}
  \begin{subfigure}{0.24\linewidth}
    \centering\includegraphics[scale = 0.18]{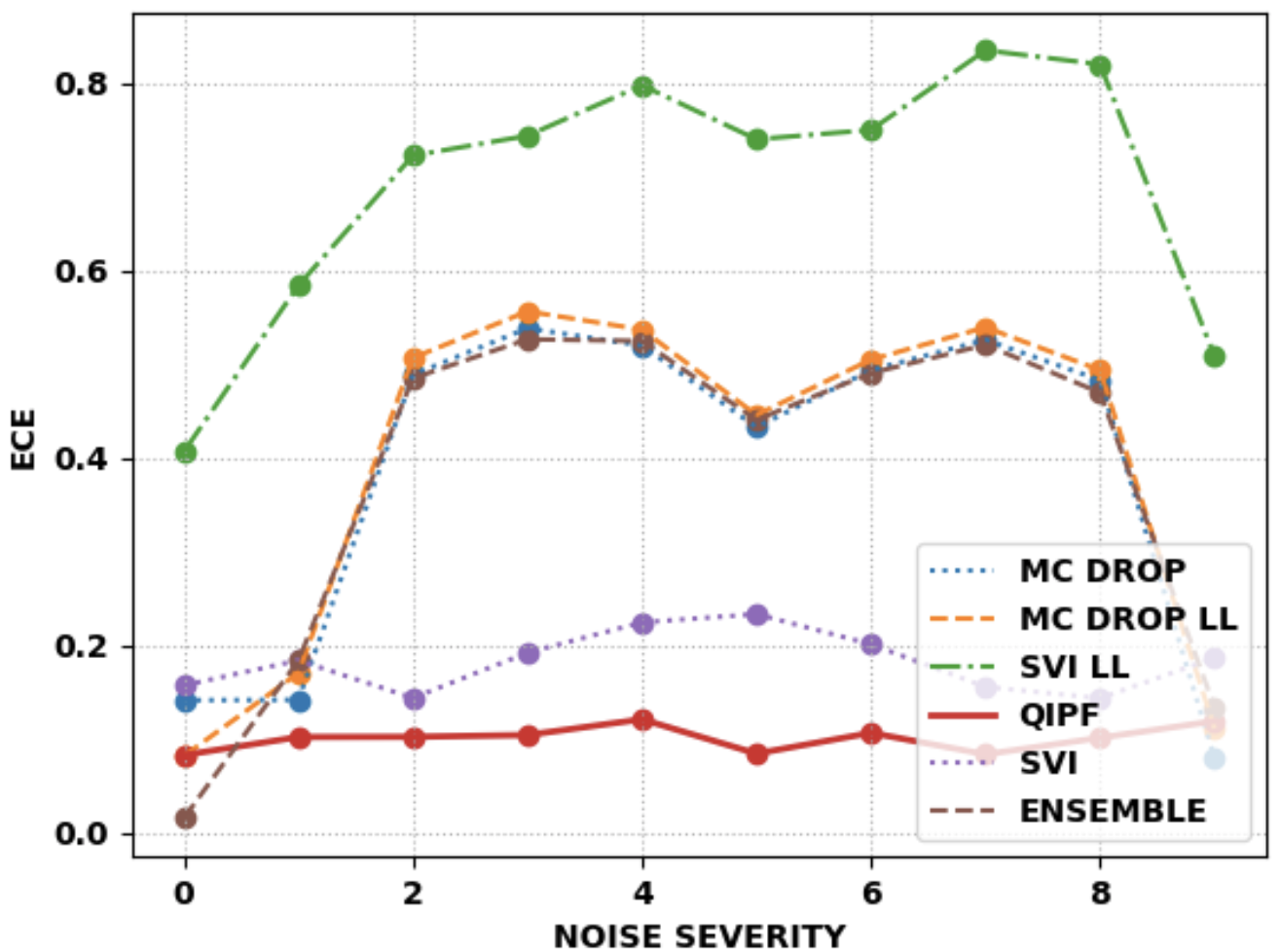}
    \caption\centering{Expected Calibration Error vs noise severity}
  \end{subfigure}
      \begin{subfigure}{0.24\linewidth}
    \centering\includegraphics[scale = 0.18]{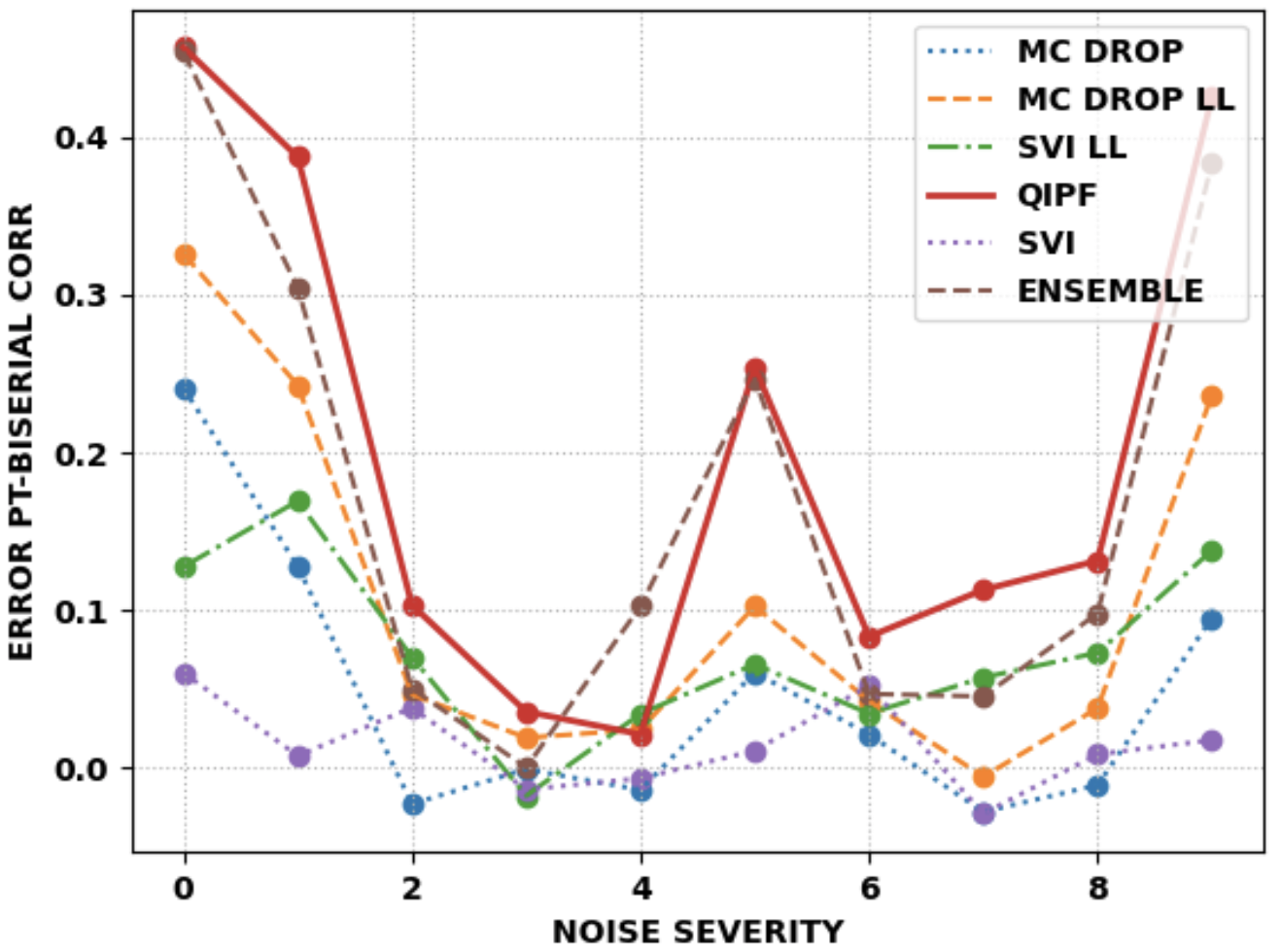}
    \caption\centering{Point-Biserial Error Correlation vs noise severity}
  \end{subfigure}
    \caption{KMNIST: Shear Corruption (All intensities)}
  \end{figure*}

      \begin{figure*}
      \begin{subfigure}{0.24\linewidth}
    \centering\includegraphics[scale = 0.18]{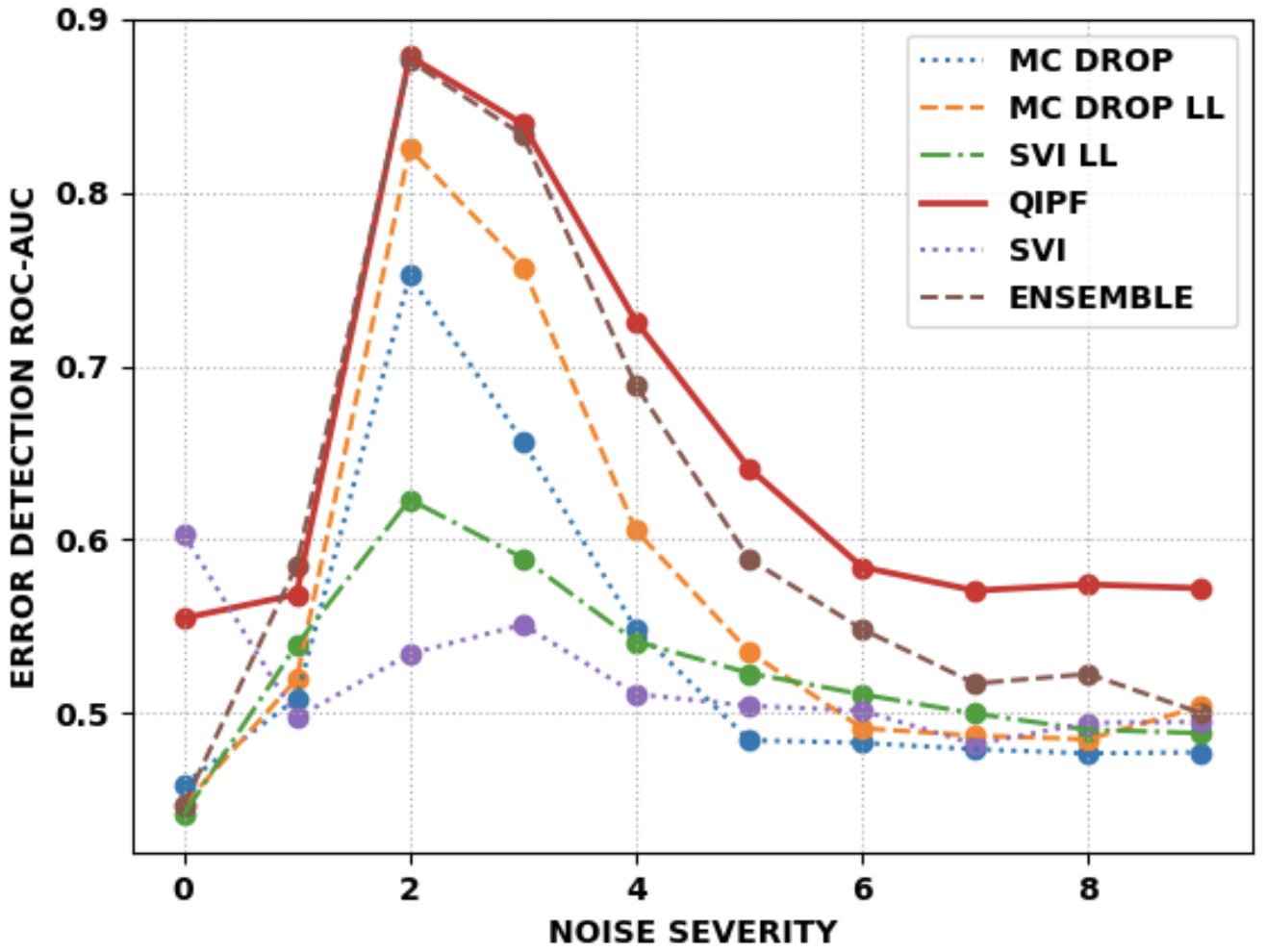}
    \caption\centering{ROC-AUC vs noise severity}
  \end{subfigure}
    \begin{subfigure}{0.24\linewidth}
    \centering\includegraphics[scale = 0.18]{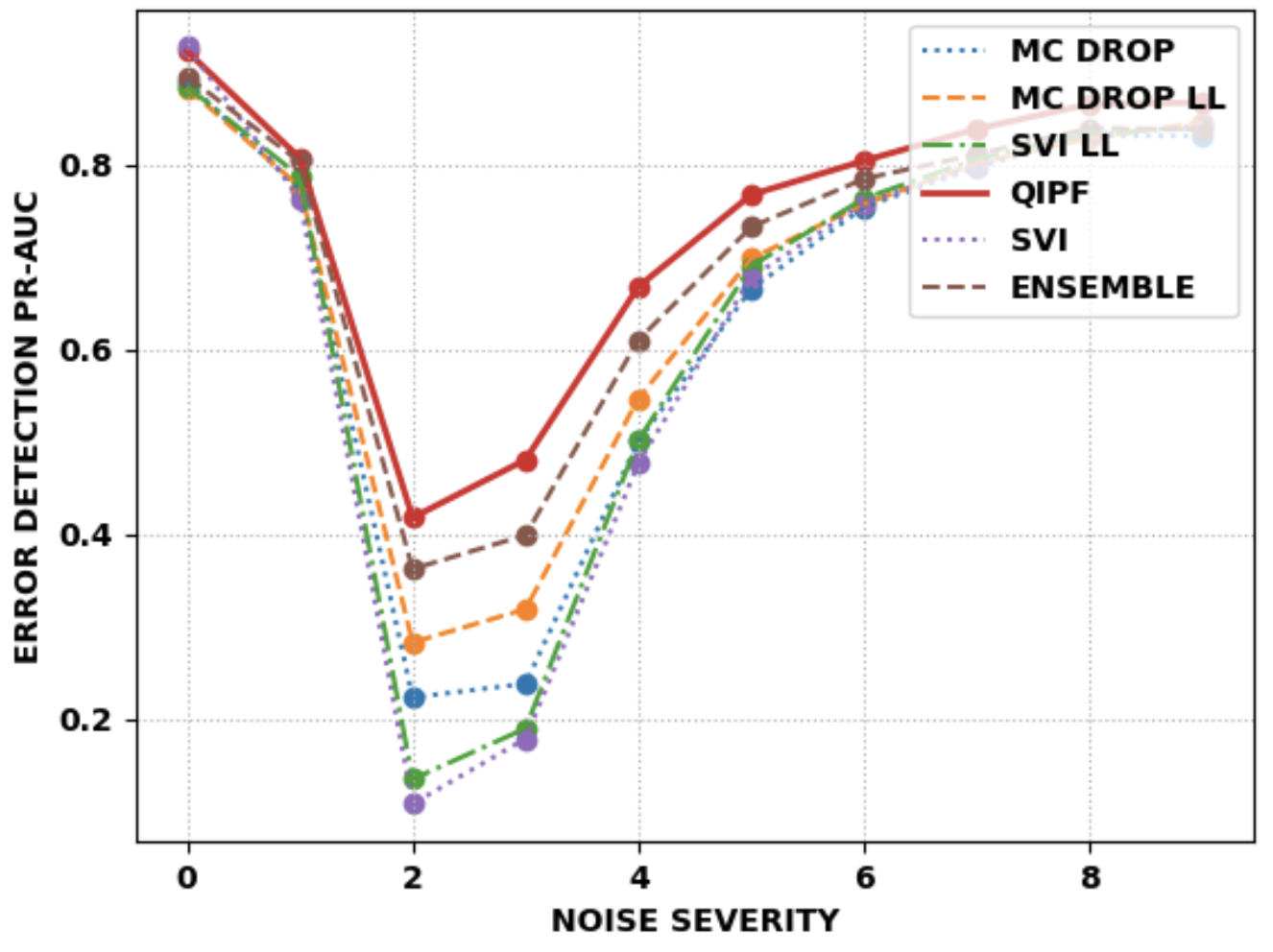}
    \caption\centering{PR-AUC vs noise severity}
  \end{subfigure}
  \begin{subfigure}{0.24\linewidth}
    \centering\includegraphics[scale = 0.18]{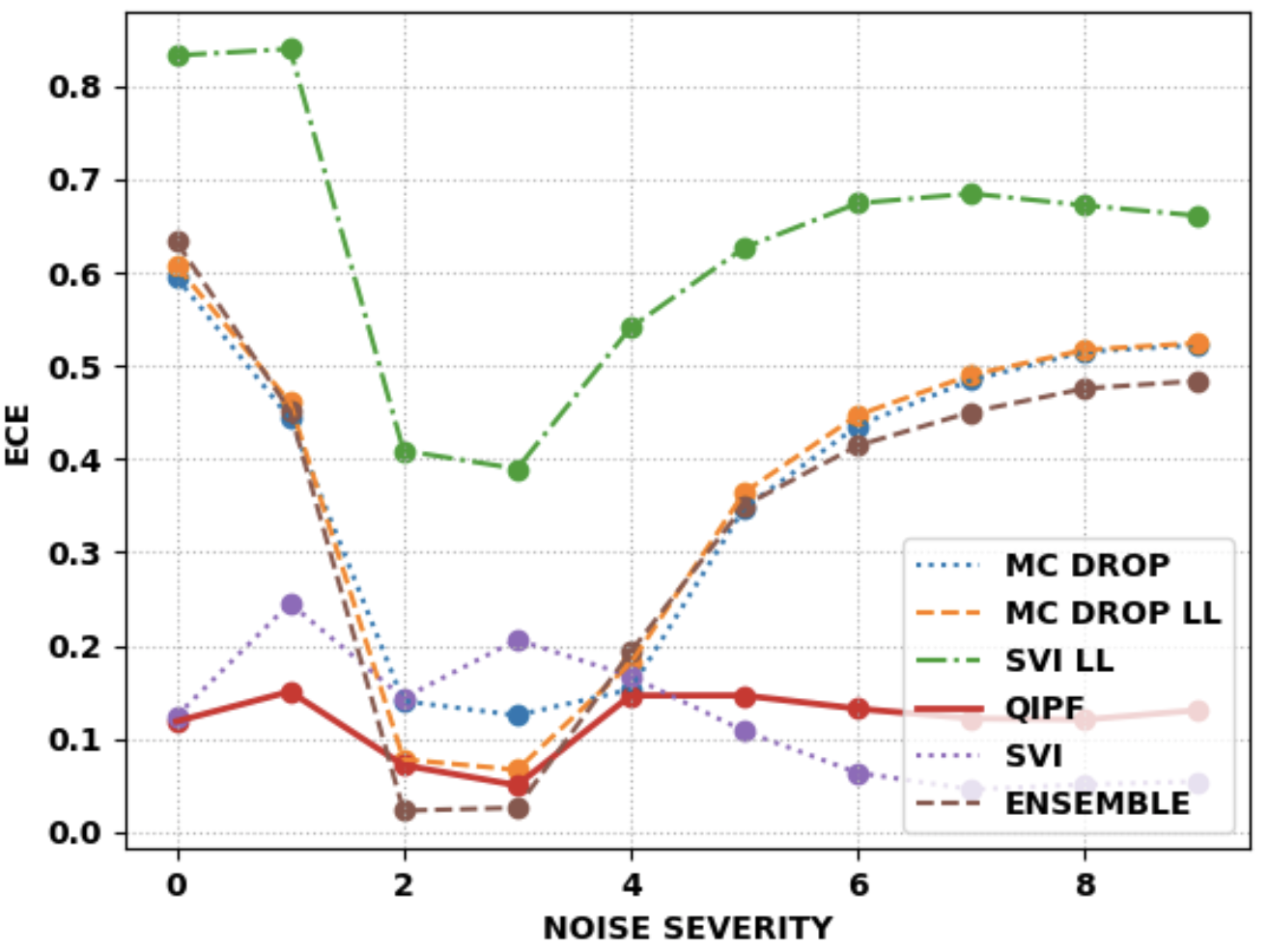}
    \caption\centering{Expected Calibration Error vs noise severity}
  \end{subfigure}
      \begin{subfigure}{0.24\linewidth}
    \centering\includegraphics[scale = 0.18]{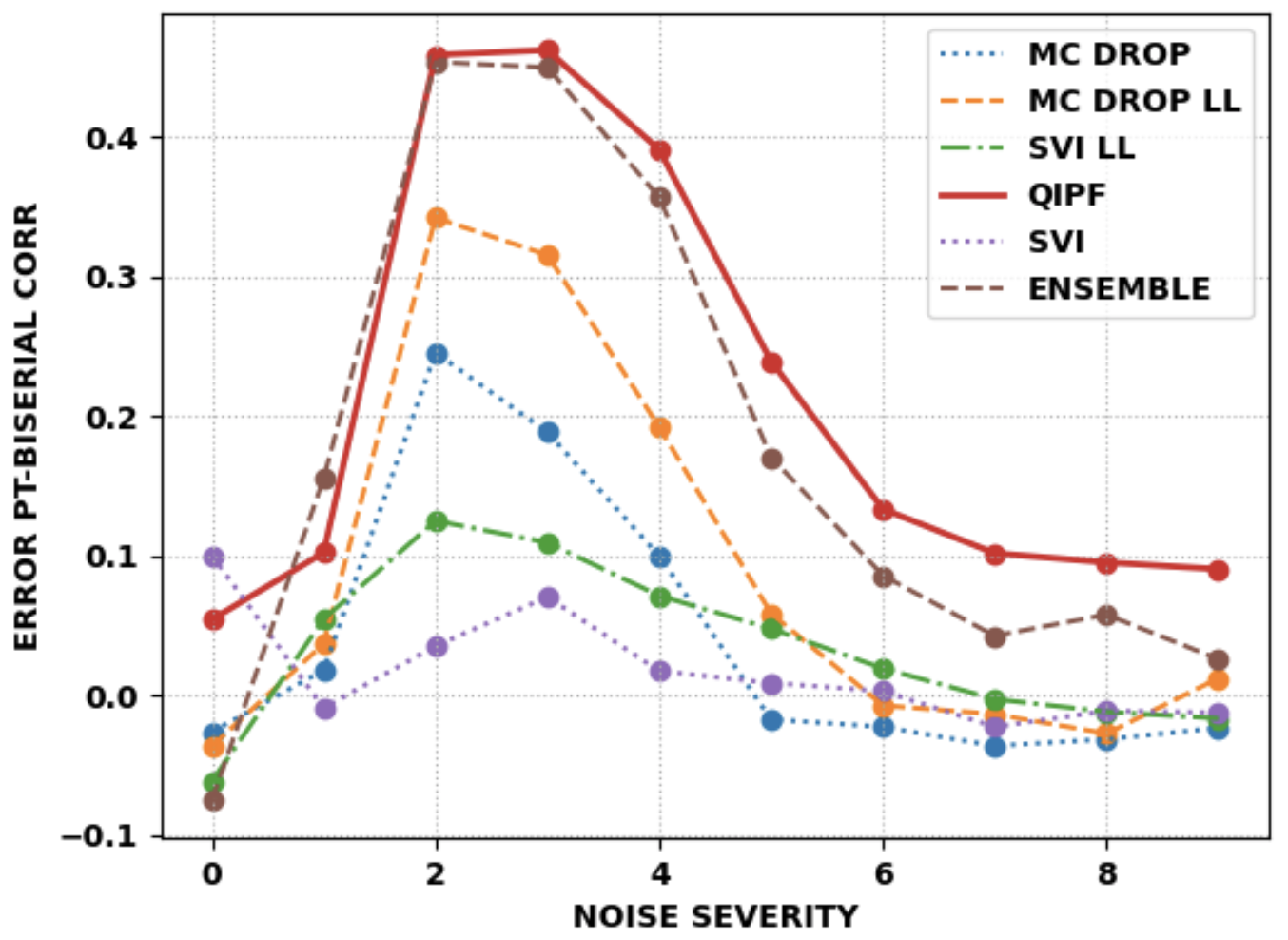}
    \caption\centering{Point-Biserial Error Correlation vs noise severity}
  \end{subfigure}
    \caption{KMNIST: Zoom Corruption (All intensities)}
  \end{figure*}



\clearpage
\subsection*{G.2. CIFAR-10 (ResNet-18)}
 \begin{figure*}[!h]
  \begin{subfigure}{0.5\linewidth}
    \centering\includegraphics[scale = 0.5]{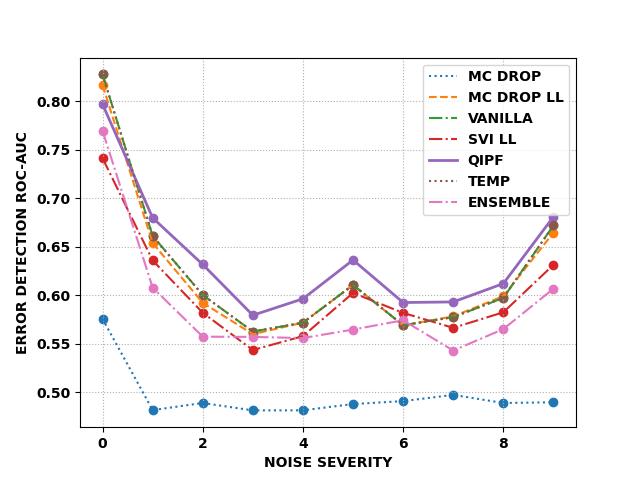}
    \caption\centering{ROC-AUC vs Noise Severity}
  \end{subfigure}
  \begin{subfigure}{0.5\linewidth}
    \centering\includegraphics[scale = 0.5]{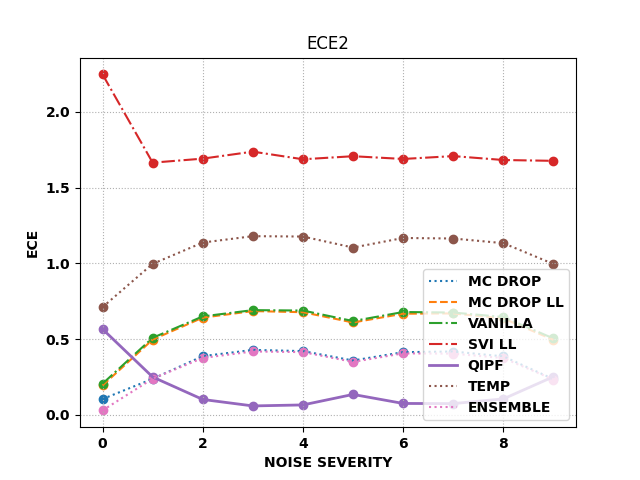}
    \caption\centering{Expected Calibration Error vs Noise Severity}
  \end{subfigure}
  \begin{subfigure}{0.5\linewidth}
    \centering\includegraphics[scale = 0.5]{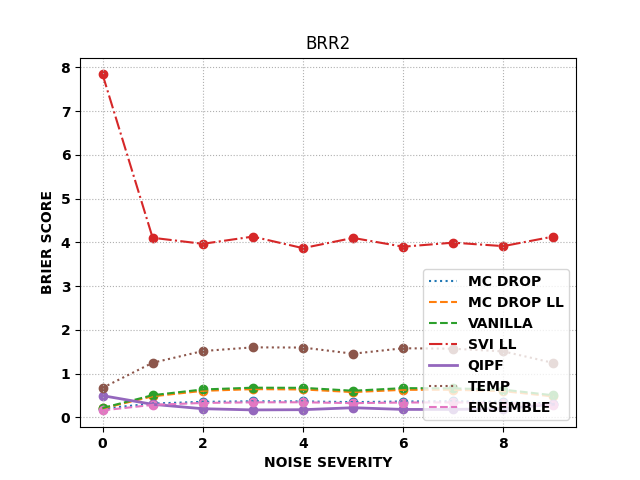}
    \caption\centering{Brier Score vs Noise Severity}
  \end{subfigure}
    \begin{subfigure}{0.5\linewidth}
    \centering\includegraphics[scale = 0.5]{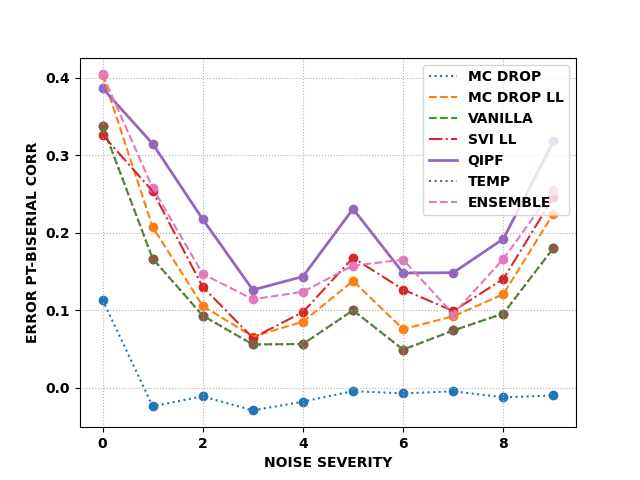}
    \caption\centering{Point Biserial Correlation with error vs Noise Severity}
  \end{subfigure}
  \label{res}
  \caption{CIFAR 10: ResNet-18 - rotation corruption (all intensities)}
  \end{figure*}

    \clearpage
    
\section*{H. Error Correlation Results}
We evaluate how correlated the uncertainty estimates of the different methods are with respect to the model prediction error by using point-biserial correlation coefficient which measures correlation between a continuous variable (uncertainty quantified by the different techniques) and a dichotomous variable (model prediction error). We also evaluate Spearman coefficient. results are shown in Table \ref{err}.
\begin{table*}[!h]
\centering
\resizebox{\textwidth}{!}{%
\begin{tabular}{|l|l|llllll|llllll|}
\hline
\multicolumn{1}{|c|}{\multirow{2}{*}{\begin{tabular}[c]{@{}c@{}}CORRUPTION\\ TYPE\end{tabular}}} & \multirow{2}{*}{DATASET}                                    & \multicolumn{6}{c|}{POINT BISERIAL CORRELATION}                                                                                                                                                                                                                         & \multicolumn{6}{c|}{SPEARMAN CORRELATION}                                                                                                                                                                                                                          \\ \cline{3-14} 
\multicolumn{1}{|c|}{}                                                                           &                                                             & \multicolumn{1}{c}{MC-DROP}         & \multicolumn{1}{c}{MC-DROP-LL}      & \multicolumn{1}{c}{SVI}             & \multicolumn{1}{c}{SVI-LL}                   & \multicolumn{1}{c}{ENSEMBLE}        & \multicolumn{1}{c|}{QIPF}                     & \multicolumn{1}{c}{MC-DROP}         & \multicolumn{1}{c}{MC-DROP-LL}      & \multicolumn{1}{c}{SVI}             & \multicolumn{1}{c}{SVI-LL}                   & \multicolumn{1}{c}{ENSEMBLE}        & \multicolumn{1}{c|}{QIPF}                     \\ \hline
\multirow{4}{*}{ROTATION}                                                                        & MNIST                                                       & \multicolumn{1}{c}{0.16 $\pm$ 0.06} & \multicolumn{1}{c}{0.29 $\pm$ 0.06} & \multicolumn{1}{c}{0.08 $\pm$ 0.05} & \multicolumn{1}{c}{0.24 $\pm$ 0.05}          & \multicolumn{1}{c}{{\color{blue}{\textbf{0.31 $\pm$ 0.06}}}} & \multicolumn{1}{c|}{\color{red}{\textbf{0.34 $\pm$ 0.08}}} & \multicolumn{1}{c}{0.16 $\pm$ 0.06} & \multicolumn{1}{c}{0.26 $\pm$ 0.07} & \multicolumn{1}{c}{0.06 $\pm$ 0.05} & \multicolumn{1}{c}{0.22 $\pm$ 0.05}          & \multicolumn{1}{c}{\color{blue}{\textbf{0.27 $\pm$ 0.06}}} & \multicolumn{1}{c|}{\color{red}{\textbf{0.33 $\pm$ 0.07}}} \\
                                                                                                 & K-MNIST                                                     & 0.00 $\pm$ 0.02                     & 0.03 $\pm$ 0.08                     & 0.03 $\pm$ 0.02                     & 0.06 $\pm$ 0.10                              & \color{blue}{\textbf{0.09 $\pm$ 0.15}}                     & \color{red}{\textbf{0.11 $\pm$ 0.08}}                      & 0.01 $\pm$ 0.02                     & 0.03 $\pm$ 0.08                     & 0.03 $\pm$ 0.02                     & 0.06 $\pm$ 0.10                              & \color{blue}{\textbf{0.08 $\pm$ 0.13}}                     & \color{red}{\textbf{0.10 $\pm$ 0.15}}                      \\
                                                                                                 & \begin{tabular}[c]{@{}l@{}}CIFAR-10\\ (VGG-3)\end{tabular}  & \multicolumn{1}{c}{0.13 $\pm$ 0.09} & \multicolumn{1}{c}{0.19 $\pm$ 0.11} & \multicolumn{1}{c}{0.06 $\pm$ 0.02} & \multicolumn{1}{c}{\color{blue}{\textbf{0.21 $\pm$ 0.05}}}          & \multicolumn{1}{c}{0.12 $\pm$ 0.08} & \multicolumn{1}{c|}{\color{red}{\textbf{0.24 $\pm$ 0.11}}} & \multicolumn{1}{c}{0.11 $\pm$ 0.02} & \multicolumn{1}{c}{0.17 $\pm$ 0.10} & \multicolumn{1}{c}{0.06 $\pm$ 0.02} & \multicolumn{1}{c}{\color{blue}{\textbf{0.20 $\pm$ 0.06}}}          & \multicolumn{1}{c}{0.10 $\pm$ 0.08} & \multicolumn{1}{c|}{\color{red}{\textbf{0.25 $\pm$ 0.08}}} \\ 
                                                                                                 & \begin{tabular}[c]{@{}l@{}}CIFAR-10\\ (RESNET-18)\end{tabular}  & \multicolumn{1}{c}{0.00 $\pm$ 0.03} & \multicolumn{1}{c}{0.15 $\pm$ 0.09} & \multicolumn{1}{c}{0.09 $\pm$ 0.03} & \multicolumn{1}{c}{0.16 $\pm$ 0.08} & \multicolumn{1}{c}{\color{blue}{\textbf{0.18 $\pm$ 0.08}}} & \multicolumn{1}{c|}{\color{red}{\textbf{0.22 $\pm$ 0.08}}}          & \multicolumn{1}{c}{0.00 $\pm$ 0.04} & \multicolumn{1}{c}{\color{blue}{\textbf{0.18 $\pm$ 0.11}}} & \multicolumn{1}{c}{0.05 $\pm$ 0.03} & \multicolumn{1}{c}{0.16 $\pm$ 0.08} & \multicolumn{1}{c}{0.14 $\pm$ 0.09} & \multicolumn{1}{c|}{\color{red}{\textbf{0.22 $\pm$ 0.10}}}          \\ \hline
\multirow{4}{*}{SHEAR}                                                                           & MNIST                                                       & 0.16 $\pm$ 0.10                     & 0.26 $\pm$ 0.14                     & 0.08 $\pm$ 0.05                     & 0.17 $\pm$ 0.10                              & \color{blue}{\textbf{0.26 $\pm$ 0.11}}                     & \color{red}{\textbf{0.28 $\pm$ 0.11}}                      & 0.15 $\pm$ 0.10                     & 0.23 $\pm$ 0.12                     & 0.07 $\pm$ 0.04                     & 0.17 $\pm$ 0.10                              & \color{blue}{\textbf{0.24 $\pm$ 0.11}}                     & \color{red}{\textbf{0.27 $\pm$ 0.10}}                      \\
                                                                                                 & K-MNIST                                                     & 0.00 $\pm$ 0.03                     & 0.06 $\pm$ 0.08                     & 0.01 $\pm$ 0.02                     & 0.13 $\pm$ 0.08                              & \color{blue}{\textbf{0.17 $\pm$ 0.16}}                     & \color{red}{\textbf{0.20 $\pm$ 0.15}}                      & 0.00 $\pm$ 0.10                     & 0.06 $\pm$ 0.08                     & 0.01 $\pm$ 0.02                     & 0.12 $\pm$ 0.08                              & \color{blue}{\textbf{0.14 $\pm$ 0.14}}                     & \color{red}{\textbf{0.18 $\pm$ 0.13}}                      \\
                                                                                                 & \begin{tabular}[c]{@{}l@{}}CIFAR-10\\ (VGG-3)\end{tabular}  & 0.20 $\pm$ 0.10                     & \color{blue}{\textbf{0.25 $\pm$ 0.11}}                     & 0.10 $\pm$ 0.04                     & 0.24 $\pm$ 0.06                              & 0.20 $\pm$ 0.08                     & \color{red}{\textbf{0.25 $\pm$ 0.12}}                      & 0.18 $\pm$ 0.10                     & 0.23 $\pm$ 0.11                     & 0.09 $\pm$ 0.03                     & \color{blue}{\textbf{0.24 $\pm$ 0.06}}                              & 0.17 $\pm$ 0.09                      & \color{red}{\textbf{0.28 $\pm$ 0.06}} \\                      
                                                                                                & \begin{tabular}[c]{@{}l@{}}CIFAR-10 \\ (RESNET-18)\end{tabular} & 0.03 $\pm$ 0.03                     & 0.21 $\pm$ 0.11                     & \multicolumn{1}{c}{0.12 $\pm$ 0.03}                     & 0.23 $\pm$ 0.06                              & \color{blue}{\textbf{0.26 $\pm$ 0.09}}                     & \color{red}{\textbf{0.28 $\pm$ 0.09}}                      & 0.04 $\pm$ 0.04                     & \color{blue}{\textbf{0.26 $\pm$ 0.13}}                     & \multicolumn{1}{c}{0.10 $\pm$ 0.02}                     & 0.22 $\pm$ 0.08                              & 0.21 $\pm$ 0.10                   &  \color{red}{\textbf{0.28 $\pm$ 0.10}} 
                                                                                                \\ \hline
\multirow{4}{*}{ZOOM}                                                                            & MNIST                                                       & 0.01 $\pm$ 0.10                     & 0.08 $\pm$ 0.14                     & 0.01 $\pm$ 0.02                     & 0.08 $\pm$ 0.10                              & \color{blue}{\textbf{0.17 $\pm$ 0.15}}                     & \color{red}{\textbf{0.27 $\pm$ 0.17}}                      & 0.01 $\pm$ 0.11                     & 0.06 $\pm$ 0.13                     & 0.01 $\pm$ 0.02                     & 0.08 $\pm$ 0.10                              & \color{blue}{\textbf{0.15 $\pm$ 0.12}}                     & \color{red}{\textbf{0.24 $\pm$ 0.15}}                      \\
                                                                                                 & K-MNIST                                                     & 0.00 $\pm$ 0.04                     & 0.05 $\pm$ 0.09                     & 0.01 $\pm$ 0.02                     & 0.11 $\pm$ 0.10                              & \color{blue}{\textbf{0.18 $\pm$ 0.17}}                     & \color{red}{\textbf{0.22 $\pm$ 0.15}}                      & 0.00 $\pm$ 0.04                     & 0.05 $\pm$ 0.09                     & 0.01 $\pm$ 0.02                     & 0.11 $\pm$ 0.10                              & \color{blue}{\textbf{0.15 $\pm$ 0.16}}                     & \color{red}{\textbf{0.21 $\pm$ 0.14}}                      \\
                                                                                                 & \begin{tabular}[c]{@{}l@{}}CIFAR-10\\ (VGG-3)\end{tabular}  & 0.14 $\pm$ 0.13                     & 0.20 $\pm$ 0.14                     & 0.08 $\pm$ 0.04                     & \color{blue}{\textbf{0.14 $\pm$ 0.10}}                              & 0.12 $\pm$ 0.11                     & \color{red}{\textbf{0.19 $\pm$ 0.10}}                      & 0.13 $\pm$ 0.12                     & \color{blue}{\textbf{0.19 $\pm$ 0.14}}                     & 0.08 $\pm$ 0.04                     & 0.15 $\pm$ 0.11                              & 0.11 $\pm$ 0.11                     & \color{red}{\textbf{0.22 $\pm$ 0.13}}      \\                
                                                                                                 & \begin{tabular}[c]{@{}l@{}}CIFAR-10\\ (RESNET-18)\end{tabular}  & 0.02 $\pm$ 0.02                     & 0.17 $\pm$ 0.10                     & \multicolumn{1}{c}{0.08 $\pm$ 0.03}                     & \color{blue}{\textbf{0.21 $\pm$ 0.06}}                              & 0.20 $\pm$ 0.10                     & \color{red}{\textbf{0.22 $\pm$ 0.10}}                      & 0.03 $\pm$ 0.02                     & \color{blue}{\textbf{0.22 $\pm$ 0.13}}                     & \multicolumn{1}{c}{0.08 $\pm$ 0.02}                     & 0.20 $\pm$ 0.07                              & 0.17 $\pm$ 0.10                     & \color{red}{\textbf{0.23 $\pm$ 0.11}}                     \\ \hline
\multirow{4}{*}{SHIFT}                                                                            & MNIST                                                       & 0.11 $\pm$ 0.07                     & \color{blue}{\textbf{0.24 $\pm$ 0.11}}                     & 0.02 $\pm$ 0.02                     & 0.16 $\pm$ 0.09                              & 0.23 $\pm$ 0.16                     & \color{red}{\textbf{0.33 $\pm$ 0.15}}                      & 0.12 $\pm$ 0.07                     & \color{blue}{\textbf{0.21 $\pm$ 0.10}}                     & 0.02 $\pm$ 0.02                     & 0.16 $\pm$ 0.10                              & 0.18 $\pm$ 0.12                     & \color{red}{\textbf{0.27 $\pm$ 0.12}}                      \\
                                                                                                 & K-MNIST                                                     & 0.03 $\pm$ 0.04                     & 0.19 $\pm$ 0.06                     & 0.04 $\pm$ 0.01                     & 0.24 $\pm$ 0.04                              & \color{blue}{\textbf{0.39 $\pm$ 0.07}}                     & \color{red}{\textbf{0.44 $\pm$ 0.05}}                      & 0.02 $\pm$ 0.05                     & 0.19 $\pm$ 0.06                     & 0.03 $\pm$ 0.01                     & 0.24 $\pm$ 0.04                              & \color{blue}{\textbf{0.35 $\pm$ 0.07}}                     & \color{red}{\textbf{0.40 $\pm$ 0.05}}                      \\
                                                                                                 & \begin{tabular}[c]{@{}l@{}}CIFAR-10\\ (VGG-3)\end{tabular}  & \color{blue}{\textbf{0.35 $\pm$ 0.02}}                     & \color{red}{\textbf{0.42 $\pm$ 0.05}}                     & 0.05 $\pm$ 0.01                     & 0.28 $\pm$ 0.01                              & 0.31 $\pm$ 0.02                     & 0.27 $\pm$ 0.12                      & 0.35 $\pm$ 0.02                     & \color{red}{\textbf{0.41 $\pm$ 0.01}}                     & 0.05 $\pm$ 0.02                     & 0.31 $\pm$ 0.01                              & 0.31 $\pm$ 0.02                     & \color{blue}{\textbf{0.36 $\pm$ 0.13}}      \\                
                                                                                                 & \begin{tabular}[c]{@{}l@{}}CIFAR-10\\ (RESNET-18)\end{tabular}  & 0.09 $\pm$ 0.01                     & \color{red}{\textbf{0.39 $\pm$ 0.01}}                     & \multicolumn{1}{c}{0.08 $\pm$ 0.03}                     & 0.31 $\pm$ 0.01                              & 0.37 $\pm$ 0.01                     & \color{blue}{\textbf{0.38 $\pm$ 0.00}}                      & 0.09 $\pm$ 0.01                     & \color{red}{\textbf{0.46 $\pm$ 0.01}}                     & \multicolumn{1}{c}{0.09 $\pm$ 0.02}                     & 0.33 $\pm$ 0.01                              & 0.37 $\pm$ 0.02                     & \color{blue}{\textbf{0.43 $\pm$ 0.01}}                      \\ \hline
\end{tabular}%
}
\caption{Average Point Biserial correlation (left) and average Spearman correlation (right) coefficients (with respect to prediction errors) of different methods for all datasets and corruption types over all corruption intensities. Best values are boldened in red and second best in blue.}
\label{err}
\end{table*}

\end{document}